# Multi-Hypothesis Classifier

## A MASTER'S THESIS

submitted for the degree

of

## MASTER OF TECHNOLOGY

in

## INFORMATION TECHNOLOGY
(*Spln. in Intelligent Systems*)

Submitted by

**Sayantan Sengupta**

**Enrol. No.  IIS2012009**

Under the supervision of

## Dr. Sudip Sanyal
Professor
IIIT-Allahabad

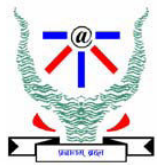

## INDIAN INSTITUTE OF INFORMATION TECHNOLOGY
## ALLAHABAD – 211012 (INDIA)

## June, 2014

# CANDIDATE'S DECLARATION

I do hereby declare that the work presented in this thesis entitled **"Multi-Hypothesis Classifier"**, submitted for the degree of Master of Technology in Information Technology (*Spln. in Intelligent Systems*) at Indian Institute of Information Technology, Allahabad, is an authentic record of my original work carried out under the guidance of **Prof. Sudip Sanyal** due acknowledgements have been made in the text of the thesis to all other material used. This thesis work was done in full compliance with the requirements and constraints of the prescribed curriculum.

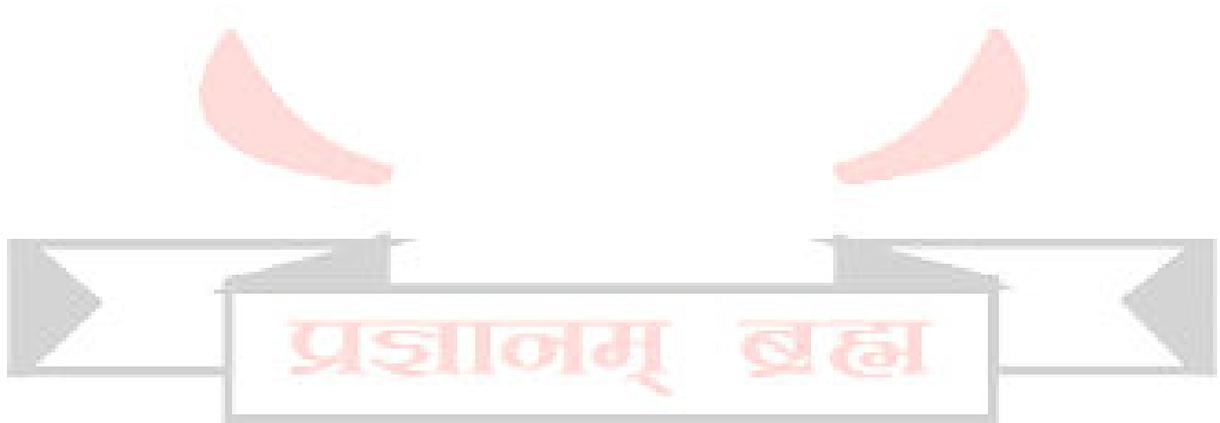

Place: Allahabad
Date:

**Sayantan Sengupta**
Enrol.No. IIS2012009

# CERTIFICATE FROM SUPERVISOR

I do hereby recommend that the thesis work prepared under my supervision by **Sayantan Sengupta** entitled **"Multi-Hypothesis Classifier"** be accepted in the partial fulfilment of the requirements of the degree of Master of Technology in Information Technology for Examination

Place: Allahabad                                                            **Dr. Sudip Sanyal**
Date:                                                                            Professor, IIITA

Countersigned by Dean (Academics) _________________________

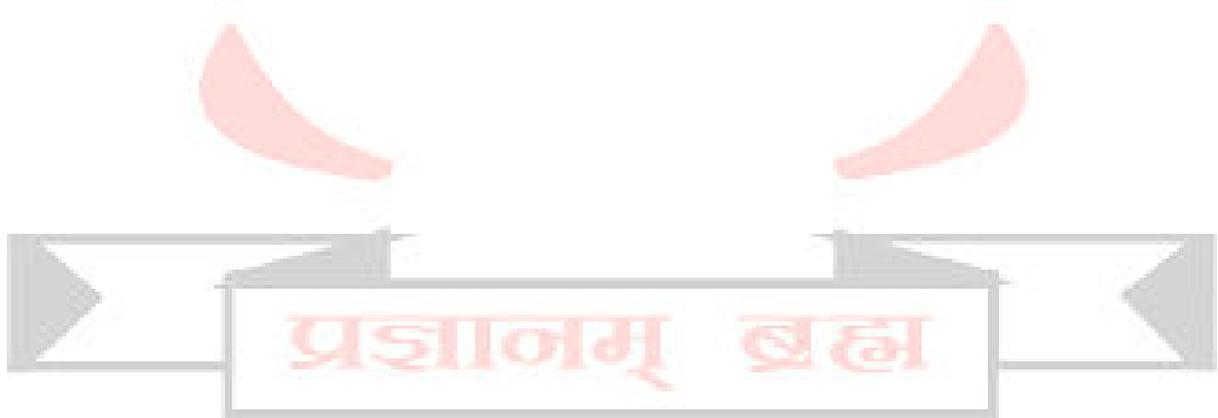

# CERTIFICATE OF APPROVAL[*]

The forgoing thesis is hereby approved as a credible study in the area of Information Technology/Electronics and Communication Engineering and its allied areas carried out and presented in a manner satisfactory to warrant its acceptance as a prerequisite to the degree for which it has been submitted. It is understood that by this approval the undersigned do not necessarily endorse or approve any statement made, opinion expressed or conclusion drawn therein but approve the thesis only for the purpose for which it is submitted.

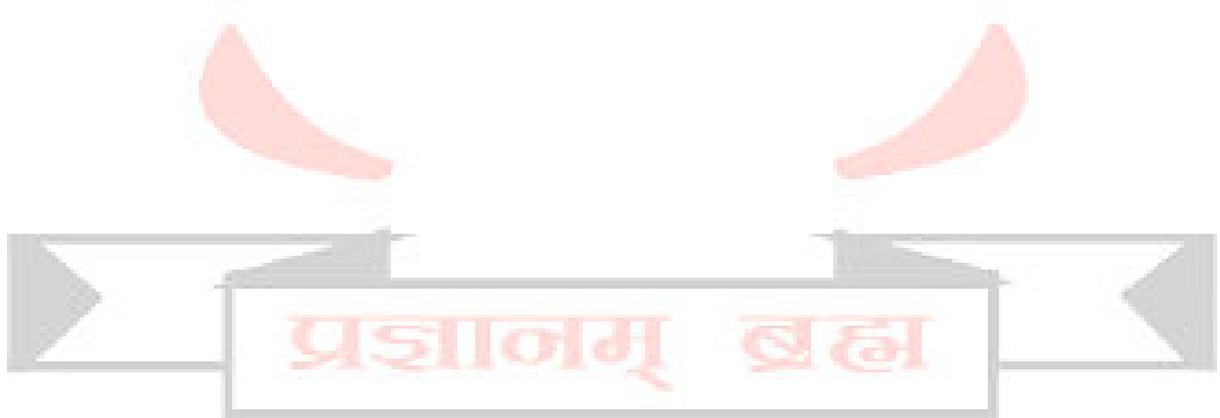

Signature & Name of the Committee members ________________________________________

On final examination and approval of the thesis________________________________________

**\*(Only in case the recommendation is concurred in)**

# PLAGIARISM REPORT

Subject: Plagiarism Report of the **M. Tech Thesis** Title **"Multi-Hypothesis Classifier".**

**Author: - Sayantan Sengupta (IIS2012009)**

Under Supervision of **Prof. Sudip Sanyal**

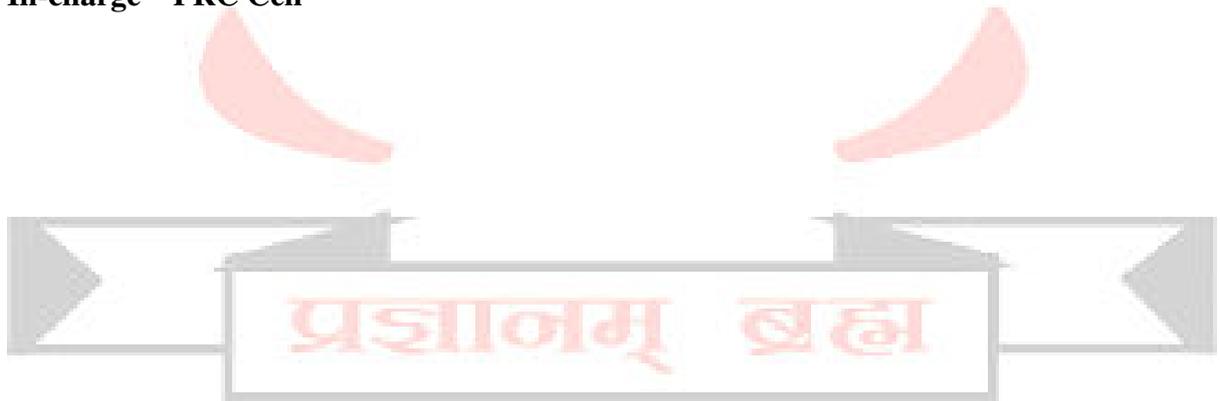

1) Reference is invited to the communication on the above subject dated 08-07-2014. PRC has performed the required plagiarism check. It has been observed that submitted **M. Tech thesis has a similarity index of 12%.**

2) The article having **Less than 15% similarity index** is considered as **Plagiarism Free.**

3) The present plagiarism check excludes plagiarism in Diagrams, Picture, Quoted materials, Bibliographic materials, and small matches (less than 1% from a document).

**In-charge – PRC Cell**

# ACKNOWLEDGEMENT

As understanding of the study like this is never the outcome of the efforts of a single person, rather it bears the imprint of a number of persons who directly or in directly helped me in completing the present study. I would be failing in my duty if I don't say a word of thanks to all those whose sincere advise make me this documentation of topic a real educative, effective and pleasurable one.

It is my privilege to study at Indian Institute of Information Technology, Allahabad where students and professors are always eager to learn new things and to make continuous improvements by providing innovative solutions. I am highly grateful to the honorable Director, IIIT-Allahabad**, Prof. Somenath Biswas**, for his ever helping attitude and encouraging us to excel in studies. I am also gratified to **Prof. G. C. Nandi** IIIT -Allahabad for providing me resources and flexibility for this dissertation work.

Regarding this thesis work, first and foremost, I would like to heartily thank my supervisor **Prof. Sudip Sanyal** for his able guidance. His fruitful suggestions, valuable comments and support were an immense help for me. In spite of his hectic schedule he took pains, with smile, in various discussions, which enriched me with new enthusiasm and vigour.

I am blessed with such wonderful family without their love, support, and encouragement, more than anything else; I would have never reached this stage in my life. I was provided with everything I required. I thank them all for all their love and support. I hope that with the completion of this course, I have made them proud.

Special thanks to Safeer Afaque and Nitish Sinha for debugging my code.

# Abstract

Accuracy is the most important parameter among few others which defines the effectiveness of a machine learning algorithm. Higher accuracy is always desirable. Now, there is a vast number of well established learning algorithms already present in the scientific domain. Each one of them has its own merits and demerits. Merits and demerits are evaluated in terms of accuracy, speed of convergence, complexity of the algorithm, generalization property, and robustness among many others. Also the learning algorithms are data-distribution dependent. Each learning algorithm is suitable for a particular distribution of data. Unfortunately, no dominant classifier exists for all the data distribution, and the data distribution task at hand is usually unknown. Not one classifier can be discriminative well enough if the number of classes are huge. So the underlying problem is that a single classifier is not enough to classify the whole sample space correctly.

This thesis is about exploring the different techniques of combining the classifiers so as to obtain the optimal accuracy. Three classifiers are implemented namely plain old nearest neighbor on raw pixels, a structural feature extracted neighbor and Gabor feature extracted nearest neighbor. Five different combination strategies are devised and tested on Tibetan character images and analyzed.

# Table of Contents

# List of Figures

# List of Tables

# *Chapter 1*

## *Introduction*

## 1.1 Overview

Accuracy is the most important parameter among few others which defines the effectiveness of a machine learning algorithm. Higher accuracy is always desirable. Now, there is a vast number of well established learning algorithms already present in the scientific domain. Each one of them has its own merits and demerits. Merits and demerits are evaluated in terms of accuracy, speed of convergence, complexity of the algorithm, generalisation property, and robustness among many others. Also the learning algorithms are data-distribution dependent. Each learning algorithm is suitable for a particular distribution of data. Unfortunately, no dominant classifier exists for all the data distribution, and the data distribution task at hand is usually unknown. Not one classifier can be discriminative well enough if the number of classes are huge. So the underlying problem is that a single classifier is not enough to classify the whole sample space correctly. Before we dive deep in to the topic, we go through the definition of "*classifier*".

**Classifier:** A "*Classifier*" is any mapping from the space of features (measurements) to a space of class labels (names, tags). A classifier is hypothesis about the real relation between features and class labels. A "*learning algorithm*" is a method to construct hypotheses. A learning algorithm applied to a set of samples (training set) outputs a classifier.

### 1.1.1 Rationale

In any application, we can use several learning algorithms. We can try many and choose the one with the best cross-validation results. On the other hand, each learning model comes with a set of assumption and thus bias. Learning is an ill-posed problem (finite data), each model converges to a different solution and fails under different circumstances. Why do not we combine multiple learners intelligently, which may lead to improved results? Now having said all this, the next question on our mind is *Is it really going to work?* If yes, then Why? What is the rationale behind it?

## 1.1.2 Why it Works?

Assume that we have 25 base classifiers. Each of the 25 base classifiers has an error rate, ε=0.35(say). Now if the base classifiers are identical, then the same examples will be misclassified by the ensemble as incorrectly predicted by the base classifiers. Assume classifiers are independent. The wrong prediction will be made only if more than half of the base classifiers are wrong in their prediction. Probability that the ensemble classifier makes a wrong prediction is $\sum_{i=13}^{25} \binom{25}{i} * \varepsilon^i (1-\varepsilon)^{25-i} = 0.06$. See the significant decrease in the error. The base classifiers should be chosen such that it does better than a random guessing. Although, it is very hard to have classifiers perfectly independent of each other.

One important point to be noted:

When multiple base learners are generated, it is not required for the individual classifiers to be very accurate, so it is not necessary for them to be optimised separately. The base learners are chosen for their simplicity, not for their accuracy.

## 1.2 Motivation

**Why do we need multiple classifiers?**

Now having gone through the definition of "classifier", the question arises is what can be the solution for this limitation of having single classifier? Is there a way, by which we can use the positive points (discriminating features) of the individual classifier to achieve an overall higher accuracy than the best accuracy obtained by a single learning algorithm? The solution is "Ensemble Classifiers" which make use of the discriminating abilities of the individual classifiers and fuses them together. Having said this, now many questions arise about the techniques, methodologies, guarantee of accuracy of the fusion strategies. These issues would be discussed in the following pages.

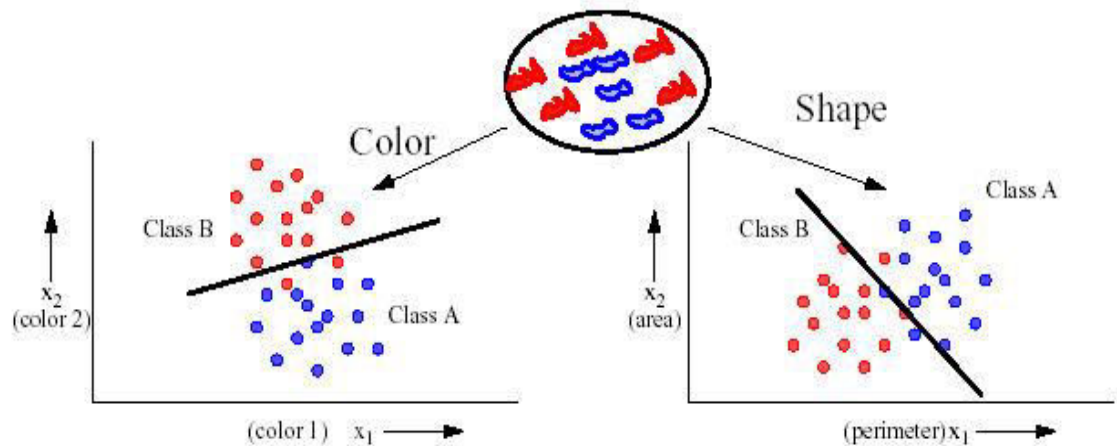

**Figure 1.1: Several classifiers in different feature spaces.**

One more advantage of "Ensemble Classifier" is, besides avoiding the selection of the worse classifier, under practical hypothesis, fusion of multiple classifiers can improve the performance of the best individual classifiers and, in some special cases, provide the optimal Bayes classifier. Also they are more resilient to noise. But, there is a necessary condition which must be fulfilled to achieve this. The necessary condition is that the individual classifiers make "different" errors. In other words, the classifiers must be independent of each other. They must not make the same mistake. Illustrating with an example, given a sample space, which is to be classified to two classes, namely CLASS-A and CLASS-B. Also two classifiers are given namely Classifier-1 and Classifier-2. So, if Classifier-1 correctly classifies the majority of samples of CLASS-A and incorrectly classifies the majority of samples of CLASS-B, then for the "Multi-hypothesis" to work, the Classifier-2 must be complementary to this, i.e. Classifier-2 must correctly classify the majority of the samples from CLASS-B.

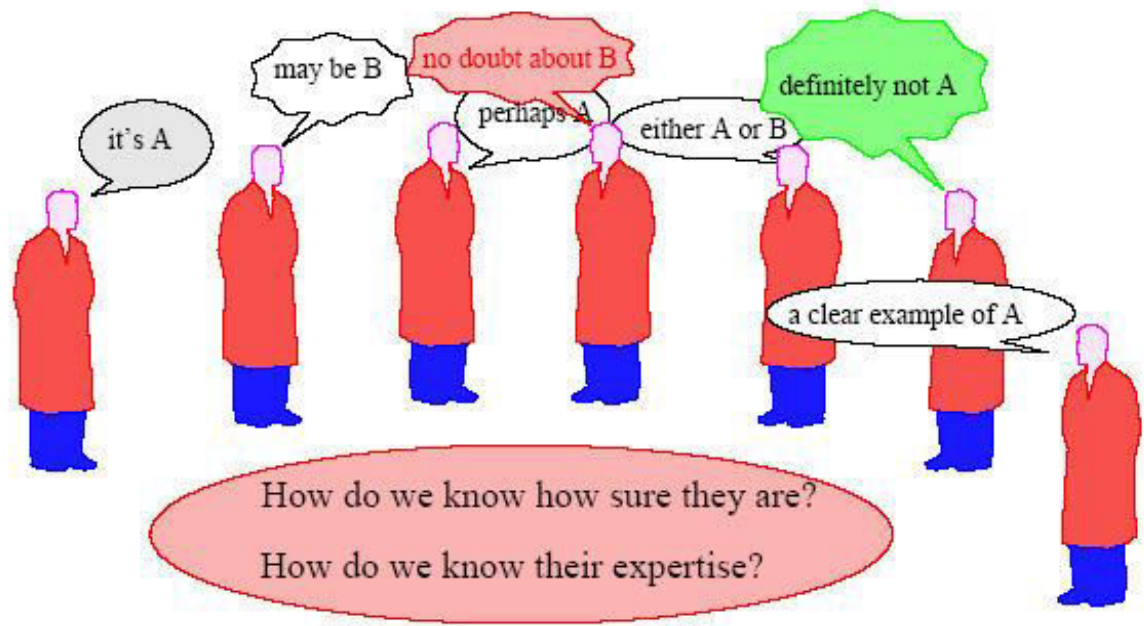

**Figure 1.2: Representation of the dilemma of several experts to reach a consensus.**

There are many advantages to an ensemble of classifiers. Three main reasons on why a classifier ensemble might be better than single classifier [1].

## 1.3    Reasons for the usefulness of  ensemble classifier

### 1.3.1  Statistical

Let us assume that there are a number of classifiers with a decent performance on a given labelled dataset as shown in Figure 1.1. Generalisation performance is different on the data for each of the classifiers. There is always a choice to pick a single classifier for solution, which may lead to a bad decision. A better choice would be to use multiple classifiers and 'average' their outputs. The new ensemble classifier may not improve the results from the best individual classifier but will eliminate the risk of picking an insufficient single classifier.

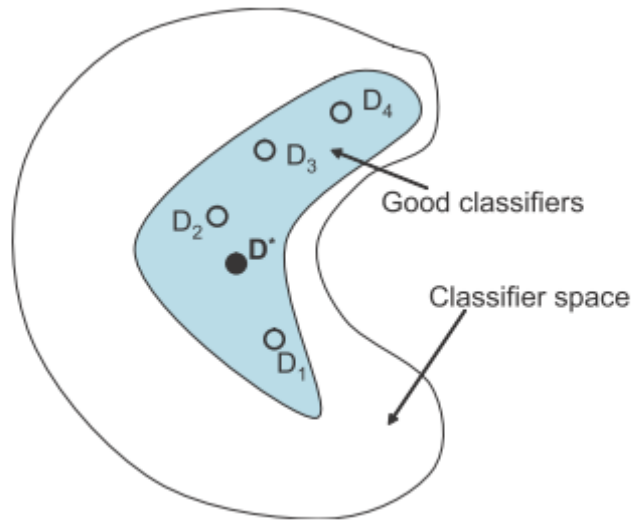

**Figure 1.3: D\* is the best performing individual classifier; the space of all classifiers is depicted as the outer curve; the shaded area depicts the space of good performance classifiers.**

### 1.3.2 Computational

Some of the learning algorithms perform hill-climbing or random search, which has a great chance to lead to a local optima as shown in Figure 1.2.

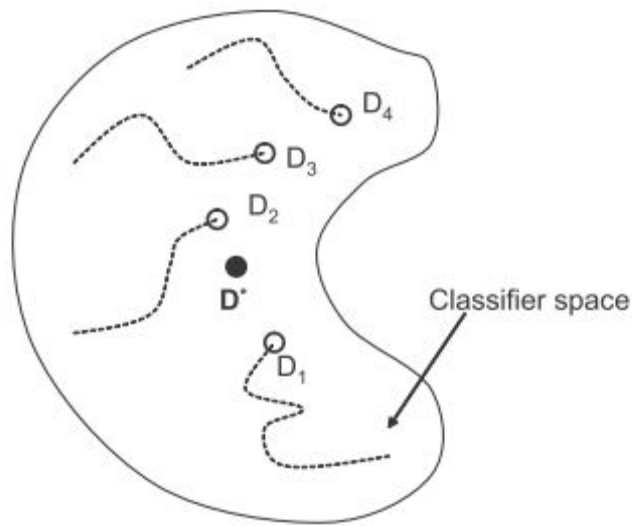

**Figure 1.4: D\* is the best classifier for the data; the space of all classifiers is shown by the outer curve; the dashed lines represents the hypothetical path or trajectories for the classifiers during training.**

We go with the assumption that the training process of each classifier starts somewhere in the space of possible classifiers and terminates closer to the optimal classifier D*. So some form of mixing or aggregating may lead to classifier that is a better approximation to D* than any single classifier Di.

### 1.3.3  Representation

It might be possible that the domain space of the classifier considered for the problem does not contain the optimal classifier. For example, for the banana dataset given in figure 1.3, the optimal classifier is nonlinear. So if we restrict the domain of all possible classifiers to linear classifiers only, then the optimal classifier for the problem will not belong in this domain. However, an ensemble of linear classifiers can approximate any decision boundary with some decent accuracy. If the classifier space is different, then the optimal classifier D* may be an element of it. The argument here is that the training an ensemble to obtain a certain high accuracy is more straightforward than training a classifier directly to achieve high complexity.

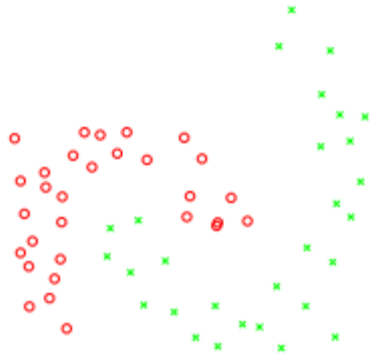

**Figure 1.5: Banana dataset; Optimal classifier is nonlinear.**

However, an improvement on the best individual classifier or on the ensemble's average performance for the general case is not guaranteed always. But still, the experimental work done so far in this field and the theories developed for a number of special cases illustrates the success of the methods of combining classifiers [2]. The fusion strategy obviously affects the improvement which aims to combine the diverse information obtained from the multiple experts. A methodical approach will be to analyse the information obtained from the different sources and find the best fusion strategy. But there is very little to be gained from combining, irrespective of the chosen scheme if the classifiers make the same mistakes, according to Turner and Ghosh [3]. In many cases we have the flexibility to create multiple classifiers. In such scenarios, we can select a fusion strategy and then a set of multiple classifiers can be formed which contains the diverse information which helps in increasing the accuracy.

## 1.4  Multiple Classifier System (MCS)

A multiple classifier system (MCS) is a structured way to combine (exploit) the outputs of individual classifiers. MCS can be thought as multiple expert systems, committees of experts, mixtures of experts, classifier ensembles, and composite classifier systems. Multiple classifier system (MCS) can be characterized by:

- The Architecture
- Fixed/Trained Combination strategy
- Others

### 1.4.1  MCS Architecture/Topology

#### 1.4.1.1   Serial

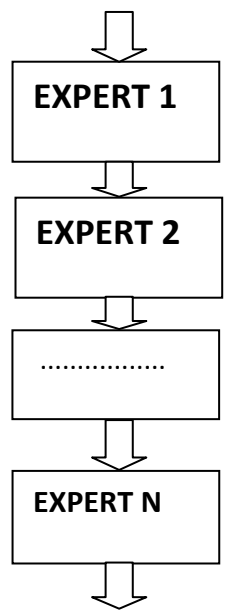

**Figure 1.6: Serial architecture.**

### 1.4.1.2 Parallel

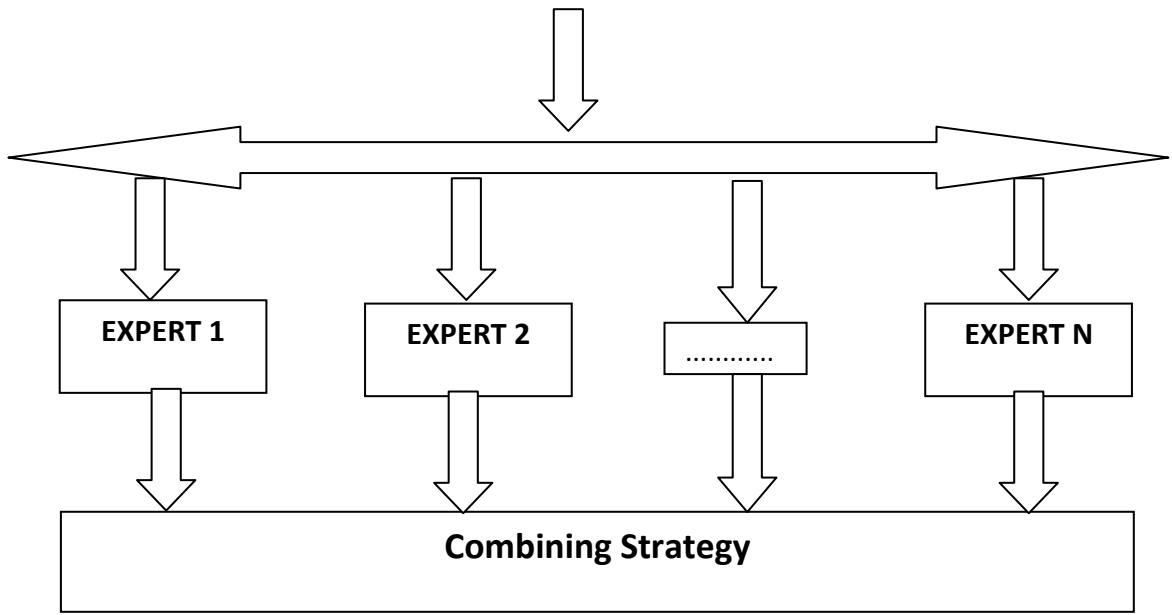

**Figure 1.7: Parallel architecture.**

### 1.4.1.3 Hybrid

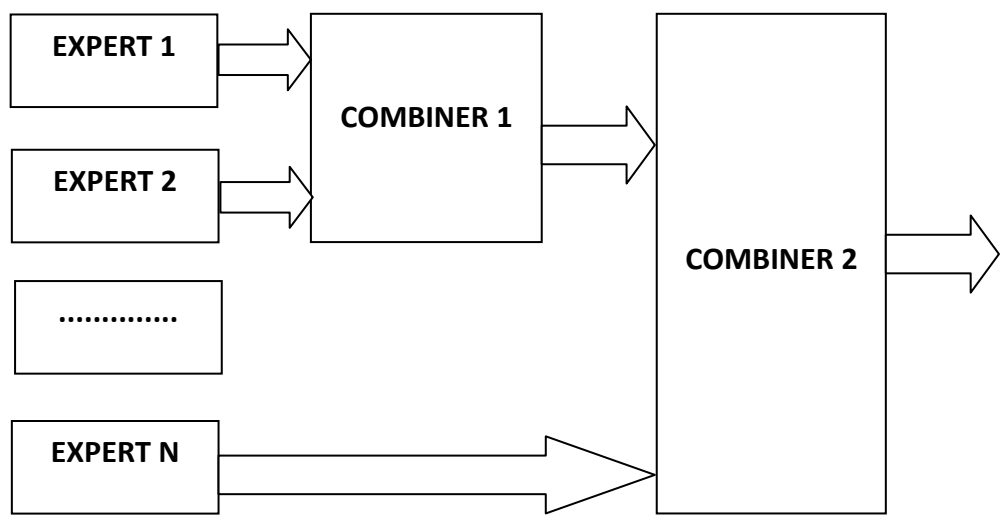

**Figure 1.8: Hybrid architecture.**

Now, having gone through the basic architectures for multiple classifier systems, the next logical question arise is, what are the multiple classifier source? Sources can be differentiated based on,

1    Different Feature spaces (Face, voice, fingerprint)

2    Different Training sets (Sampling, Boosting, Bagging)

3    Different Classifiers (KNN,ANN,SVM)

4    Different Architectures (Neural Net: layers, Units, Transfer function)

5    Different parameter values ( k in kNN, Kernel in SVM)

6    Different initialisations

**Combination based on a single space but different classifiers**

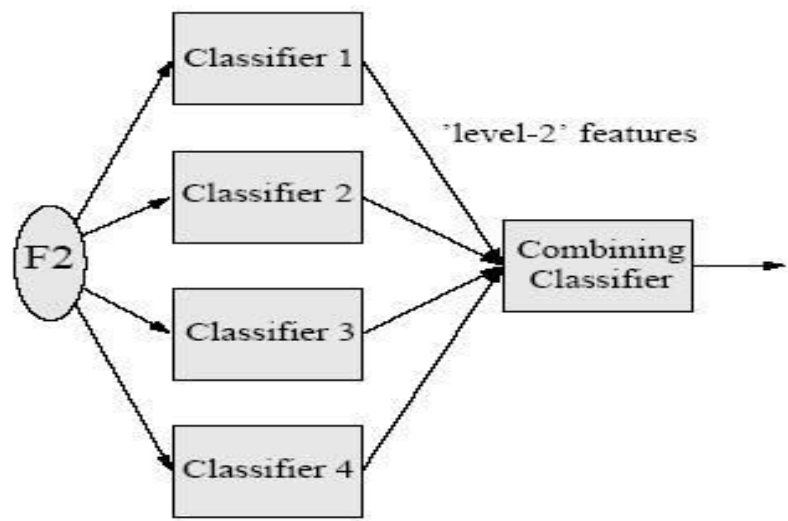

Figure 1.9: Single feature set, different classifiers.

**Combination based on different feature spaces**

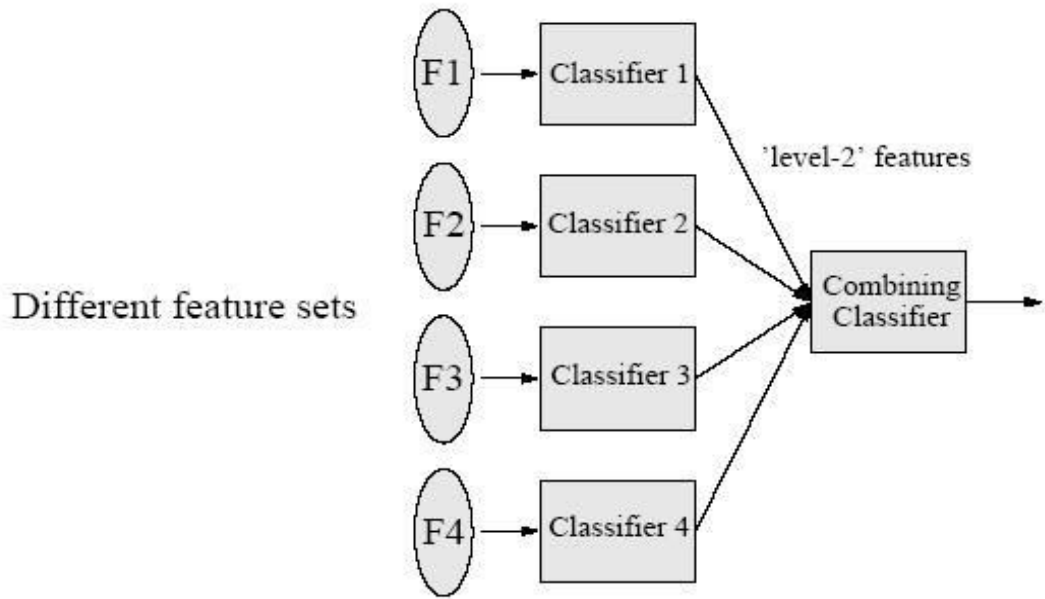

**Figure 1.10: Different feature space.**

### 1.4.2 Fixed Combination Strategy

Some of the popular fixed combination rules are Product Rule (minimum), Sum rule (mean), Median rule, Majority vote and Maximum rule. These rules will be discussed in the later chapter in details. Some features of each of the rules are as below:

➢ **Product Rule** (Minimum)
  - Independent feature spaces
  - Different areas of expertise
  - Error free posterior probability estimates

➢ **Sum (Mean), Median, Majority Vote**
  - Equal posterior-estimation distribution in same feature space.
  - Differently trained classifiers, but drawn from the same distribution
  - Bad if some classifiers (experts) are very good or very bad.

➢ **Maximum Rule**
  - Trust the most confident classifier/expert
  - Bad if some classifiers (experts) are badly trained.

Having listed the above rules, one important factor which put it in disadvantage than the trained combiner is that the fixed combining rules are sub-optimal. Base classifiers are never really independent (Product Rule).Base classifiers are never really equally imperfectly trained (sum, median, majority vote). Also the sensitivity to over-confident base classifiers (Product, min, max). This leads to the emergence of the emergence of the trained combiners.

### 1.4.3   Trained Combiners

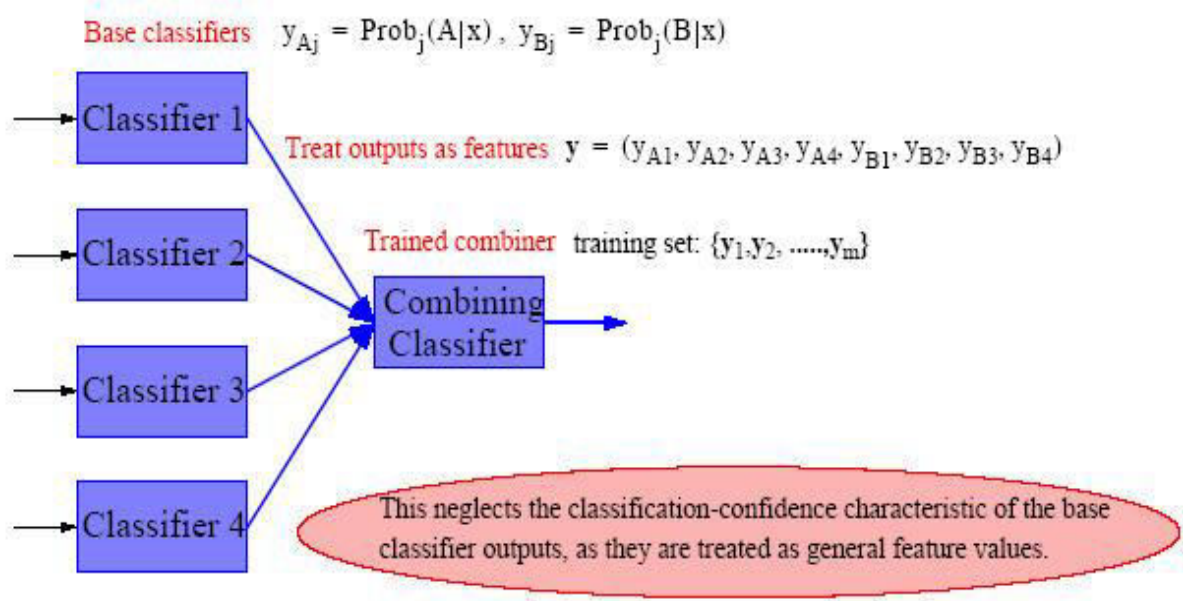

**Figure 1.11: Trained combiners**

Trained combiners perform potentially better than the fixed rules because of their flexibility. Trained rules are claimed to be more suitable than the fixed ones for classifiers correlated or exhibiting different performances. Trained combiner neglects the classification confidence characteristic of the base classifier output, as they are treated as general feature values. One disadvantage of the trained rules from fixed rules is that it takes high memory and requires large time to get trained. Fixed rules are known for their simplicity which makes them easier to use whereas trained rules are more complex.

# *Chapter 2*

## *Literature Survey*

### *What has already been done?*

## 2.1 Boosting

Boosting is a old effective way to combine classifiers. It was introduced by Schapire and Freund in 1990s. Boosting convert a weak learning algorithm into a strong one. Having said that, the analogy is still not clear. Let me summarize how the pioneers of this field Freund and Schapire approached the understanding of this method.

***2.1.1 Analogy:*** "There was a gambler, who was frustrated by the non-stop losses in the horse racing and was jealous of his friend's success in the same events. Not being able to find out the reason behind his failure, he allows his group of best pals cum gamblers to make bets on his behalf. He makes up his mind that he will wager a fixed sum of money in every race. But there was also a catch. He decides that he will proportionally divide his money among his friends depending on how well they are performing. Certainly, if he knew psychically beforehand which of his friend would win the most, he would obviously have that friend handle all his wagers. But due to the lack of such clear vision, he allocates each race's wager in such a way that his total earning for the whole season will be reasonably close to what he would have won had he bet on his luckiest friend . This method is all about dynamic allocation problem.

Now suppose the gambler gets tired of choosing among the experts and instead he would like to create a computer program which would predict the winner of the race using some usual information (races won by individual horses, betting odds, etc.). To create this sort of a program, he asks all his experts to articulate their strategy. Not surprisingly, the experts are unable to come up with a grand set of rules for selecting a horse. On the other hand they were able to come up with a solution when presented data for a specific set of races. Such rules of thumbs (horse with most favoured odds) are usually very inaccurate and rough.

They are expected to give results which might be slightly better than random guessing, which will not be unreasonable to expect. By asking the experts their opinion again and again on different collection of races, the gambler is able to gather many rules-of-thumb".

Now in order to use these rules to maximum advantage, there arise two issues.

First, how must he choose the collection of races presented to the expert so as to extract the most useful rules of thumb from the expert? Second, how can the rules be combined to form a single highly effective and accurate rule, once he has collected many rules-of-thumb? Boosting provides the combination technique to for accurate prediction for moderately inaccurate rules-of-thumb.

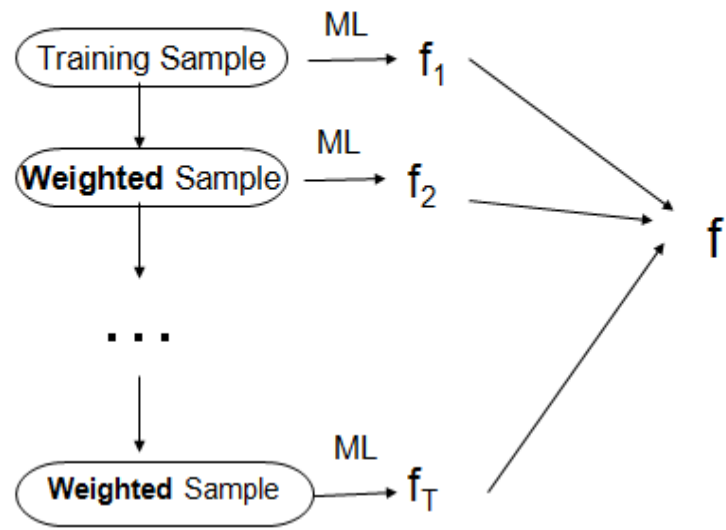

**Figure 2.1: Flowchart of Boosting.**

Now, having established the analogy lets dive in to the technical know-how of this method. The main idea is to combine many weak classifiers to produce powerful committee. So the classifiers are produced sequentially, one after the other. Each classifier is dependent on the previous one, and focuses on the previous one's errors. Examples that are incorrectly predicted in the previous classifiers are chosen more often or weighted more heavily. Records that are wrongly classified will have their weights increased. Records that are classified correctly will have their weights decreased. Given an example below,

| Original Data | 1 | 2 | 3 | 4 | 5 | 6 | 7 | 8 | 9 | 10 |
|---|---|---|---|---|---|---|---|---|---|---|
| Boosting (Round 1) | 7 | 3 | 2 | 8 | 7 | 9 | 4 | 10 | 6 | 3 |
| Boosting (Round 2) | 5 | 4 | 9 | 4 | 2 | 5 | 1 | 7 | 4 | 2 |
| Boosting (Round 3) | 4 | 4 | 8 | 10 | 4 | 5 | 4 | 6 | 3 | 4 |

**Figure 2.2: Example of data sampling in subsequent rounds of boosting.**

As we can see from the above table that, example 4 is hard to classify. So its weight is increased, therefore it is more likely to be chosen again in the subsequent rounds. Boosting algorithm differs in terms of (1) how the weights of the training examples are updated at the end of each round, and (2) how the predictions made by each classifier are combined.

### 2.1.2 The basic Ada-Boosting Algorithm

For t=1 . . . , T

- Train weak learner using training data and $d_t$.

- Get $h_t : X \rightarrow \{-1,1\}$ with error $\varepsilon_t = \sum_{i:ht(xi) \neq yi} D_t(i)$

- Choose $\alpha_t = \frac{1}{2} \ln \frac{1 - \varepsilon_t}{\varepsilon_t}$

- Update $D_{t+1}(i) = \frac{D_t(i)}{Z_t} * e^{-\alpha t}$ if $h_t(x_i) = y_i$

$$D_{t+1}(i) = \frac{D_t(i)}{Z_t} * e^{-\alpha t} \quad \text{if} \quad h_t(x_i) \neq y_i$$

$$= \frac{D_t(i)}{Z_t} e^{\alpha_t y_i h_t(x_i)} , \qquad (2.1)$$

Where $Z_t$ is the normalisation factor (chosen so that $D_{t+1}$ will be a distribution).

The hypothesis weight $\alpha_t$ is decided at round t. The weight distribution of training examples is updated at every round t. But Boosting comes with its own sets of issues. The main issues are-

➢ Given $h_t$, how to choose $\alpha_t$ ?

➢ How to select $h_t$ ?

➢ How to deal with multi-class problems?

### 2.1.3 Strengths of Ada-boost

- It has no parameters to tune (except for the number of rounds)
- It is fast, simple and easy to program
- It comes with a set of theoretical guarantee (training error, test error)
- Instead of designing a learning algorithm that is accurate over the entire space, we can focus on finding the base learning algorithms that only need to be better than random.
- It can identify outliners i.e. examples that are either mislabelled or that are inherently ambiguous and hard to categorize.

### 2.1.4 Weakness of Ada-Boost

- The actual performance of boosting depends on the data and the base learner.
- Boosting seems to be especially susceptible to noise.
- When the number of outliners is very large, the emphasis placed on the hard examples can hurt the performance.

### 2.2  Bagging

Introduced by Breiman[4]. Derived from bootstrap (Efron,1993). It creates classifiers using training sets that are bootstrapped (drawn with replacement).Each bootstrap sample D has the same size as the original data. Some instances could appear several times in the training set, while others may be omitted. The idea is to build classifier on each bootstrap sample D. D will contain approximately 63% of the original data. Each data object has probability of $1-(1-\dfrac{1}{n})^{n}$ of being selected in D. Bagging improves generalization performance by reducing the variance of the base classifiers.

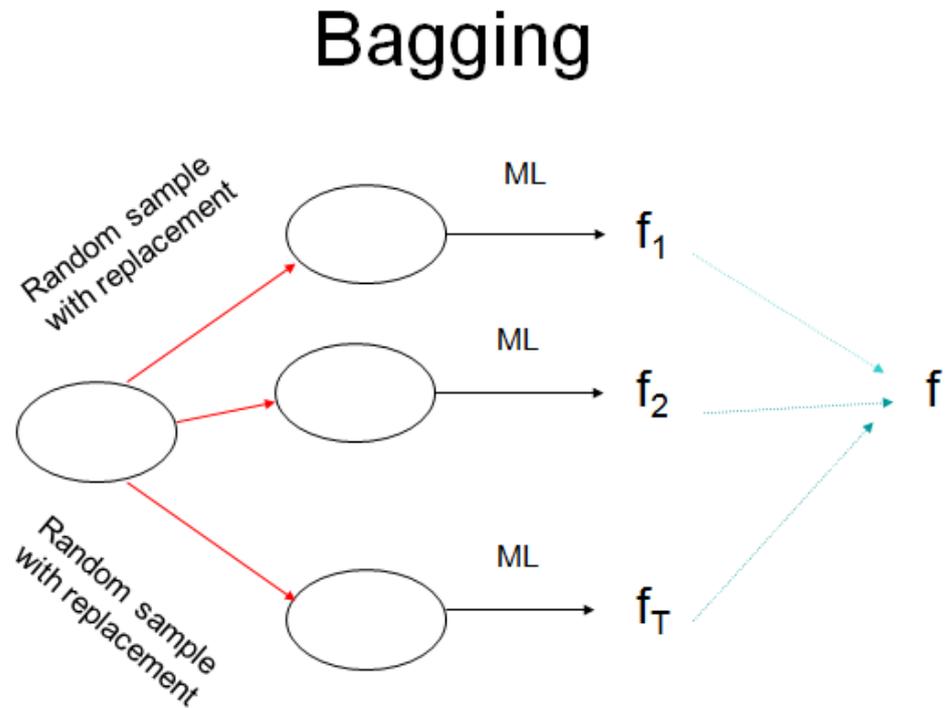

**Figure 2.3: Flow chart of Bagging**

The performance of the bagging depends on the stability of the base classifier. If a base classifier is unstable, bagging helps to reduce the errors associated with random fluctuations in the training data. If the base classifier is stable, bagging may not be able to improve; rather it could degrade the performance. Also one more advantage of Bagging is that, it is less susceptible to model over fitting when applied.

### 2.2.1  Bagging vs. Boosting

An example provided below shows the difference between the two techniques. The example shown below indicates that sample no. 1 is prone to error. So in boosting this sample has a higher chance of getting picked up in the set. Whereas in bagging, the sampling is independent of the previous set.

| Training Data |
|---|
| **1** , 2 , 3 , 4 , 5 , 6 , 7 , 8 |

| Bagging Training set | Boosting Training set |
|---|---|
| Set 1: 2, 7, 8,  3, 7, 6, 3, **1** | Set 1: 2, 7, 8, 3, 7, 6, 3, **1** |
| Set 2: 7, 8, 5, 6, 4, 2, 7, **1** | Set 2: **1**, 4, 5, 4, **1**, 5, 6, 4 |
| Set 3: 3, 6, 2, 7, 5, 6, 2, 2 | Set 3: 7, **1**, 5, 8, **1**, 8, **1**, 4 |
| Set 4: 4, 5, **1**, 4, 6, 4, 3, 8 | Set 4: **1**, **1**, 6, **1**, **1**, 3, **1**, 5 |

**Table 2.1: Training set (above); Data sampling in bagging and boosting.**

Some important conclusions from the above two techniques:

- Bagging always uses re-sampling rather than re-weighting.
- Bagging does not modify the distribution over examples or mislabels, but instead always uses the uniform distribution.
- In forming the final hypothesis, bagging gives equal weight to each of the weak hypotheses.

Experimental Results on Ensembles (Freund and Schapire, 1996; Quinlan, 1996) [5]:

- Ensembles improve generalization accuracy on a wide variety of problems.
- On average, Boosting provides a larger increase in accuracy than Bagging.
- Boosting on rare occasion can degrade accuracy.
- Bagging more consistently provides a modest improvement.
- Boosting is particularly subject to over-fitting when there is significant noise in the training data.

## 2.3   Issues in Ensembles

- Parallelism in ensembles: Bagging is easily parallelized whereas Boosting is not.
- How "weak" should a base learner for Boosting be?
- Variants of Boosting to handle noisy data.
- What is the theoretical explanation of Boosting's ability to improve generalization?

- Exactly how does the diversity of ensembles affect their ensemble performance?
- Combining Boosting and Bagging.

The above two techniques exploits the fact that multiple training set leads to the formation of multiple classifiers. Instead we can design multiple classifiers with the same dataset. Now broadly there are two large groups of methods of classification, (1) Feature Vector based methods and (2) Structural and Syntactic methods. Also each group again includes different algorithms based on different methodologies. For the first group itself, there exists k-NN classifier, Bayes classifier, neural network based classifiers and various distance based classifiers among others. So there is a big challenge to devise an effective strategy/methodology to combine the outputs of all the different breeds of classifiers which outputs different levels of information. Also how to obtain a consensus on the result of each individual classifier?

Methods for fusing classifiers can be differentiated according to the type of information produced by the individual classifiers [7]. They are:

- **Abstract level:** A classifier only outputs a unique label for each input pattern.
- **Rank level:** Each classifier outputs a list of possible classes, with ranking for each input pattern.
- **Measurement level:** Each classifier outputs class confidence levels for each input pattern.

For each of the above categories, methods can be further subdivided into Integration vs. Selection rules and Fixed rules vs. trained rules. Example of a fixed rule at abstract-level is the *majority voting rule*.

Next we talk about Fuser (Combination rule). Broadly, fuser can be classified into two main categories:

- **Integration (fusion) function:** For each pattern, all the classifiers contribute to the final decision. Integration assumes competitive classifiers.
- **Selection functions:** For each pattern, just one classifier, or a subset, is responsible for the final decision. Selection assumes complementary classifiers.

Integration and selection can be merged for designing the hybrid fuser. For this, multiple functions for non-parallel architecture can be necessary.

### 2.3.1 Diversity in Classifiers

Next we talk briefly about *classifier's diversity*. Measure of diversity in classifier ensembles are a matter of ongoing research (L. I. Kuncheva). The key issue in this domain is: *How are the diversity measures related to the accuracy of the ensembles?*

Fusion is obviously useful only if the combined classifiers are mutually complementary (ideally, classifiers with and high diversity high accuracy). The required degree of error diversity depends on the fuser complexity. An example, the whole can be divided in to four diversity levels (A. Sharkey,1999) [11].

-   Level 1: No more than one classifier is wrong for each pattern.
-   Level 2: The majority is always correct.
-   Level 3: At least one classifier is correct for each pattern.
-   Level 4: All classifiers are wrong for some pattern.

Simple fusers can be used for classifiers that exhibit a simple complementary pattern (e.g. majority voting). Complex fusers, for example, a dynamic selector, are necessary for classifiers with a complex dependency model. The required complexity of the fuser depends on the degree of classifier's diversity.

The design of MCS (Multiple classifier systems) involves two main phases, (1) The design of the classifier ensemble, and (2) The design of the fuser. The design of the classifier ensemble is aimed to create a set of complementary/diverse classifiers. On the other hand, the design of the combination function/fuser is aimed to create a fusion mechanism that can exploit the complementary/diversity of the classifiers and optimally combine them. The above two phases are obviously linked. In this thesis, we focus more on the second phase i.e. design of the fuser.

### 2.4 Average Bayes Classifier and its versions

This section deals with the combination problem where the output from each classifier is available in measurement level. First, we consider that all classifiers are Bayes classifiers. For a Bayes classifier e, the classification is based actually on a real set of measurements-post probabilities:

$$P(x \in C_i/x), i = 1,........, M$$

$$(2.2)$$

Where sample x comes from class Ci. These probabilities are not related to individual classifier Ek. In practice, each classifier classifies a sample to a particular class is not really based on the true values of the above equation, which are not available. For each sample x, a set of approximations are estimated by classifier Ek itself. These approximations are based on what features the classifier is used on and how the classifiers are trained. To clarify such dependence, we denote it as follows:

$$P_k(x \in C_i/x), i = 1,.....,M, k = 1,.....,K.$$

(2.3)

The above equation gives the approximation for combining the classification results on the same sample by all the K classifiers. One simple approach given by Lei Xu (1992) [ ], calls for a little modification in the above approach. They use the following average value as a new estimation of the combined classifier:

$$P_E(x \in C_i/x) = \frac{1}{K}\sum_{k=1}^{K} P_k(x \in C_i/x), i = 1,.....,M$$

(2.4)

The final decision made by the classifier E is given by:

$$E(x) = j, \quad \text{with} \quad P_E(x \in C_j/x) = \max_{i \in \wedge} P_E(x \in C_i/x)$$

(2.5)

The above Bayes decision is based on the newly estimated post-probabilities. Such a combined classifier is called as an *Averaged Bayes classifier*. The above approach could as well be extended to cover several cases where there are different kinds of classifiers. Generally, any classifiers where some kinds of post probabilities are computable could be combined by means of equation (2.5).

## 2.5 Combining Multiple Classifiers using Voting Principle and its variants [8]

It may so happen that the individual classifier does not agree on a particular sample about which class it belongs to. A simple rule used for resolving this kind of disputes in human social life is by majority voting. Now, the basic principle can be altered/modified a bit to bring out the variants which may be suitable under certain conditions. First, for our convenience, let us represent the event $e_k(x) = i$ in a binary characteristic function form.

$$T_k(x \in C_i) = 1, \text{ when } e_k(x) = i \text{ and } i \in \wedge \qquad (2.6)$$

$$T_k(x \in C_i) = 0, \text{ otherwise.}$$

The most conservative voting rule is the following:

$$E(x) = j, \text{ if } \exists j \in \wedge \prod_{k=1}^{K} T_k(x \in C_j) > 0 \qquad (2.7)$$

$$E(x) = M + 1, \text{ Otherwise}$$

The above equation means that if all K classifiers decide that sample x belongs to class $C_j$ unanimously, then classifier E decides that x comes from $C_j$, otherwise it rejects x. In the above equation I denotes the logical AND operator or binary multiplication and in the following equations Y denotes the logical OR operator or binary summation.

A slight altering of the above equation will lead to a version which is less rigid as shown below.

$$E(x) = j, \text{ if } \exists j \in \wedge \prod_{k=1}^{K} \{T_k(x \in C_j) Y(1 - \coprod_{q=1}^{M} T_K(x \in C_q)\} > 0 \qquad (2.8)$$

$$E(x) = M + 1, \text{ Otherwise}$$

The above modified equation, means that sample x belongs to a particular class $C_k$ as long as some of the classifiers support that class $C_k$ and no other classifier support a different class. In other words, the classifiers that reject the sample x will have no impact on the combined classifier unless all the classifier reject x.

Now for a case, where there are more than two labels that receive the maximal votes or the maximum vote obtained is not much larger than the second maximum votes. In such cases, even if the vote received by a label is large, but there is equally another large vote which goes against the first label. To tackle this problem, a new kind of majority voting is proposed below:

$$E(x) = j, \quad T_E(x \in C_j) = \max_1 \text{ and } \max_1 - \max_2 \geq \alpha * K \qquad (2.9)$$

$$E(x) = M + 1, \text{ Otherwise}$$

Where $0 < \alpha <= 1$. As the number of classifiers (K) are constant. The vote of the second maximum is taken as the implicit objections to the label j.

## 2.6 Combination of Classifiers in Dempster-Shafer Formalism

This combination technique is useful only when the output information provided by each classifier is in abstract form, i.e. only class label information is provided as output. As a prior knowledge, only the substitution, recognition and rejection rates of each classifier are used.

### 2.6.1 Dempster-Shafer Theory [9, 10]:

Let me first give the key points of this theory. Given a number of exhaustive and mutually exclusive propositions $A_i, i = 1, \ldots, M$, of the universal set $\Theta = \{A_1, \ldots, A_M\}$. Any subset $\{A_{i1}, \ldots, A_{iq}\} \subset \Theta$ is a proposition denoting the disjunction $A_{i1} \curlyvee \ldots \curlyvee A_{iq}$. Each element $A_i \subset \Theta$ called a *singleton* which corresponds to a one element subset. All subsets possible of $\Theta$ forms a superset $2^\Theta$. The *Dempster-Shafer theory* uses a value in the range of 0 and 1, inclusive to give a belief in a proposition, given the occurrence of some evidence *e*. It is denoted as *bel(A),* gives information about the degree to which the evidence *e* supports the proposition A. The belief function *bel(A)* is calculated from another function which is known as the Basic Probability Assignment(BPA). This function gives information about the individual impact of each evidences on our propositions. The BPA is denoted by m, and can also be generalized as probability mass distribution. It assigns numeric values in the closed range of 0 and 1 to each subset of $\Theta$ i.e. every element of $2^{\Theta}$, such that the values sum up to 1. Three distinct features about BPA:

- m(A) is just a small part of the total belief assigned exactly to A. It can't be subdivided among the subsets of A.

- The singletons are only some of the part of the whole elements of the superset $2^{\Theta}$. So it is quite possible that $\sum_{i=1}^{M} m(A_i) < 1$, and also $A_i$ and $\neg A_i$ are the sole two elements of the superset $2^{\Theta}$, so the relation $m(A_i) + m(\neg A_i) < 1$ holds true. This violates the basic axioms of Bayesian theory. So the BPA provides an incomplete probabilistic model.

- A subset *A* of the superset $2^{\Theta}$ with *m(A)*>0 is called a focal element. When there is only one focal element in the superset, then m($\Theta$) absorbs the unassigned part of the total belief. m($\Theta$) = 1 − m(*A*).

Now as subset A is the disjunction of all elements in A, if B⊂A, then the truth of B implies the truth of A. Hence, the belief function bel(A) is given by,

$$bel(A) = \sum_{B \subseteq A} m(B)$$

(2.10)

If more than two evidences exists, two or more sets of BPA's and bel(.. )'s will be applied to the same subset of $\Theta$. If $m_1$, $m_2$ and $bel_1, bel_2$ denote the two BPA's and their corresponding belief functions respectively, then the D-S rule defines a new BPA given by $m = m_1 \oplus m_2$, which gives the combined effect of both $m_1$ and $m_2$, i.e. for A is not equal to an empty set:

$$m(A) = m_1 \oplus m_2(A) = k \sum_{X \cap Y = A, A \neq \Phi} m_1(X) m_2(Y)$$

(2.11)

Where,

$$k^{-1} = 1 - \sum_{X \cap Y = \Phi} m_1(X) m_2(Y) = \sum_{X \cap Y \neq \Phi} m_1(X) m_2(Y)$$

(2.12)

The necessary condition for the BPA to exist is : $k^{-1} \neq 0$. If $k^{-1} = 0$, then the two evidences are said to be in conflict, i,e. They are in total contradiction. $m_1 \oplus m_2$ does not exists. If there is no contradiction, then the overall belief function bel(A) can be calculated from the result of the combined BPA.

## 2.6.2 Modelling Multi classifier combination using Dempster-Shafer Theory

In this problem, the M mutually exclusive and exhaustive propositions are given by $A_1 = x \in C_i, \forall i \in \Lambda$ which denotes that sample $x$ comes from class label $C_i$. The universal proposition is $\Theta = \{A_1, \ldots, A_M\}$. There are K classifiers $e_1, \ldots, e_K$ which will give K evidences, $e_k(x) = j_k, k = 1, 2, \ldots, K$ with each evidence denoting that sample x is assigned a class label by classifier $e_K$.

We have prior knowledge of the recognition rate $\varepsilon_r^{(k)}$ and the substitution rate $\varepsilon_s^{(k)}$ of $e_K$. For each evidences produced $e_k(x) = j_k$, one could infer uncertain beliefs that the proposition $A_{jk} = x \in C_{jk}$ is true with a degree $\varepsilon_r^{(k)}$ (recognition rate) and is not true with a degree $\varepsilon_s^{(k)}$. If $x$ is rejected by $e_K$ i.e. when $j_k = M + 1$, it has no idea of the any given propositions, and it infers the full support of the universal proposition $\Theta$. Now we can define a Basic Probability Assignment function $m_k$ on the universal proposition $\Theta$ for evidence in the given way:

- $m_k$ has only one focal element $\Theta$ , when $j_k = M + 1$ , with $m_k(\Theta) = 1$ .This is a degenerate case as the evidence $e_K$ says nothing about the any of the M propositions.

- If $j_k \in \Lambda$, only two focal elements are there in $m_k$ , namely $A_{jk}$ and $\neg A_{jk} = \Theta - \{A_{jk}\}$ . Now we have $m_k(A_{jk}) = \varepsilon_r^{(k)}$ and $m_k(\neg A_{jk}) = \varepsilon_s^{(k)}$. As $e_K$ says nothing about other propositions, we have $m_k(\Theta) = 1 - \varepsilon_r^{(k)} - \varepsilon_s^{(k)}$ . As we have K evidences, we will obtain K basic probability assignment functions $m_k, k = 1..., K$ . Now as we have formulated the problem in a usable format, the next step is to use the D-S rule to obtain a combined BPA $m = m_1 \oplus m_2 \oplus m_3 \oplus ....... \oplus m_K$ and use this new BPA to calculate the belief functions $bel(A_i)$ and $bel(\neg A_i)$ based on the K evidences we gathered. After this the combined classifier could be formed by using the decision rules derived from these beliefs.

### 2.6.3  Conclusion of D-S Theory

Now having concluded the above section, we have seen three approaches namely Bayesian Formalism, Voting principle and Dempster Shafer theory. Let us summarize the comparative advantages/disadvantages of each over one another.

- We found out that the D-S approach is quite robust. Inaccurate learning doesn't affect the performance much [9].

- If the confusion matrix of each algorithm is well learned, then Bayesian approach is the best method. Although it is unstable. Rough learning will degrade the performance quickly [9].

- The D-S approach is better than the voting approach when high reliability is required [9].

## 2.7 Some fixed rules of combination [8]:

There are broadly two combination scenarios. In the first scenario, the classifiers use the same representation of the input pattern. An example being a set of k-NN classifiers, each using the same measurement vectors, but classifier parameters are different (k varies, distance metrics varies). When given an input pattern, this produces an estimate of the same a posterior class probability. In the second scenario, the classifiers use their own representation of the input pattern. The measurements extracted by each classifier from the pattern are unique to each classifier. The combination of this type requires to physically integrate the different types of measurement/features as the computed a posterior probabilities are not the estimate of the same functional values. Kittler et al(1998), devise a framework which under some assumptions can lead to a proper formulation of some commonly used combining strategy.

### 2.7.1 Theoretical framework

Let us consider a pattern recognition problem, in which pattern Z is to be assigned to one of the possible classes $(w_1, \ldots, w_m)$. Also we have R classifiers each one representing the pattern with a unique measurement vector. The measurement vector used by the *ith* classifier is denoted by $x_i$. The class $w_k$ can be modelled by the PDF $p(x_i/w_k)$ and its prior probability of occurrence can be denoted as $p(w_k)$. The models are assumed to be mutually exclusive, i.e. each pattern is associated with only one model. Now according to the Bayesian theory,

$$p(w_k/x_1, \ldots, x_R) = \frac{p(x_1, \ldots, x_R \mid w_k)P(w_k)}{p(x_1, \ldots, x_R)}$$

(2.13)

Where,

$$p(x_1,......,x_R) = \sum_{j=1}^{m} p(x_1,......,x_R \mid w_j) P(w_j)$$

(2.14)

### 2.7.2  Product Rule

The basic assumption made to arrive at the product rule is that the representations used are conditionally independent. This consequence can be formulated as,

$$p(x_1,......,x_R \mid w_k) = \prod_{i=1}^{R} p(x_i \mid w_k)$$

(2.15)

Now substituting the above equation in the previously obtained Bayes equation, we find

$$P(w_k \mid x_1,......,x_R) = \frac{P(w_k) \prod_{i=1}^{R} p(x_i \mid w_k)}{\sum_{j=1}^{m} P(w_j) \prod_{i=1}^{R} p(x_i \mid w_j)}$$

(2.16)

And the decision rule formulated is,

Assign    Z $\rightarrow$ $w_j$  *if*

$$P(w_j) \prod_{i=1}^{R} p(x_i \mid w_j) = \max_{k=1}^{m} P(w_k) \prod_{i=1}^{R} P(x_i \mid w_k)$$

(2.17)

Or in terms of the a posterior probabilities obtained by the respective classifiers

Assign    Z $\rightarrow$ $w_j$  *if*

$$P^{-(R-1)}(w_j) \prod_{i=1}^{R} P(w_j \mid x_i) = \max_{k=1}^{m} P^{-(R-1)}(w_k) \prod_{i=1}^{R} P(w_k \mid x_i)$$

(2.18)

### 2.7.3  Sum Rule

To derive at the sum rule a strong assumption is made. It is assumed that the a posterior probability computed by the respective classifiers will not deviate drastically much from the prior probabilities. The sum decision rule can be denoted as, *assign* $\quad$ Z $\rightarrow W_j \quad$ *if*

$$(1-R)P(W_j) + \sum_{i=1}^{R} P(W_j \mid x_i) = \max_{k=1}^{m} \left[ (1-R)P(W_k) + \sum_{i=1}^{R} P(W_k \mid x_i) \right]$$

$$(2.19)$$

Due to the assumption made, if the patterns convey discriminating information, the sum rule will introduce gross approximation error.

The decision rules ( ) and ( ) lays the foundation for classifier combination. Other popular combination schemes can be derived from these rules by noting that,

$$\prod_{i=1}^{R} P(W_k \mid x_i) \leq \min_{i=1}^{R} P(W_k \mid x_i) \leq \frac{1}{R} \sum_{i=1}^{R} P(W_k \mid x_i) \leq \max_{i=1}^{R} P(W_k \mid x_i)$$

$$(2.20)$$

The above relationship suggests that the sum and the product combination rules can be approximated by the above upper and lower bounds. The a posterior probabilities $P(W_k \mid x_i)$ can be hardened to produce binary valued function $\Delta_{ki}$ as

$$\Delta_{ki} = \begin{cases} 1, if \dots\dots P(W_k \mid x_i) = \max_{j=1}^{m} P(W_j \mid x_i) \\ 0, \dots\dots Otherwise \end{cases}$$

$$(2.21)$$

The above function, instead of combining a posteriori probabilities, combines decision outcomes. The above approximations will lead to the following rules.

### 2.7.4 Max Rule

Using (2.20) and approximating the sum using the maximum of the posterior probabilities,

*assign* $\quad$ Z $\rightarrow W_j \quad$ *if*

$$(1-R)P(W_j) + R \max_{i=1}^{R} P(W_j \mid x_i) = \max_{i=1}^{R} \left[ (1-R)P(W_k) + R \max_{i=1}^{R} P(W_k \mid x_i) \right]$$

$$(2.22)$$

Assuming the priors are equal, it simplifies to,

*assign* $\quad Z \rightarrow w_j \quad if$

$$\max_{i=1}^{R} P(w_j \mid x_i) = \max_{k=1}^{m} \max_{i=1}^{R} P(w_k \mid x_i)$$

(2.23)

### 2.7.5  Min Rule

Using (2.18) and bounding the product of the posterior probabilities we obtain,

*assign* $\quad Z \rightarrow w_j \quad if$

$$P^{-(R-1)}(w_j) \min_{i=1}^{R} P(w_j \mid x_i) = \max_{k=1}^{m} P^{-(R-1)}(w_k) \min_{i=1}^{R} P(w_k \mid x_i)$$

(2.24)

Assuming the priors are equal, it simplifies to

*assign* $\quad Z \rightarrow w_j \quad if$

$$\min_{i=1}^{R} P(w_j \mid x_i) = \max_{k=1}^{m} \min_{i=1}^{R} P(w_k \mid x_i)$$

(2.25)

### 2.7.6  Mean Rule

The sum rule in (2.20), under the equal prior assumption can be viewed to be computing the average a posterior probability over all the classifier outputs for each class, i.e.

*assign* $\quad Z \rightarrow w_j \quad if$

$$\frac{1}{R} \sum_{i=1}^{R} P(w_j \mid x_i) = \max_{k=1}^{m} \frac{1}{R} \sum_{i=1}^{R} P(w_k \mid x_i)$$

(2.26)

Thus the class with the maximum average a posterior probability get assigned to the pattern.

### 2.7.7  Median Rule

Taking a clue from the above rule, in case of an outlier, the average posterior probability will be affected, which in turn will lead to an incorrect decision. Also a popularly known fact is that the median is a better or robust estimate of the mean. So it could be wise to model the combined decision on the median of the a posterior probabilities.

*assign*     $Z \rightarrow w_j$     *if*

$$\operatorname*{med}_{i=1}^{R} P\left(w_j \mid x_i\right) = \max_{k=1}^{m} \operatorname*{med}_{i=1}^{R} P\left(w_k \mid x_i\right)$$

(2.27)

# *Chapter 3*

## *Proposed Methodology*

In the earlier work done in the combination approaches, majority of the researchers have played with weak classifiers. Common approach to form a weak classifier is by tweaking the training data or by using different sampling techniques and feeding it to a learning algorithm. Bagging and Boosting as discussed above are the examples of combining weak classifiers. In my research, I have experimented with the combination of few strong classifiers. In all, three classifiers are build, namely nearest neighbour on raw image pixels, Nearest neighbour on a structural feature extracted image, and Nearest neighbour on GABOR feature extracted image. PCA is also used after the GABOR feature extraction process due to the high dimensionality and redundancy of the feature vector. The bank of GABOR filters used is formed by using different orientation, kernel size and frequency of the filter. After the classifiers are formed having their individual accuracy, now each classifier forms a hypothesis. A hypothesis is a proposed explanation for a phenomenon. Each classifier gives its own hypothesis in the form of a confusion matrix. My task is to use the relevant information from each classifier and suggest a suitable framework and technique to combine all the classifiers in such a way, which would take in to account the positive discriminating features of each classifier and help to increase the overall accuracy of the ensemble better than what is obtained by the individual classifiers. A confidence chart is prepared using the confusion matrix or test sample accuracy which gives the confidence about the classifiers on each class labels or on particular samples of data. Formation techniques of the confidence chart lay the foundation of this whole research. There can be many techniques to form the confidence chart (apart from using the information from the confusion matrix). Based on the technique, corresponding hypotheses are proposed for the ensemble. In my research, I have used 5 different hypotheses based on different ways of creating the confidence chart. To combine the three classifiers, maximum rule and weighted majority voting rules are used and analysed.

Also the basic assumption still holds true, which is about the conditional independence of the data i.e. the classifiers, will not make the same mistake. But there is no guarantee to it. Although one can't say for sure, how much the accuracy will increase, if at all it increases due to the different ways of collecting the data and percolation of noise into it. But one thing that can be said about the combination approaches are that, this will stabilise the accuracy to a certain point, i.e individual accuracies might change when the data samples change, but an effective combination strategy will ensure that the overall accuracy of the ensemble doesn't change drastically due to the dynamic nature of the data. This is one of the biggest advantages of using an ensemble.

Before we dive deep, let us first have a look at the data set.

## 3.1 Dataset

The dataset comprises of the Tibetan characters. Total 232 unique indexed class labels are there. Many class's data are missing though. Some class's have less than 5 samples while some have more than 200. So the distribution of the samples among the classes are not uniform. 4 different books, namely *"drugpa kunleg", "Losang gongyen"," milai gurbam", "Nyang choejung"* are scanned and characters are extracted and stored. Each book has its own set of fonts. Each books have many classes (not necessarily all the classes) as encountered in the text. The following is the distribution of the data among the number of classes.

| INDEX | IMAGES | | INDEX | IMAGES | | INDEX | IMAGES | | INDEX | IMAGES |
|---|---|---|---|---|---|---|---|---|---|---|
| 0 | 0 | | 31 | 0 | | 62 | 0 | | 93 | 2 |
| 1 | 0 | | 32 | 117 | | 63 | 43 | | 94 | 132 |
| 2 | 0 | | 33 | 101 | | 64 | 0 | | 95 | 103 |
| 3 | 0 | | 34 | 57 | | 65 | 0 | | 96 | 26 |
| 4 | 0 | | 35 | 1 | | 66 | 0 | | 97 | 0 |
| 5 | 0 | | 36 | 21 | | 67 | 44 | | 98 | 0 |
| 6 | 0 | | 37 | 100 | | 68 | 0 | | 99 | 13 |
| 7 | 0 | | 38 | 7 | | 69 | 2 | | 100 | 8 |
| 8 | 0 | | 39 | 5 | | 70 | 7 | | 101 | 0 |

| | | | | | | | |
|---|---|---|---|---|---|---|---|
| 9 | 0 | 40 | 17 | 71 | 70 | 102 | 0 |
| 10 | 0 | 41 | 133 | 72 | 66 | 103 | 0 |
| 11 | 0 | 42 | 0 | 73 | 0 | 104 | 130 |
| 12 | 0 | 43 | 89 | 74 | 53 | 105 | 0 |
| 13 | 0 | 44 | 0 | 75 | 228 | 106 | 0 |
| 14 | 0 | 45 | 0 | 76 | 0 | 107 | 109 |
| 15 | 0 | 46 | 0 | 77 | 0 | 108 | 0 |
| 16 | 0 | 47 | 0 | 78 | 2 | 109 | 0 |
| 17 | 0 | 48 | 0 | 79 | 0 | 110 | 0 |
| 18 | 0 | 49 | 0 | 80 | 0 | 111 | 95 |
| 19 | 0 | 50 | 0 | 81 | 44 | 112 | 0 |
| 20 | 103 | 51 | 26 | 82 | 99 | 113 | 0 |
| 21 | 109 | 52 | 133 | 83 | 1 | 114 | 0 |
| 22 | 104 | 53 | 49 | 84 | 0 | 115 | 94 |
| 23 | 55 | 54 | 0 | 85 | 49 | 116 | 0 |
| 24 | 0 | 55 | 98 | 86 | 2 | 117 | 0 |
| 25 | 0 | 56 | 0 | 87 | 26 | 118 | 7 |
| 26 | 0 | 57 | 47 | 88 | 1 | 119 | 0 |
| 27 | 0 | 58 | 108 | 89 | 120 | 120 | 0 |
| 28 | 0 | 59 | 46 | 90 | 5 | 121 | 0 |
| 29 | 0 | 60 | 0 | 91 | 0 | 122 | 0 |
| 30 | 0 | 61 | 9 | 92 | 40 | 123 | 0 |
| 124 | 15 | 156 | 34 | 188 | 0 | 220 | 148 |
| 125 | 0 | 157 | 0 | 189 | 0 | 221 | 117 |
| 126 | 0 | 158 | 41 | 190 | 0 | 222 | 234 |
| 127 | 0 | 159 | 77 | 191 | 0 | 223 | 10 |
| 128 | 0 | 160 | 0 | 192 | 0 | 224 | 147 |
| 129 | 0 | 161 | 13 | 193 | 0 | 225 | 0 |
| 130 | 0 | 162 | 0 | 194 | 0 | 226 | 0 |
| 131 | 0 | 163 | 0 | 195 | 0 | 227 | 0 |
| 132 | 0 | 164 | 2 | 196 | 0 | 228 | 0 |
| 133 | 0 | 165 | 0 | 197 | 0 | 229 | 0 |
| 134 | 0 | 166 | 105 | 198 | 0 | 230 | 0 |
| 135 | 210 | 167 | 6 | 199 | 0 | 231 | 245 |
| 136 | 0 | 168 | 0 | 200 | 0 | | |

| 137 | 0 | | 169 | 2 | | 201 | 124 | | | |
| 138 | 6 | | 170 | 118 | | 202 | 103 | | | |
| 139 | 0 | | 171 | 112 | | 203 | 8 | | | |
| 140 | 0 | | 172 | 39 | | 204 | 7 | | | |
| 141 | 59 | | 173 | 0 | | 205 | 0 | | | |
| 142 | 61 | | 174 | 6 | | 206 | 1 | | | |
| 143 | 125 | | 175 | 247 | | 207 | 3 | | | |
| 144 | 9 | | 176 | 83 | | 208 | 5 | | | |
| 145 | 83 | | 177 | 0 | | 209 | 3 | | | |
| 146 | 0 | | 178 | 32 | | 210 | 7 | | | |
| 147 | 0 | | 179 | 0 | | 211 | 3 | | | |
| 148 | 86 | | 180 | 25 | | 212 | 2 | | | |
| 149 | 17 | | 181 | 99 | | 213 | 0 | | | |
| 150 | 0 | | 182 | 0 | | 214 | 0 | | | |
| 151 | 0 | | 183 | 13 | | 215 | 0 | | | |
| 152 | 51 | | 184 | 20 | | 216 | 0 | | | |
| 153 | 0 | | 185 | 7 | | 217 | 95 | | | |
| 154 | 21 | | 186 | 0 | | 218 | 8 | | | |
| 155 | 1 | | 187 | 0 | | 219 | 84 | | | |

**Table 3.1: Overview of the data distribution**

The above data gives the rough idea about how the samples are distributed among the various classes. Now, having seen the distribution of data among various classes, we find that the data is not uniform. The number of images per class varies drastically from 0(lowest) to 245(highest). Thus to ensure uniformity, we consider only those classes which has at least 25 samples. The number of such classes is 64. We take 25 samples for each class. Out of these, 15 samples are for training, 5 samples are for testing and 5 samples are for validation. The number of samples for each class from each book is shown:

| CLASS INDEX | CLASS LABEL | BOOK #1 | BOOK #2 | BOOK #3 | BOOK #4 |
|---|---|---|---|---|---|
| 0 | 20 | 103 | 0 | 0 | 0 |
| 1 | 21 | 109 | 0 | 0 | 0 |
| 2 | 22 | 104 | 0 | 0 | 0 |
| 3 | 23 | 53 | 0 | 2 | 0 |
| 4 | 32 | 117 | 0 | 0 | 0 |
| 5 | 33 | 101 | 0 | 0 | 0 |
| 6 | 34 | 30 | 20 | 7 | 0 |
| 7 | 37 | 100 | 0 | 0 | 0 |
| 8 | 41 | 18 | 4 | 1 | 110 |
| 9 | 43 | 14 | 19 | 8 | 48 |
| 10 | 51 | 14 | 7 | 2 | 3 |
| 11 | 52 | 59 | 11 | 39 | 24 |
| 12 | 53 | 49 | 0 | 0 | 0 |
| 13 | 55 | 92 | 6 | 0 | 0 |
| 14 | 57 | 4 | 18 | 24 | 1 |
| 15 | 58 | 108 | 0 | 0 | 0 |
| 16 | 59 | 46 | 0 | 0 | 0 |
| 17 | 63 | 12 | 7 | 0 | 24 |
| 18 | 67 | 35 | 1 | 2 | 6 |
| 19 | 71 | 70 | 0 | 0 | 0 |
| 20 | 72 | 66 | 0 | 0 | 0 |
| 21 | 74 | 53 | 0 | 0 | 0 |
| 22 | 75 | 228 | 0 | 0 | 0 |
| 23 | 81 | 1 | 5 | 0 | 38 |
| 24 | 82 | 98 | 1 | 0 | 0 |
| 25 | 85 | 14 | 19 | 4 | 12 |
| 26 | 87 | 4 | 0 | 0 | 22 |
| 27 | 89 | 88 | 3 | 16 | 13 |
| 28 | 92 | 29 | 2 | 4 | 5 |
| 29 | 94 | 58 | 74 | 0 | 0 |
| 30 | 95 | 103 | 0 | 0 | 0 |
| 31 | 96 | 24 | 2 | 0 | 0 |
| 32 | 104 | 130 | 0 | 0 | 0 |
| 33 | 107 | 109 | 0 | 0 | 0 |

| 34 | 111 | 77  | 18  | 0   | 0   |
|----|-----|-----|-----|-----|-----|
| 35 | 115 | 0   | 0   | 0   | 94  |
| 36 | 135 | 210 | 0   | 0   | 0   |
| 37 | 141 | 43  | 14  | 2   | 0   |
| 38 | 142 | 43  | 7   | 11  | 0   |
| 39 | 143 | 23  | 102 | 0   | 0   |
| 40 | 145 | 83  | 0   | 0   | 0   |
| 41 | 148 | 86  | 0   | 0   | 0   |
| 42 | 152 | 51  | 0   | 0   | 0   |
| 43 | 156 | 16  | 4   | 0   | 14  |
| 44 | 158 | 40  | 1   | 0   | 0   |
| 45 | 159 | 77  | 0   | 0   | 0   |
| 46 | 166 | 105 | 0   | 0   | 0   |
| 47 | 170 | 78  | 8   | 5   | 27  |
| 48 | 171 | 109 | 0   | 0   | 3   |
| 49 | 172 | 36  | 2   | 1   | 0   |
| 50 | 175 | 68  | 175 | 0   | 4   |
| 51 | 176 | 83  | 0   | 0   | 0   |
| 52 | 178 | 26  | 1   | 3   | 2   |
| 53 | 180 | 21  | 0   | 1   | 3   |
| 54 | 181 | 90  | 4   | 0   | 5   |
| 55 | 201 | 34  | 13  | 59  | 18  |
| 56 | 202 | 103 | 0   | 0   | 0   |
| 57 | 217 | 65  | 30  | 0   | 0   |
| 58 | 219 | 70  | 1   | 7   | 6   |
| 59 | 220 | 19  | 0   | 6   | 123 |
| 60 | 221 | 0   | 0   | 0   | 117 |
| 61 | 222 | 232 | 2   | 0   | 0   |
| 62 | 224 | 89  | 36  | 9   | 13  |
| 63 | 231 | 0   | 0   | 245 | 0   |

**Table 3.2: Distribution of data for classes with samples greater than 25.**

Now, having seen the distribution of the data above, the next question arise is *"how many samples for each class from each book should be collected? "* We have to ensure uniformity in all respect. Equal number of samples should ideally be collected from each book. A logic is written which will determine the number of samples to be taken from each book for that class, given the distribution of data. The algorithm is,

*1) Check for the book with the minimum sample.*

*2) If ( Number of samples < 6 ) ; Take all the samples from that book.*

*3) Else ; Take only 6 samples from that book.*

*4) IF ( Last book? ), GOTO Step 6.*

*5) Go to the book with next minimum sample and repeat steps 2 &3.*

*6) Sum all the samples taken so far (sum) and take 25 - sum number of samples from the last book.*

Thus we have got a somewhat equally distributed samples from each book. Now, the next dilemma is - *"Which samples to collect from each class of each book, given the number of samples to be collected from each book for each class?"*

The best bet would be to use sampling theory, which would ensure good samples, which represents the whole class distribution. But to keep things simple, a random function is used, which will collect the required number of samples randomly from each book for each class.

After doing this the data distribution looks something like this:

| CLASS INDEX | CLASS LABEL | BOOK #1 | BOOK #2 | BOOK #3 | BOOK #4 |
|---|---|---|---|---|---|
| 0 | 20 | 25 | 0 | 0 | 0 |
| 1 | 21 | 25 | 0 | 0 | 0 |
| 2 | 22 | 25 | 0 | 0 | 0 |
| 3 | 23 | 23 | 0 | 2 | 0 |
| 4 | 32 | 25 | 0 | 0 | 0 |
| 5 | 33 | 25 | 0 | 0 | 0 |
| 6 | 34 | 13 | 6 | 6 | 0 |
| 7 | 37 | 25 | 0 | 0 | 0 |
| 8 | 41 | 6 | 4 | 1 | 14 |
| 9 | 43 | 6 | 6 | 6 | 7 |

| 10 | 51 | 14 | 6 | 2 | 3 |
|----|-----|----|----|----|----|
| 11 | 52 | 7 | 6 | 6 | 6 |
| 12 | 53 | 25 | 0 | 0 | 0 |
| 13 | 55 | 19 | 6 | 0 | 0 |
| 14 | 57 | 4 | 6 | 14 | 1 |
| 15 | 58 | 25 | 0 | 0 | 0 |
| 16 | 59 | 25 | 0 | 0 | 0 |
| 17 | 63 | 6 | 6 | 0 | 13 |
| 18 | 67 | 16 | 1 | 2 | 6 |
| 19 | 71 | 25 | 0 | 0 | 0 |
| 20 | 72 | 25 | 0 | 0 | 0 |
| 21 | 74 | 25 | 0 | 0 | 0 |
| 22 | 75 | 25 | 0 | 0 | 0 |
| 23 | 81 | 1 | 5 | 0 | 19 |
| 24 | 82 | 24 | 1 | 0 | 0 |
| 25 | 85 | 6 | 9 | 4 | 6 |
| 26 | 87 | 4 | 0 | 0 | 21 |
| 27 | 89 | 10 | 3 | 6 | 6 |
| 28 | 92 | 14 | 2 | 4 | 5 |
| 29 | 94 | 6 | 19 | 0 | 0 |
| 30 | 95 | 25 | 0 | 0 | 0 |
| 31 | 96 | 23 | 2 | 0 | 0 |
| 32 | 104 | 25 | 0 | 0 | 0 |
| 33 | 107 | 25 | 0 | 0 | 0 |
| 34 | 111 | 19 | 6 | 0 | 0 |
| 35 | 115 | 0 | 0 | 0 | 25 |
| 36 | 135 | 25 | 0 | 0 | 0 |
| 37 | 141 | 17 | 6 | 2 | 0 |
| 38 | 142 | 13 | 6 | 6 | 0 |
| 39 | 143 | 6 | 19 | 0 | 0 |
| 40 | 145 | 25 | 0 | 0 | 0 |
| 41 | 148 | 25 | 0 | 0 | 0 |
| 42 | 152 | 25 | 0 | 0 | 0 |
| **43** | 156 | 15 | 4 | 0 | 6 |
| **44** | 158 | 24 | 1 | 0 | 0 |
| **45** | 159 | 25 | 0 | 0 | 0 |
| **46** | 166 | 25 | 0 | 0 | 0 |
| **47** | 170 | 8 | 6 | 5 | 6 |
| **48** | 171 | 22 | 0 | 0 | 3 |
| **49** | 172 | 22 | 2 | 1 | 0 |
| **50** | 175 | 6 | 15 | 0 | 4 |
| **51** | 176 | 25 | 0 | 0 | 0 |
| **52** | 178 | 19 | 1 | 3 | 2 |
| **53** | 180 | 21 | 0 | 1 | 3 |
| **54** | 181 | 16 | 4 | 0 | 5 |
| **55** | 201 | 6 | 6 | 7 | 6 |
| **56** | 202 | 25 | 0 | 0 | 0 |
| **57** | 217 | 19 | 6 | 0 | 0 |
| **58** | 219 | 12 | 1 | 6 | 6 |
| **59** | 220 | 6 | 0 | 6 | 13 |
| **60** | 221 | 0 | 0 | 0 | 25 |
| **61** | 222 | 23 | 2 | 0 | 0 |
| **62** | 224 | 7 | 6 | 6 | 6 |
| **63** | 231 | 0 | 0 | 25 | 0 |

**Table 3.3: Distribution of samples selected from different books for each class.**

Now let's take a look at the actual samples for each class. How do they actually look?

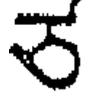

CLASS #20

CLASS INDEX  #0

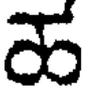

CLASS #21

CLASS INDEX  #1

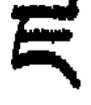

CLASS #22

CLASS INDEX  #2

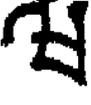

CLASS #23

CLASS INDEX  #3

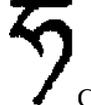

CLASS #32

CLASS INDEX  #4

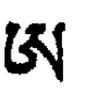

CLASS #33

CLASS INDEX  #5

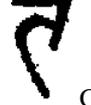

CLASS #34

CLASS INDEX  #6

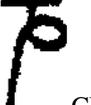

CLASS #37

CLASS INDEX  #7

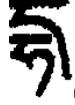 CLASS #41

CLASS INDEX #8

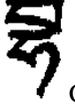 CLASS #43

CLASS INDEX #9

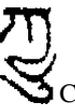 CLASS #51

CLASS INDEX #10

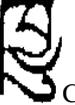 CLASS #52

CLASS INDEX #11

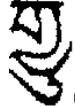 CLASS #53

CLASS INDEX #12

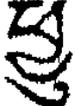 CLASS #55

CLASS INDEX #13

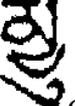 CLASS #57

CLASS INDEX #14

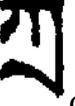 CLASS #58

CLASS INDEX #15

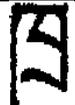 CLASS #59

CLASS INDEX #16

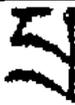 CLASS #63

CLASS INDEX #17

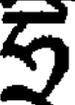 CLASS #67

CLASS INDEX #18

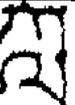 CLASS #71

CLASS INDEX #19

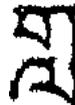 CLASS #72

CLASS INDEX #20

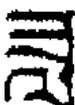 CLASS #74

CLASS INDEX #21

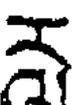 CLASS #75

CLASS INDEX #22

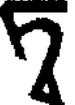 CLASS #81

CLASS INDEX #23

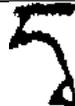 CLASS #82

CLASS INDEX #24

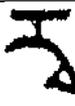 CLASS #85

CLASS INDEX #25

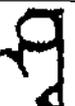 CLASS #87

CLASS INDEX #26

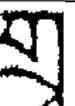 CLASS #89

CLASS INDEX #27

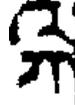 CLASS #92

CLASS INDEX #28

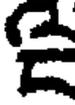 CLASS #94

CLASS INDEX #29

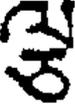 CLASS #95

CLASS INDEX #30

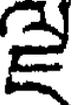 CLASS #96

CLASS INDEX #31

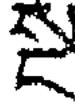 CLASS #104

CLASS INDEX #32

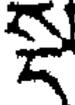 CLASS #107

CLASS INDEX #33

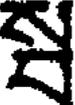 CLASS #111

CLASS INDEX #34

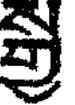 CLASS #115

CLASS INDEX #35

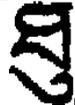 CLASS #135

CLASS INDEX #36

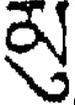 CLASS #141

CLASS INDEX #37

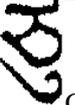 CLASS #142

CLASS INDEX #38

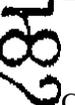 CLASS #143

CLASS INDEX #39

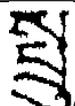 CLASS #145

CLASS INDEX #40

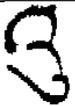 CLASS #148

CLASS INDEX #41

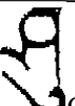 CLASS #152

CLASS INDEX #42

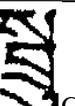 CLASS #156

CLASS INDEX #43

| | | | |
|---|---|---|---|
| 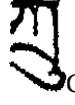CLASS #158<br>CLASS INDEX #44 | 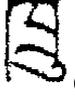CLASS #159<br>CLASS INDEX #45 | 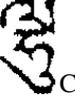CLASS #166<br>CLASS INDEX #46 | 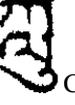CLASS #170<br>CLASS INDEX #47 |
| 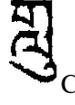CLASS #171<br>CLASS INDEX #48 | 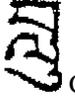CLASS #172<br>CLASS INDEX #49 | 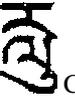CLASS #175<br>CLASS INDEX #50 | 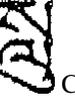CLASS #176<br>CLASS INDEX #51 |
| 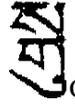CLASS #178<br>CLASS INDEX #52 | 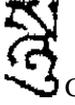CLASS #180<br>CLASS INDEX #53 | 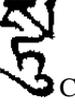CLASS #181<br>CLASS INDEX #54 | 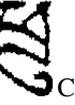CLASS #201<br>CLASS INDEX #55 |
| 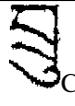CLASS #202<br>CLASS INDEX #56 | 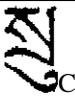CLASS #217<br>CLASS INDEX #57 | 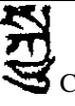CLASS #219<br>CLASS INDEX #58 | 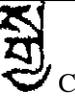CLASS #220<br>CLASS INDEX #59 |
| 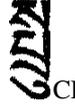CLASS #221<br>CLASS INDEX #60 | 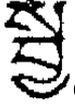CLASS #222<br>CLASS INDEX #61 | 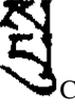CLASS #224<br>CLASS INDEX #62 | CLASS #231<br>CLASS INDEX #63 |

**Table 3.4: Visual appearance of the dataset**

Now, having known the dataset, we move forward to the formation of the individual classifiers. Also one more important factor to be noted, due to the different sizes of the images collected, all the images are first resized to a fixed dimension of 32x32. And to remove any unwanted grey values in the images, a threshold is applied to completely binarize the images.

## 3.2 Software and source code used

- The language used to code is C++.
- Library used is OPEN-CV.

**Classifier-1:** Classifier-1 is formed by simply using a nearest neighbour approach on the raw pixel values. The images are read, and stored as column vectors in three matrices, namely training, testing and validation. The metrics used for measurement is Euclidean distance. This is the cheapest algorithm used among the three. The result of this classifier is represented in the form of a confusion matrix.

**Classifier-2:** Classifier-2 is a tricky one. In this classifier, instead of directly using the raw pixels as feature vector, we extract feature explicitly from the images. Next query would be *what is the feature vector?* To extract the feature from a given image, we first have to find the centroid of the image. Centroid can be found using a weighted mean approach in both horizontal and vertical direction and arriving at a particular value of *x* and *y* which becomes the centroid. Now, from the centroid, a vector is drawn up to the edge of the image vertically, and rotated a full 360 degrees. While doing so, the distance between the centroids and the last black pixel encountered on that vector, while moving from the centroids towards the edge of the image is noted down. This distance is noted down for all the vectors in the entire span of the image. Thus from an image, we obtain a feature vector of all the distances from the centroids to the edge points.

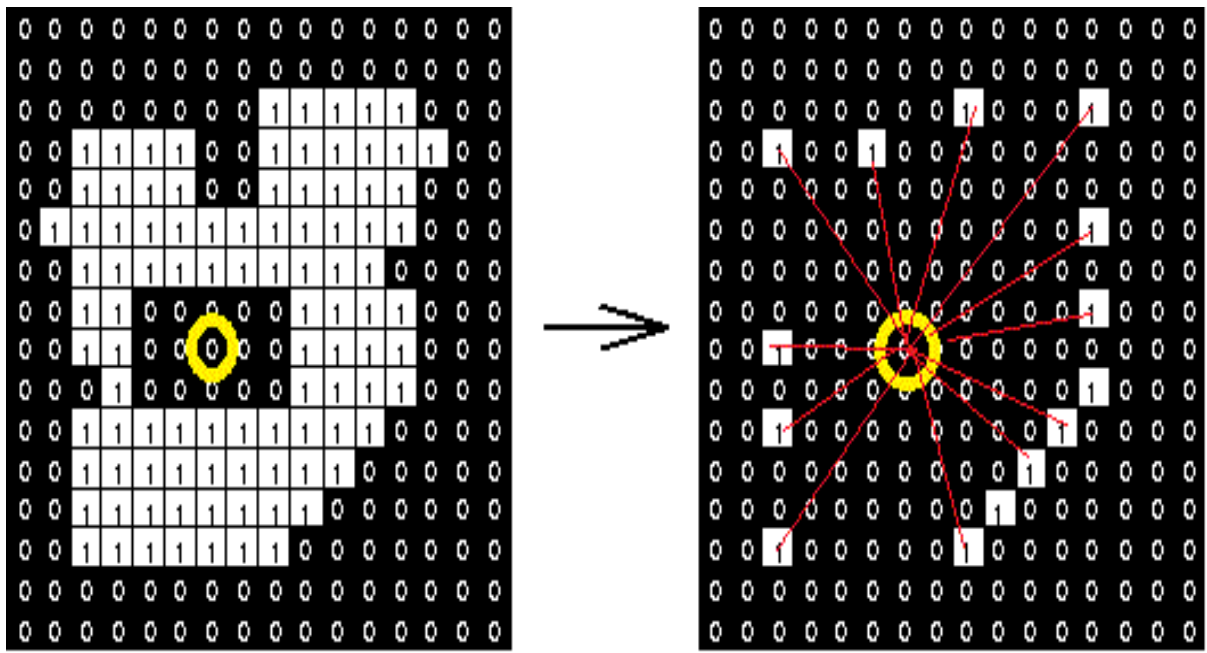

**Fig 3.1: Calculating the distance of the last black pixel along the vector drawn from the centroids to the perimeter of the image.**

## 3.3 Scientific tools used

### 3.3.1 Bressenham's Line drawing Algorithm

To draw a line in the image to connect two points, we used Bressenham's Line drawing algorithm [12] to determine which pixels are activated on that particular line. The algorithm takes input the starting and the ending points and gives out the corresponding pixels illuminated. One practical problem faced is that, due to the discrete nature of the images, the vector can't be rotated 360 degrees. A simple explanation for this is even if we consider a vector from the centroids to each and every pixels on the circumference of the image, then also for a 32x32 image, the maximum number of such vectors could be 32x4 i.e. 128. So the size of our feature vector is 128. After the features have been extracted, again a nearest neighbour algorithm is applied and the corresponding confusion matrix is formed.

Before we proceed further, let me briefly introduce GABOR filters and PCA which will be used in the next classifier.

### 3.3.2 Gabor Filter [13] [14] [15]:

Its a linear filter used for edge detection and named after Denis Gabor. The representation of frequency and orientation of Gabor filters are quite similar to those of human visual systems. They have found to be quite useful for texture representation and discrimination. It has two components namely real and complex which are orthogonal to each other. A set of gabor filters with different frequency and orientation helps to extract relevant features from an image. It is formed by multiplication of a sinusoidal waveform with a Gaussian. Due to this, it has a unique shape as shown in the next page. A Gabor filter is characterized by its frequency, standard deviation, mean, directionality, and aspect ratio. These are the important parameters which decide the final effect of the output after the filter is applied. It can also be thought of as a tuneable band pass filter. It is similar to windowed Fourier transform.

The basic equation of a 2-D Gabor filter is:

$$f\left(x,y,w,\theta,\sigma_x,\sigma_y\right) = \frac{1}{2\pi\sigma_x\sigma_y}\exp\left[\frac{-1}{2}\left(\left(\frac{x}{\sigma_x}\right)^2 + \left(\frac{y}{\sigma_y}\right)^2\right) + j\omega\left(x\cos\theta + y\sin\theta\right)\right] \quad (3.1)$$

Where,

$\sigma_x$ = Standard deviation along x-axis

$\sigma_y$ = Standard deviation along y-axis.

Θ = Orientation or the angle

ω = Frequency

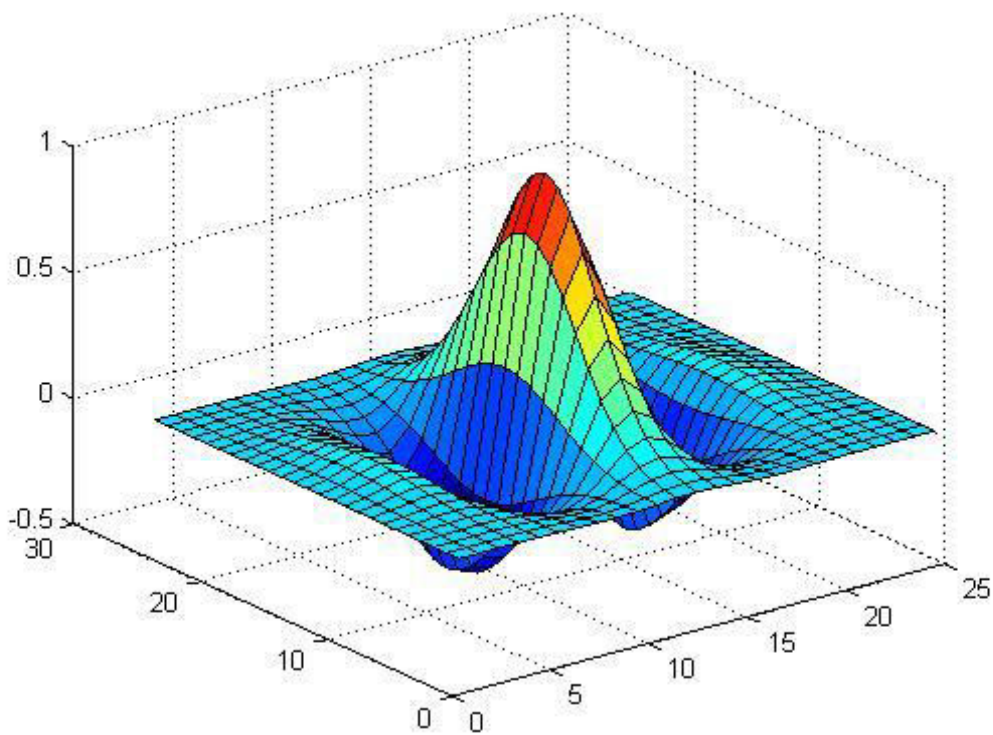

**Figure 3.2: A visual representation of a 2-D Gabor filter**

This is especially important for OCR applications due to its unique shape and flexibility. It gives varied response (weak/strong) depending on the orientation and frequency of the filter which makes it ideal for stroke detection applications. A typical example of the effect of a Gabor bank with different orientation on a Chinese character is shown below:

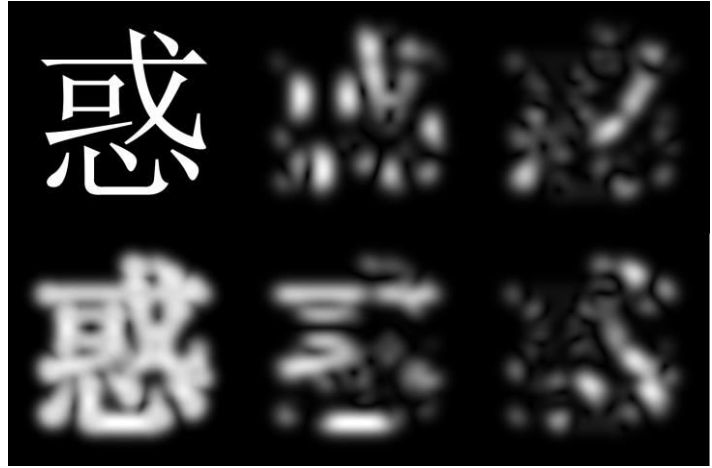

**Figure 3.3: Gabor Filter bank with four different orientations (0, 45, 90, 135) are shown on the right. The original is shown in the top left and below it is the superposition of all the Gabor images.**

Now after having a brief idea about GABOR filters, let's introduce our next topic, PCA (Principal component analysis).

### 3.3.3  Principal Component Analysis [16]

PCA is a dimension reduction technique and is used extensively in modern day to day applications. Due to the large size of our feature vectors and also due to the redundancy of information, PCA comes as a rescue to help distinguish relevant features which conveys relevant information from the irrelevant ones. The basic principle on which PCA works is that it tries to fit in a lower dimension hyper-plane, on which the projections of the data points would have maximum variance. In simple words, it tries to minimize the projection error. It sounds quite similar to regression, although there is a very simple difference between the two,

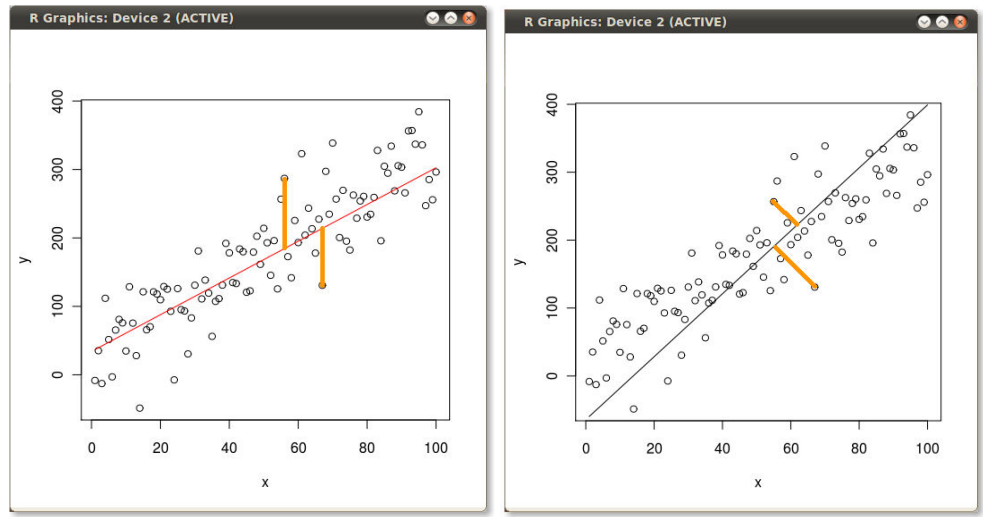

**Figure 3.4: Regression (left) v/s PCA (right).**

The Figure in the left shows regression whose objective is to minimise the output error. On the other hand PCA minimises the projection error. The basic steps to implement PCA are as follows:

1) Get some data
2) Subtract the mean
3) Calculate the Covariance matrix
4) Calculate the Eigen values and corresponding Eigen vectors of the covariance matrix.
5) Choosing the required number of components and forming the feature vector.

The last step is the deciding step in maintaining the required variance of the data. After computing the Eigen values of the covariance matrix, it is found that only few Eigen values are significant compared to the rest. It signifies that the data is distributed along that Eigen vector more whose values are higher. This will help to discriminate between different classes even if we discard the Eigen vectors where there is no visible variance, which leads to dimension reduction. Now we have the required tools ready to discuss the Classifier-3.

**Classifier-3:** This is the most decorated and expensive classifier used among the three. This classifier again extracts some features from the images. Feature extracted are by the means of a GABOR filter bank. The important point to be noted is how the Gabor filter has been used on the image? A GABOR kernel of size 5x5 is selected and convoluted on the whole image (size 32x32) starting from the top left each one with four different orientations (0, 45, 90,

135) which will generate a feature vector of size 4096. After this, we apply PCA to reduce the dimension from 4096 to 100 (optimum). After the feature vectors are formed, this is fed to our plain old nearest neighbour algorithm and the corresponding confusion matrix is created.

Now having done with the formation of the classifiers, we now proceed to the formation of the hypotheses. Five different hypotheses are formed. The description of each one of them is given below:

## 3.4   HYPOTHESIS #1

The confusion matrix is formed for each classifier. The confusion matrix is a square matrix (no. of rows = no. of class labels) which gives a general idea of which samples are classified in to which class. Using the confusion matrix, we create the confidence matrix which shows the confidence of each classifier for a given class label. Confidence matrix is created by taking the ratio of the diagonal element and the sum of that particular column and doing this for all the class labels. After this is done, next when a validation data is entered, all the classifiers predict its class label for that particular sample. Now we can use the confidence matrix as a lookup table and compare the confidence of the given classifiers on the predicted class labels. In case two or more classifiers predict the same class label then the confidence for that predicted class adds up. Finally to take a final decision, the hypothesis checks which class label has got the highest confidence value and that class label is selected as the final predicted class label of the hypothesis.

## 3.5   HYPOTHESIS #2

The basic strategy of this hypothesis is to exploit the distance between the best match and the second best match. More the distance of the best match and second best, more is the probability of that sample belonging to that particular class (best match). To explain in more detail, for a given test sample, the best match is found from the same class label from the training set and denoted as *dis1*. Then the same sample is found the second best match from the remaining training set (apart from the samples already scanned) and denoted as *dis2*. Now the ratio *dis2/dis1* gives a general idea about how far is the second best from the best match. More the ratio more the confidence. Now for every validation data that enters, we already have a set of built confidence values for each test sample from each classifier. Now, we just apply nearest neighbour on the validation set and form a result matrix which shows which classifiers predicted which class labels. After this, just compare the confidence obtained earlier for that sample number, and the classifier with the maximum confidence will have its final say in the voting.

## 3.6  HYPOTHESIS #3

This hypothesis takes in to account the centroids of each class labels. Each sample is compared to the class centroids, and the nearest match is noted. If the class label index matches with the test sample index, then we have a correct classification otherwise not. Using this information the confusion matrix is created for each classifier. Using the confusion matrix, confidence matrix is formed. Now, when a validation data comes, it is passed through the same procedure as the test data set has gone through. The classifiers will output a certain class label as its output. Now, the confidence matrix comes handy to decide which classifier has a higher confidence for that particular class label. Using weighted majority voting, the final decision of the hypothesis is decided.

## 3.7  HYPOTHESIS #4

This hypothesis takes off from the previous hypothesis. In this method, along with the centroids, the average distances of the samples from their respective centroids are calculated. The test samples are compared with the centroids and the minimum distance of that sample from all the centroids are found out. Now this distance is compared with all the "average distances for each class" that was previously calculated and the nearest match is found and the index is noted. This metric is used to form the confusion matrix. From the confusion matrix, confidence matrix is again created. Now for the validation data, the same procedure is repeated and the index is noted for the nearest match, which becomes the predicted class for that particular classifier. Now to arrive at a final decision, the confidence matrix is referred and based on the confidence of each classifier on particular predicted class labels, the final decision is made.

## 3.8 HYPOTHESIS #5

This hypothesis avoids the confusion matrix and directly calculates the confidence matrix using the ratio of average class distances and the nearest match from the centroids. If the nearest match is smaller, then the ratio will be higher indicating strong confidence of the classifier for that class label. Ideally the confidence of each classifier for each class should be close to 1.

The result and the analysis for each of the above hypotheses will be discussed in the next section.

# *Chapter 4*

## *Results and Analysis*

## 4.1   Hypothesis #1

### 4.1.1   Analysis of Classifier #1

**Figure 4.1: Confusion matrix of classifier-1(Nearest neighbour algorithm)**

**Conclusion:** Accuracy obtained is 84.06 %.

## 4.1.2   Analysis of Classifier #2

**Figure 4.2: Confusion matrix of classifier-2 (Centroid feature extraction method)**

**Conclusion:** Accuracy obtained is 79.06%.

### 4.1.3   Analysis of the PCA on accuracy of the Classifier #3

Before proceeding to the next classifier results (GABOR+PCA+Nearest neighbour), the analysis for the optimum GABOR filter and PCA combination is shown. The below table shows the comparison between the number of principal components retained and its effect on the overall accuracy of the classifier.

| Number of Principal components retained | Final accuracy of the classifier |
|:---:|:---:|
| 20 | 60.62 % |
| 30 | 63.75 % |
| 40 | 68.75 % |
| 50 | 69.37 % |
| 60 | 70.62 % |
| 70 | 71.56 % |
| 80 | 71.56 % |
| 90 | 72.81 % |
| 100 | 73.12 % |
| **110** | **74.37 %** |
| 120 | 74.37 % |
| 130 | 74.06 % |
| 140 | 74.06 % |

**Table 4.1: Number of components retained v/s classifier accuracy.**

**Conclusion:** From the above table, it is concluded that retaining 110 principal components gives the maximum accuracy. Also the accuracy decreases if the number of principal components is more than 120.

### 4.1.4 Effect of the size of GABOR kernel on the accuracy of Classifier #3

The effect of the size of the GABOR kernel on the accuracy of the classifier is shown below. The below is calculated keeping the number of principal components at 110.

| Size of the GABOR kernel | Accuracy |
|:---:|:---:|
| 3 | 64.68 % |
| 4 | 70.62 % |
| 5 | 70.62 % |
| 6 | 74.06 % |
| 7 | 74.37 % |
| 8 | 73.75 % |
| 9 | 73.75 % |
| 10 | 71.56 % |

**Table 4.2: Size of Gabor kernel v/s classifier accuracy**

**Conclusion:** From the above table we conclude that accuracy increases as the size of the kernel increases up to a certain point and then it starts to decrease. For the size of kernel being at 7, the maximum accuracy of the classifier is achieved at 74.37 %.

## 4.1.5   Analysis of classifier-3 (GABOR + PCA + Nearest Neighbour)

**Fig 4.3: Confusion matrix of classifier #3**

**Conclusion:** The accuracy obtained is 74.37 %.

## 4.1.6 Confidence Matrix

```
THE CONFIDENCE OF EACH CLASSIFIER ON INDIVIDUAL CLASSES

CLASS    CLASSIFIER-1    CLASSIFIER-2    CLASSIFIER-3

0        1.000000        1.000000        1.000000
1        1.000000        1.000000        1.000000
2        1.000000        1.000000        1.000000
3        1.000000        1.000000        1.000000
4        1.000000        1.000000        1.000000
5        1.000000        1.000000        1.000000
6        0.800000        0.800000        0.600000
7        1.000000        1.000000        1.000000
8        1.000000        1.000000        0.800000
9        0.400000        0.800000        0.200000
10       0.200000        0.400000        0.600000
11       0.600000        0.400000        1.000000
12       0.600000        0.600000        0.600000
13       0.800000        0.800000        0.800000
14       1.000000        0.800000        0.400000
15       1.000000        0.800000        1.000000
16       1.000000        1.000000        1.000000
17       0.800000        0.200000        0.600000
18       0.200000        0.200000        0.200000
19       1.000000        1.000000        1.000000
20       0.800000        0.800000        0.600000
21       1.000000        1.000000        1.000000
22       1.000000        1.000000        1.000000
23       1.000000        1.000000        1.000000
24       1.000000        1.000000        1.000000
25       1.000000        0.800000        0.600000
26       0.800000        0.600000        0.600000
27       1.000000        0.800000        0.800000
28       0.800000        1.000000        0.600000
29       1.000000        1.000000        1.000000
30       1.000000        1.000000        1.000000
31       1.000000        1.000000        1.000000
32       0.800000        1.000000        0.600000
33       1.000000        1.000000        1.000000
34       0.800000        1.000000        0.800000
35       1.000000        1.000000        1.000000
36       1.000000        1.000000        1.000000
37       0.400000        0.600000        0.400000
38       1.000000        0.600000        0.400000
39       1.000000        1.000000        1.000000
40       0.600000        0.600000        0.400000
41       1.000000        1.000000        1.000000
42       1.000000        1.000000        1.000000
43       0.000000        0.000000        0.200000
44       1.000000        1.000000        0.400000
45       1.000000        1.000000        1.000000
46       0.800000        0.400000        0.800000
47       0.800000        0.800000        0.000000
48       0.800000        0.600000        0.200000
49       1.000000        0.800000        1.000000
50       1.000000        1.000000        1.000000
51       1.000000        1.000000        1.000000
52       0.800000        0.600000        0.800000
53       0.600000        0.400000        0.600000
54       1.000000        1.000000        0.000000
55       0.800000        0.400000        1.000000
56       1.000000        0.600000        0.600000
57       0.800000        0.600000        0.400000
58       0.600000        0.400000        0.400000
59       0.000000        0.000000        0.000000
60       0.800000        0.800000        0.400000
61       1.000000        1.000000        1.000000
62       0.600000        0.600000        0.400000
63       1.000000        1.000000        1.000000
```

**Figure 4.4: Confidence matrix of the 3 classifiers**

## 4.1.7 Summary of Hypothesis #1

| SAMPLE NUMBER | CLASSIFIER-1 | CLASSIFIER-2 | CLASSIFIER-3 | ACTUAL-CLASS | PREDICTED CLASS |
|---|---|---|---|---|---|
| 0 | 0 | 0 | 0 | 0 | 0 |
| 1 | 0 | 0 | 0 | 0 | 0 |
| 2 | 0 | 0 | 0 | 0 | 0 |
| 3 | 0 | 0 | 0 | 0 | 0 |
| 4 | 0 | 0 | 0 | 0 | 0 |
| 5 | 1 | 1 | 1 | 1 | 1 |
| 6 | 1 | 1 | 1 | 1 | 1 |
| 7 | 1 | 1 | 1 | 1 | 1 |
| 8 | 1 | 1 | 1 | 1 | 1 |
| 9 | 1 | 1 | 1 | 1 | 1 |
| 10 | 2 | 2 | 2 | 2 | 2 |
| 11 | 2 | 2 | 2 | 2 | 2 |
| 12 | 2 | 2 | 2 | 2 | 2 |
| 13 | 2 | 2 | 2 | 2 | 2 |
| 14 | 2 | 2 | 2 | 2 | 2 |
| 15 | 3 | 3 | 3 | 3 | 3 |
| 16 | 3 | 3 | 3 | 3 | 3 |
| 17 | 3 | 3 | 3 | 3 | 3 |
| 18 | 3 | 27 | 3 | 3 | 3 |
| 19 | 3 | 3 | 3 | 3 | 3 |
| 20 | 25 | 4 | 25 | 4 | 25 |
| 21 | 4 | 4 | 4 | 4 | 4 |
| 22 | 4 | 4 | 4 | 4 | 4 |
| 23 | 4 | 4 | 4 | 4 | 4 |
| 24 | 4 | 4 | 4 | 4 | 4 |
| 25 | 5 | 5 | 5 | 5 | 5 |
| 26 | 5 | 5 | 5 | 5 | 5 |
| 27 | 5 | 5 | 5 | 5 | 5 |
| 28 | 5 | 5 | 5 | 5 | 5 |
| 29 | 5 | 5 | 5 | 5 | 5 |
| 30 | 6 | 9 | 6 | 6 | 6 |
| 31 | 6 | 63 | 6 | 6 | 63 |
| 32 | 6 | 9 | 6 | 6 | 6 |
| 33 | 6 | 6 | 6 | 6 | 6 |
| 34 | 6 | 6 | 6 | 6 | 6 |
| 35 | 7 | 7 | 7 | 7 | 7 |
| 36 | 7 | 7 | 7 | 7 | 7 |
| 37 | 7 | 7 | 7 | 7 | 7 |
| 38 | 7 | 7 | 7 | 7 | 7 |
| 39 | 7 | 7 | 7 | 7 | 7 |
| 40 | 8 | 8 | 8 | 8 | 8 |
| 41 | 9 | 8 | 8 | 8 | 8 |
| 42 | 9 | 8 | 8 | 8 | 8 |
| 43 | 8 | 9 | 8 | 8 | 8 |
| 44 | 63 | 34 | 25 | 8 | 63 |
| 45 | 9 | 9 | 3 | 9 | 9 |
| 46 | 8 | 9 | 8 | 9 | 8 |
| 47 | 9 | 9 | 4 | 9 | 9 |
| 48 | 9 | 9 | 9 | 9 | 9 |
| 49 | 2 | 9 | 63 | 9 | 63 |
| 50 | 10 | 13 | 10 | 10 | 10 |
| 51 | 10 | 56 | 12 | 10 | 56 |
| 52 | 10 | 10 | 63 | 10 | 63 |
| 53 | 10 | 10 | 63 | 10 | 63 |
| 54 | 47 | 56 | 12 | 10 | 47 |
| 55 | 11 | 58 | 11 | 11 | 11 |
| 56 | 11 | 10 | 11 | 11 | 11 |
| 57 | 11 | 52 | 10 | 11 | 52 |
| 58 | 11 | 11 | 63 | 11 | 11 |
| 59 | 56 | 56 | 37 | 11 | 56 |
| 60 | 12 | 12 | 12 | 12 | 12 |
| 61 | 12 | 12 | 12 | 12 | 12 |
| 62 | 12 | 10 | 12 | 12 | 12 |
| 63 | 12 | 10 | 12 | 12 | 12 |
| 64 | 12 | 10 | 12 | 12 | 12 |
| 65 | 14 | 14 | 14 | 13 | 14 |
| 66 | 14 | 14 | 63 | 13 | 14 |
| 67 | 14 | 10 | 10 | 13 | 10 |
| 68 | 14 | 10 | 10 | 13 | 10 |
| 69 | 14 | 10 | 10 | 13 | 10 |
| 70 | 14 | 14 | 14 | 14 | 14 |
| 71 | 14 | 14 | 14 | 14 | 14 |
| 72 | 14 | 14 | 14 | 14 | 14 |
| 73 | 14 | 14 | 14 | 14 | 14 |
| 74 | 35 | 30 | 63 | 14 | 63 |
| 75 | 15 | 63 | 15 | 15 | 15 |
| 76 | 34 | 34 | 34 | 15 | 34 |
| 77 | 15 | 15 | 15 | 15 | 15 |
| 78 | 15 | 21 | 15 | 15 | 15 |
| 79 | 15 | 15 | 15 | 15 | 15 |
| 80 | 16 | 16 | 16 | 16 | 16 |
| 81 | 16 | 16 | 16 | 16 | 16 |
| 82 | 16 | 16 | 16 | 16 | 16 |
| 83 | 16 | 16 | 16 | 16 | 16 |
| 84 | 16 | 16 | 16 | 16 | 16 |
| 85 | 17 | 15 | 19 | 17 | 19 |
| 86 | 17 | 9 | 17 | 17 | 17 |
| 87 | 15 | 20 | 35 | 17 | 35 |
| 88 | 17 | 17 | 63 | 17 | 17 |
| 89 | 17 | 19 | 63 | 17 | 63 |
| 90 | 25 | 8 | 25 | 18 | 25 |
| 91 | 4 | 63 | 63 | 18 | 63 |
| 92 | 63 | 8 | 63 | 18 | 63 |
| 93 | 23 | 30 | 63 | 18 | 63 |
| 94 | 17 | 8 | 63 | 18 | 63 |
| 95 | 19 | 19 | 63 | 19 | 19 |
| 96 | 19 | 19 | 19 | 19 | 19 |
| 97 | 19 | 19 | 19 | 19 | 19 |
| 98 | 19 | 19 | 19 | 19 | 19 |
| 99 | 19 | 19 | 19 | 19 | 19 |
| 100 | 20 | 19 | 19 | 19 | 19 |
| 101 | 20 | 20 | 20 | 20 | 20 |
| 102 | 20 | 20 | 20 | 20 | 20 |
| 103 | 20 | 20 | 20 | 20 | 20 |
| 104 | 20 | 20 | 20 | 20 | 20 |
| 105 | 52 | 21 | 21 | 21 | 21 |
| 106 | 21 | 21 | 21 | 21 | 21 |
| 107 | 21 | 21 | 21 | 21 | 21 |
| 108 | 21 | 21 | 21 | 21 | 21 |
| 109 | 21 | 21 | 21 | 21 | 21 |
| 110 | 22 | 22 | 22 | 22 | 22 |
| 111 | 22 | 22 | 22 | 22 | 22 |
| 112 | 22 | 22 | 22 | 22 | 22 |
| 113 | 22 | 22 | 22 | 22 | 22 |
| 114 | 22 | 22 | 22 | 22 | 22 |
| 115 | 23 | 23 | 23 | 23 | 23 |
| 116 | 23 | 23 | 23 | 23 | 23 |
| 117 | 23 | 23 | 63 | 23 | 23 |
| 118 | 23 | 23 | 63 | 23 | 23 |
| 119 | 23 | 23 | 63 | 23 | 23 |
| 120 | 24 | 24 | 24 | 24 | 24 |
| 121 | 24 | 24 | 24 | 24 | 24 |
| 122 | 24 | 24 | 24 | 24 | 24 |
| 123 | 24 | 24 | 24 | 24 | 24 |
| 124 | 25 | 24 | 63 | 24 | 63 |
| 125 | 25 | 25 | 25 | 25 | 25 |
| 126 | 25 | 25 | 25 | 25 | 25 |
| 127 | 25 | 25 | 25 | 25 | 25 |
| 128 | 25 | 25 | 25 | 25 | 25 |
| 129 | 24 | 25 | 63 | 25 | 63 |
| 130 | 26 | 26 | 10 | 26 | 26 |
| 131 | 26 | 26 | 26 | 26 | 26 |
| 132 | 26 | 26 | 26 | 26 | 26 |
| 133 | 26 | 26 | 26 | 26 | 26 |
| 134 | 26 | 10 | 25 | 26 | 26 |
| 135 | 27 | 27 | 27 | 27 | 27 |
| 136 | 3 | 25 | 28 | 27 | 3 |
| 137 | 3 | 27 | 44 | 27 | 3 |
| 138 | 27 | 27 | 37 | 27 | 27 |
| 139 | 27 | 27 | 25 | 27 | 27 |
| 140 | 28 | 17 | 5 | 28 | 5 |
| 141 | 28 | 36 | 51 | 28 | 51 |
| 142 | 28 | 28 | 63 | 28 | 28 |
| 143 | 28 | 28 | 51 | 28 | 28 |
| 144 | 28 | 28 | 51 | 28 | 28 |
| 145 | 29 | 29 | 29 | 29 | 29 |
| 146 | 29 | 32 | 29 | 29 | 29 |
| 147 | 29 | 32 | 29 | 29 | 63 |
| 148 | 29 | 31 | 63 | 29 | 63 |
| 149 | 29 | 29 | 29 | 29 | 29 |
| 150 | 30 | 30 | 30 | 30 | 30 |
| 151 | 30 | 30 | 30 | 30 | 30 |
| 152 | 30 | 30 | 30 | 30 | 30 |
| 153 | 30 | 30 | 30 | 30 | 30 |
| 154 | 30 | 30 | 30 | 30 | 30 |
| 155 | 31 | 31 | 31 | 31 | 31 |
| 156 | 31 | 31 | 31 | 31 | 31 |
| 157 | 31 | 31 | 31 | 31 | 31 |
| 158 | 29 | 32 | 29 | 31 | 29 |
| 159 | 29 | 29 | 63 | 31 | 29 |
| 160 | 32 | 32 | 32 | 32 | 32 |
| 161 | 32 | 32 | 32 | 32 | 32 |
| 162 | 32 | 32 | 32 | 32 | 32 |
| 163 | 32 | 32 | 32 | 32 | 32 |
| 164 | 32 | 32 | 32 | 32 | 32 |
| 165 | 33 | 33 | 33 | 33 | 33 |
| 166 | 33 | 33 | 33 | 33 | 33 |
| 167 | 33 | 33 | 33 | 33 | 33 |
| 168 | 33 | 33 | 33 | 33 | 33 |
| 169 | 33 | 33 | 33 | 33 | 33 |
| 170 | 33 | 32 | 25 | 34 | 33 |
| 171 | 29 | 32 | 29 | 34 | 29 |
| 172 | 29 | 9 | 63 | 34 | 63 |
| 173 | 32 | 34 | 25 | 34 | 34 |
| 174 | 34 | 34 | 30 | 34 | 34 |
| 175 | 35 | 35 | 35 | 35 | 35 |
| 176 | 35 | 35 | 35 | 35 | 35 |
| 177 | 35 | 35 | 35 | 35 | 35 |
| 178 | 35 | 35 | 35 | 35 | 35 |
| 179 | 35 | 35 | 35 | 35 | 35 |
| 180 | 36 | 36 | 36 | 36 | 36 |
| 181 | 36 | 36 | 10 | 36 | 36 |
| 182 | 36 | 42 | 25 | 36 | 42 |
| 183 | 36 | 36 | 36 | 36 | 36 |
| 184 | 36 | 36 | 36 | 36 | 36 |
| 185 | 66 | 39 | 42 | 37 | 66 |
| 186 | 3 | 36 | 42 | 37 | 42 |
| 187 | 36 | 56 | 25 | 37 | 36 |
| 188 | 37 | 37 | 37 | 37 | 37 |
| 189 | 37 | 36 | 25 | 37 | 36 |
| 190 | 38 | 38 | 38 | 38 | 38 |
| 191 | 38 | 38 | 38 | 38 | 38 |
| 192 | 38 | 37 | 25 | 38 | 38 |
| 193 | 42 | 36 | 37 | 38 | 42 |
| 194 | 38 | 38 | 38 | 38 | 38 |
| 195 | 57 | 36 | 10 | 39 | 36 |
| 196 | 39 | 39 | 39 | 39 | 39 |
| 197 | 39 | 39 | 39 | 39 | 39 |
| 198 | 39 | 39 | 39 | 39 | 39 |
| 199 | 39 | 39 | 39 | 39 | 39 |
| 200 | 40 | 43 | 40 | 40 | 40 |
| 201 | 40 | 62 | 40 | 40 | 40 |
| 202 | 40 | 43 | 40 | 40 | 40 |
| 203 | 43 | 40 | 40 | 40 | 40 |
| 204 | 43 | 62 | 43 | 40 | 62 |
| 205 | 41 | 41 | 41 | 41 | 41 |
| 206 | 41 | 41 | 41 | 41 | 41 |
| 207 | 41 | 41 | 41 | 41 | 41 |
| 208 | 41 | 41 | 41 | 41 | 41 |
| 209 | 41 | 41 | 41 | 41 | 41 |
| 210 | 42 | 42 | 42 | 42 | 42 |
| 211 | 42 | 42 | 42 | 42 | 42 |
| 212 | 42 | 42 | 42 | 42 | 42 |
| 213 | 42 | 42 | 42 | 42 | 42 |
| 214 | 42 | 42 | 42 | 42 | 42 |
| 216 | 60 | 60 | 60 | 43 | 60 |
| 217 | 60 | 60 | 10 | 43 | 60 |
| 218 | 60 | 60 | 60 | 43 | 60 |
| 219 | 60 | 52 | 60 | 43 | 60 |
| 220 | 44 | 44 | 63 | 44 | 44 |
| 221 | 44 | 44 | 44 | 44 | 44 |
| 222 | 44 | 47 | 44 | 44 | 44 |
| 223 | 27 | 44 | 57 | 44 | 44 |
| 224 | 55 | 62 | 55 | 44 | 55 |
| 225 | 45 | 45 | 45 | 45 | 45 |
| 226 | 45 | 45 | 63 | 45 | 45 |
| 227 | 45 | 45 | 45 | 45 | 45 |
| 228 | 45 | 45 | 45 | 45 | 45 |
| 229 | 45 | 45 | 45 | 45 | 45 |
| 230 | 46 | 46 | 46 | 46 | 46 |
| 231 | 46 | 46 | 46 | 46 | 46 |
| 232 | 46 | 46 | 37 | 46 | 46 |
| 233 | 46 | 46 | 46 | 46 | 46 |
| 234 | 46 | 46 | 46 | 46 | 46 |
| 235 | 19 | 49 | 19 | 47 | 19 |
| 236 | 47 | 48 | 63 | 47 | 63 |
| 237 | 49 | 47 | 51 | 47 | 51 |
| 238 | 47 | 56 | 37 | 47 | 47 |
| 239 | 48 | 49 | 63 | 47 | 63 |
| 240 | 48 | 48 | 48 | 48 | 48 |
| 241 | 48 | 48 | 48 | 48 | 48 |
| 242 | 47 | 47 | 48 | 48 | 48 |
| 243 | 49 | 60 | 19 | 48 | 49 |
| 244 | 48 | 49 | 19 | 48 | 19 |
| 245 | 49 | 48 | 49 | 49 | 49 |
| 246 | 49 | 49 | 38 | 49 | 49 |
| 247 | 12 | 14 | 63 | 49 | 63 |
| 248 | 48 | 48 | 50 | 49 | 48 |
| 249 | 57 | 53 | 43 | 49 | 57 |
| 250 | 50 | 50 | 50 | 50 | 50 |
| 251 | 50 | 11 | 55 | 50 | 55 |
| 252 | 50 | 58 | 25 | 50 | 50 |
| 253 | 50 | 58 | 10 | 50 | 50 |
| 254 | 50 | 11 | 37 | 50 | 50 |
| 255 | 51 | 51 | 51 | 51 | 51 |
| 256 | 51 | 51 | 51 | 51 | 51 |
| 257 | 57 | 48 | 57 | 51 | 57 |
| 258 | 51 | 51 | 51 | 51 | 51 |
| 259 | 51 | 51 | 51 | 51 | 51 |
| 260 | 52 | 62 | 52 | 52 | 52 |
| 261 | 52 | 62 | 42 | 52 | 42 |
| 262 | 52 | 62 | 42 | 52 | 52 |
| 263 | 60 | 60 | 25 | 52 | 60 |
| 264 | 60 | 60 | 63 | 52 | 60 |
| 265 | 53 | 53 | 53 | 53 | 53 |
| 266 | 53 | 53 | 63 | 53 | 53 |
| 267 | 54 | 58 | 63 | 53 | 63 |
| 268 | 53 | 53 | 63 | 53 | 53 |
| 269 | 53 | 53 | 63 | 53 | 53 |
| 270 | 33 | 54 | 29 | 54 | 54 |
| 271 | 60 | 53 | 63 | 54 | 63 |
| 272 | 29 | 54 | 63 | 54 | 29 |
| 273 | 29 | 53 | 63 | 54 | 63 |
| 274 | 60 | 53 | 63 | 54 | 63 |
| 275 | 55 | 56 | 56 | 55 | 55 |
| 276 | 55 | 55 | 13 | 55 | 35 |
| 277 | 55 | 35 | 10 | 55 | 55 |
| 278 | 55 | 10 | 55 | 55 | 55 |
| 279 | 55 | 43 | 51 | 55 | 55 |
| 280 | 56 | 55 | 44 | 56 | 56 |
| 281 | 56 | 10 | 63 | 56 | 63 |
| 282 | 56 | 56 | 56 | 56 | 56 |
| 283 | 56 | 56 | 56 | 56 | 56 |
| 284 | 56 | 56 | 56 | 56 | 56 |
| 285 | 25 | 30 | 63 | 57 | 63 |
| 286 | 57 | 57 | 26 | 57 | 57 |
| 287 | 55 | 62 | 55 | 57 | 55 |
| 288 | 55 | 30 | 63 | 57 | 63 |
| 289 | 55 | 30 | 63 | 57 | 63 |
| 290 | 35 | 60 | 63 | 58 | 63 |
| 291 | 35 | 58 | 10 | 58 | 35 |
| 292 | 58 | 43 | 10 | 58 | 58 |
| 293 | 35 | 36 | 35 | 58 | 35 |
| 294 | 60 | 60 | 35 | 58 | 60 |
| 295 | 35 | 35 | 35 | 59 | 35 |
| 296 | 35 | 35 | 35 | 59 | 35 |
| 297 | 35 | 35 | 35 | 59 | 35 |
| 298 | 35 | 35 | 35 | 59 | 35 |
| 299 | 35 | 35 | 35 | 59 | 35 |
| 300 | 60 | 60 | 60 | 60 | 60 |
| 301 | 60 | 60 | 60 | 60 | 60 |
| 302 | 60 | 60 | 60 | 60 | 60 |
| 303 | 60 | 60 | 60 | 60 | 60 |
| 304 | 60 | 60 | 60 | 60 | 60 |
| 305 | 61 | 61 | 61 | 61 | 61 |
| 306 | 61 | 61 | 61 | 61 | 61 |
| 307 | 61 | 61 | 61 | 61 | 61 |
| 308 | 59 | 62 | 59 | 61 | 62 |
| 309 | 35 | 62 | 63 | 61 | 63 |
| 310 | 48 | 49 | 55 | 62 | 55 |
| 311 | 52 | 52 | 36 | 62 | 52 |
| 312 | 52 | 62 | 57 | 62 | 52 |
| 313 | 52 | 52 | 55 | 62 | 52 |
| 314 | 61 | 58 | 10 | 62 | 61 |
| 315 | 63 | 63 | 63 | 63 | 63 |
| 316 | 63 | 63 | 63 | 63 | 63 |
| 317 | 63 | 63 | 63 | 63 | 63 |
| 318 | 63 | 63 | 63 | 63 | 63 |
| 319 | 63 | 63 | 63 | 63 | 63 |

**Figure 4.5: Summary of the final hypothesis.**

### 4.1.8 Conclusion from Hypothesis #1

From the above result we infer that Class label #43 is always getting classified as Class label #60 and Class label #59 is getting classified as Class label #35, due to visual similarity between the two class samples. Some examples pair of data with visual similarity which our classifiers are not able to classify properly are given below:

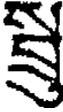CLASS #156
CLASS INDEX #43

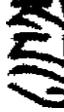CLASS #221
CLASS INDEX #60

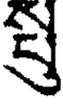CLASS #224
CLASS INDEX #62

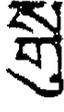CLASS #178
CLASS INDEX #52

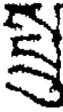CLASS #145
CLASS INDEX #40

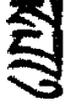CLASS #221
CLASS INDEX #60

Some examples of bad classes on which all the classifiers fail most of the time:

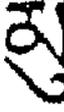CLASS #141
CLASS INDEX #37

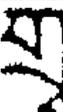CLASS #89
CLASS INDEX #27

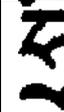CLASS #67
CLASS INDEX #18

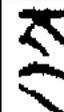CLASS #217
CLASS INDEX #57

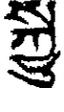CLASS #219
CLASS INDEX #58

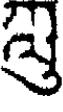CLASS #170
CLASS INDEX #47

Now, for our hypothesis to work decently at least one classifier should give the correct prediction. The number of such samples from the above observation is 255. Among them, the number of correctly classified samples by the hypothesis is 223. So our final accuracy stands at 87.45 %.

<div style="border:1px solid black; padding:20px; text-align:center;">

**ACCURACY = 87.45 %**

</div>

The samples discarded are:  44, 54, 59, 65, 66, 67, 68, 69, 74, 76, 87, 90, 91, 92, 93, 94, 136, 158, 159, 170, 171, 172, 185, 186, 187, 193, 195, 204, 215, 216, 217, 218, 219, 224, 235, 239, 243, 247, 248, 249, 257, 263, 264, 267, 271, 273, 274, 285, 287, 288, 289, 290, 293, 294, 295, 296, 297, 298, 299, 308, 309, 310, 311, 313, 314.

## 4.2   HYPOTHESIS #2

| | | | |
|---|---|---|---|
| 1 | 2.136986 | 0.002096 | 1.263013 |
| 2 | 1.989899 | 0.000000 | 1.385125 |
| 3 | 2.852941 | 0.173469 | 1.956405 |
| 4 | 3.125000 | 0.093897 | 2.653465 |
| 5 | 2.657534 | 0.000000 | 1.657507 |
| 6 | 1.258621 | 0.025641 | 1.261087 |
| 7 | 2.549296 | 0.000000 | 1.318405 |
| 8 | 2.507246 | 0.030172 | 1.711597 |
| 9 | 3.500000 | 0.000000 | 1.712746 |
| 10 | 2.207317 | 0.198157 | 1.304607 |
| 11 | 4.204545 | 0.000000 | 1.894830 |
| 12 | 3.196078 | 0.000000 | 1.684205 |
| 13 | 2.567164 | 0.005291 | 1.965306 |
| 14 | 2.967213 | 0.163655 | 1.577400 |
| 15 | 2.370370 | 0.000000 | 1.967923 |
| 16 | 3.203704 | 0.000000 | 1.754631 |
| 17 | 1.905882 | 0.000000 | 1.915720 |
| 18 | 1.806818 | 0.000000 | 2.285208 |
| 19 | 2.250000 | 0.000000 | 1.402420 |
| 20 | 1.773810 | 0.000000 | 1.638558 |
| 21 | 1.915663 | 0.000000 | 1.931738 |
| 22 | 2.744681 | 0.006593 | 1.584982 |
| 23 | 2.156627 | 0.002500 | 1.543043 |
| 24 | 2.048780 | 0.004057 | 1.943871 |
| 25 | 2.098039 | 0.000000 | 1.913143 |
| 26 | 3.700000 | 0.000000 | 1.705083 |
| 27 | 1.700000 | 0.000000 | 1.165490 |
| 28 | 1.708661 | 0.000000 | 1.883417 |
| 29 | 2.057143 | 0.001171 | 1.460387 |
| 30 | 0.882096 | 0.028616 | 0.776377 |
| 31 | 1.240741 | 0.000000 | 1.145703 |
| 32 | 1.515526 | 0.000000 | 1.426469 |
| 33 | 3.490566 | 0.018868 | 1.909027 |
| 34 | 1.045000 | 0.005263 | 0.965768 |
| 35 | 1.621212 | 0.040000 | 1.199740 |
| 36 | 2.400000 | 0.015564 | 1.822224 |
| 37 | 2.684211 | 0.005988 | 1.940009 |
| 38 | 1.476627 | 0.033473 | 1.116754 |
| 39 | 2.761194 | 0.006944 | 1.659989 |
| 40 | 2.500000 | 0.000000 | 1.477625 |
| 41 | 1.303867 | 0.000000 | 0.870387 |
| 42 | 1.471831 | 0.000000 | 1.109599 |
| 43 | 1.265306 | 0.000000 | 1.118019 |
| 44 | 1.539568 | 0.000000 | 1.320641 |
| 45 | 0.591304 | 0.000000 | 0.665224 |
| 46 | 1.239726 | 0.012270 | 1.246424 |
| 47 | 0.707627 | 0.000000 | 0.743910 |
| 48 | 1.003937 | 0.000000 | 0.719869 |
| 49 | 0.921053 | 0.000000 | 0.642028 |
| 50 | 0.895817 | 0.005848 | 1.519414 |
| 51 | 2.026316 | 0.007812 | 1.228635 |
| 52 | 0.789062 | 0.001712 | 1.490690 |
| 53 | 0.629969 | 0.000000 | 0.773983 |
| 54 | 0.920290 | 0.000000 | 0.793614 |
| 55 | 1.633540 | 0.000000 | 1.436280 |
| 56 | 0.996255 | 0.000000 | 1.068056 |
| 57 | 1.631579 | 0.000000 | 1.571265 |
| 58 | 1.080972 | 0.000000 | 1.230230 |
| 59 | 0.919463 | 0.000000 | 1.005440 |
| 60 | 0.823151 | 0.000000 | 0.646983 |
| 61 | 1.633929 | 0.000000 | 1.723647 |
| 62 | 2.179245 | 0.000000 | 1.432352 |
| 63 | 0.660668 | 0.000000 | 0.678467 |
| 64 | 1.207101 | 0.000000 | 1.979271 |
| 65 | 2.008547 | 0.000000 | 1.640450 |
| 66 | 1.664062 | 0.000000 | 1.565577 |
| 67 | 1.752066 | 0.000000 | 1.480491 |
| 68 | 2.681818 | 0.000000 | 1.884850 |
| 69 | 0.620991 | 0.012702 | 0.707969 |
| 70 | 1.290000 | 0.000000 | 1.081333 |
| 71 | 1.075862 | 0.000000 | 0.702792 |
| 72 | 1.806897 | 0.000000 | 0.957703 |
| 73 | 1.276498 | 0.000000 | 0.984750 |
| 74 | 2.297521 | 0.000000 | 1.659832 |
| 75 | 5.119048 | 0.000000 | 1.911069 |
| 76 | 1.747475 | 0.000000 | 1.247233 |
| 77 | 1.685185 | 0.000000 | 1.231524 |
| 78 | 2.562500 | 0.000000 | 1.548611 |
| 79 | 2.743243 | 0.000000 | 1.606819 |
| 80 | 2.016393 | 0.000000 | 1.534326 |
| 81 | 3.617647 | 0.000000 | 1.748230 |
| 82 | 1.223256 | 0.000000 | 1.173973 |
| 83 | 2.379310 | 0.025641 | 2.073656 |
| 84 | 3.281690 | 0.000000 | 2.078940 |
| 85 | 1.368421 | 0.000000 | 1.303816 |
| 86 | 1.207792 | 0.000000 | 1.088093 |
| 87 | 1.175325 | 0.000000 | 1.031101 |
| 88 | 0.928910 | 0.000000 | 0.636514 |
| 89 | 1.087629 | 0.000000 | 0.720903 |
| 90 | 1.637500 | 0.000000 | 1.225901 |
| 91 | 0.798450 | 0.003460 | 0.822406 |
| 92 | 0.722772 | 0.002502 | 0.571465 |
| 93 | 0.798507 | 0.001451 | 0.538463 |
| 94 | 0.859649 | 0.001134 | 0.421656 |
| 95 | 2.566265 | 0.000000 | 1.403662 |
| 96 | 1.752294 | 0.000000 | 1.948492 |
| 97 | 1.720000 | 0.000000 | 1.260662 |
| 98 | 2.197802 | 0.000000 | 1.582018 |
| 99 | 1.414414 | 0.000000 | 1.106527 |
| 100 | 2.056632 | 0.000000 | 1.132345 |
| 101 | 1.909091 | 0.000000 | 1.103694 |
| 102 | 0.619433 | 0.001020 | 0.536200 |
| 103 | 1.098592 | 0.000000 | 0.739935 |
| 104 | 1.809091 | 0.000000 | 1.007284 |
| 105 | 1.720339 | 0.000000 | 1.491874 |
| 106 | 1.515152 | 0.000000 | 1.714816 |
| 131 | 0.876777 | 0.000000 | 0.577453 |
| 132 | 1.525714 | 0.000540 | 1.222539 |
| 133 | 1.573333 | 0.000000 | 0.951481 |
| 134 | 2.320000 | 0.000000 | 1.474577 |
| 135 | 1.837183 | 0.000000 | 1.388002 |
| 136 | 2.029197 | 0.000000 | 1.989971 |
| 137 | 1.199052 | 0.000000 | 1.050087 |
| 138 | 1.130841 | 0.000000 | 1.099810 |
| 139 | 1.326829 | 0.000000 | 0.894519 |
| 140 | 1.112299 | 0.000000 | 0.927591 |
| 141 | 1.327485 | 0.000000 | 1.379680 |
| 142 | 0.611801 | 0.000000 | 0.529729 |
| 143 | 1.288344 | 0.002918 | 1.395814 |
| 144 | 1.812903 | 0.000000 | 1.440043 |
| 145 | 2.197368 | 0.000000 | 2.082267 |
| 146 | 1.892473 | 0.000000 | 1.853712 |
| 147 | 1.239130 | 0.000000 | 1.415528 |
| 148 | 2.240385 | 0.002273 | 1.727825 |
| 149 | 1.537313 | 0.000000 | 1.101374 |
| 150 | 1.656489 | 0.000000 | 1.413806 |
| 151 | 2.333333 | 0.000000 | 1.865949 |
| 152 | 1.990000 | 0.000000 | 1.286057 |
| 153 | 2.247312 | 0.000000 | 1.300984 |
| 154 | 1.925234 | 0.000000 | 1.477153 |
| 155 | 1.918182 | 0.000000 | 1.612056 |
| 156 | 1.878261 | 0.002967 | 1.917728 |
| 157 | 1.833333 | 0.000000 | 1.560226 |
| 158 | 1.831933 | 0.000000 | 1.410103 |
| 159 | 1.552000 | 0.005714 | 1.626016 |
| 160 | 1.075472 | 0.000000 | 0.925677 |
| 161 | 1.486957 | 0.000000 | 1.393619 |
| 162 | 0.927928 | 0.031807 | 0.776190 |
| 163 | 1.580357 | 0.000000 | 1.226880 |
| 164 | 2.578125 | 0.021127 | 1.488431 |
| 165 | 2.493976 | 0.000000 | 2.355993 |
| 166 | 1.715328 | 0.000000 | 1.625539 |
| 167 | 1.900000 | 0.000000 | 1.965490 |
| 168 | 2.280488 | 0.000000 | 2.257118 |
| 169 | 2.101266 | 0.000000 | 1.743911 |
| 170 | 2.746479 | 0.000000 | 1.514929 |
| 171 | 1.375000 | 0.000000 | 1.278319 |
| 172 | 1.640000 | 0.000000 | 1.010604 |
| 173 | 1.697248 | 0.000000 | 1.392763 |
| 174 | 0.814978 | 0.000000 | 0.500167 |
| 175 | 1.121495 | 0.000000 | 1.196960 |
| 176 | 1.561224 | 0.000000 | 1.080588 |
| 177 | 1.221239 | 0.000000 | 1.102521 |
| 178 | 1.594937 | 0.000000 | 1.207084 |
| 179 | 1.189873 | 0.000000 | 1.122015 |
| 180 | 3.200000 | 0.000000 | 1.588804 |
| 181 | 2.967033 | 0.000000 | 1.666216 |
| 182 | 2.103448 | 0.000000 | 1.328102 |
| 183 | 2.141593 | 0.000000 | 1.330758 |
| 184 | 3.320000 | 0.000000 | 1.436261 |
| 185 | 3.350000 | 0.000000 | 2.229983 |
| 186 | 1.082251 | 0.000000 | 1.116800 |
| 187 | 0.712538 | 0.000000 | 0.623428 |
| 188 | 0.732000 | 0.000000 | 0.523114 |
| 189 | 0.942623 | 0.000000 | 0.920398 |
| 190 | 1.309392 | 0.001965 | 0.921063 |
| 191 | 0.755556 | 0.001773 | 0.612048 |
| 192 | 1.809524 | 0.000000 | 1.596463 |
| 193 | 0.744681 | 0.000838 | 0.578954 |
| 194 | 1.243781 | 0.000000 | 0.999945 |
| 195 | 2.244444 | 0.000000 | 1.577920 |
| 196 | 2.017857 | 0.000000 | 1.679910 |
| 197 | 2.428571 | 0.000000 | 1.756892 |
| 198 | 1.426901 | 0.000000 | 1.090870 |
| 199 | 2.097561 | 0.000000 | 1.455045 |
| 200 | 1.670455 | 0.000000 | 1.250905 |
| 201 | 1.696203 | 0.000000 | 0.794497 |
| 202 | 1.500000 | 0.000000 | 1.506836 |
| 203 | 0.747059 | 0.000000 | 0.937645 |
| 204 | 0.888235 | 0.000000 | 0.913456 |
| 205 | 4.214286 | 0.011173 | 2.329536 |
| 206 | 3.632353 | 0.000000 | 2.591949 |
| 207 | 4.035714 | 0.000000 | 2.065948 |
| 208 | 2.158879 | 0.000000 | 2.152945 |
| 209 | 2.594059 | 0.000000 | 2.581111 |
| 210 | 2.182692 | 0.000000 | 2.988493 |
| 211 | 2.400000 | 0.000000 | 1.898015 |
| 212 | 1.978261 | 0.254902 | 1.712349 |
| 213 | 2.245455 | 0.000000 | 2.122535 |
| 214 | 3.484375 | 0.048458 | 1.470787 |
| 215 | 0.830565 | 0.000000 | 0.707175 |
| 216 | 0.696203 | 0.000000 | 1.124280 |
| 217 | 0.875303 | 0.005556 | 0.640981 |
| 218 | 0.496094 | 0.000000 | 0.507228 |
| 219 | 0.559259 | 0.000000 | 0.498173 |
| 220 | 1.213836 | 0.000000 | 0.750128 |
| 221 | 2.247059 | 0.000000 | 1.148501 |
| 222 | 1.306569 | 0.000000 | 0.760524 |
| 223 | 1.371212 | 0.000000 | 0.926439 |
| 224 | 1.441667 | 0.000000 | 1.090312 |
| 225 | 2.225225 | 0.000000 | 2.145909 |
| 226 | 2.774510 | 0.020619 | 1.331742 |
| 227 | 2.515464 | 0.000000 | 1.644069 |
| 228 | 1.973214 | 0.000000 | 2.386902 |
| 229 | 2.111111 | 0.000000 | 2.368995 |
| 230 | 1.816901 | 0.000000 | 1.387988 |
| 231 | 1.003300 | 0.007005 | 1.144148 |
| 232 | 1.780000 | 0.004141 | 1.997734 |
| 233 | 2.017857 | 0.000000 | 1.526808 |
| 234 | 0.865248 | 0.000000 | 0.885687 |
| 235 | 1.461538 | 0.000000 | 0.845695 |
| 236 | 1.019512 | 0.000000 | 0.591792 |
| 237 | 1.101064 | 0.000000 | 0.633319 |
| 238 | 0.957031 | 0.000000 | 0.693058 |
| 239 | 1.009852 | 0.000000 | 0.966898 |
| 240 | 1.614815 | 0.000000 | 0.838680 |
| 241 | 1.530435 | 0.000000 | 0.898347 |
| 242 | 1.446281 | 0.000000 | 0.898848 |
| 243 | 1.416667 | 0.000000 | 1.260484 |
| 244 | 0.923077 | 0.000000 | 0.977772 |
| 245 | 1.066265 | 0.000000 | 1.084416 |
| 246 | 1.243902 | 0.000000 | 1.280361 |
| 247 | 1.680000 | 0.000000 | 1.457644 |
| 248 | 2.202532 | 0.000000 | 1.364526 |
| 249 | 1.727273 | 0.000000 | 1.837176 |
| 250 | 1.902439 | 0.000000 | 1.560375 |
| 251 | 1.679389 | 0.000000 | 1.271866 |
| 252 | 1.865079 | 0.000000 | 1.719077 |
| 253 | 1.423619 | 0.000000 | 1.446599 |
| 254 | 1.835714 | 0.000000 | 1.455150 |
| 255 | 1.736842 | 0.000000 | 1.763403 |
| 256 | 1.870000 | 0.000000 | 1.593390 |
| 257 | 1.202797 | 0.000000 | 1.587684 |
| 258 | 2.110000 | 0.000000 | 1.679523 |
| 259 | 2.011111 | 0.000000 | 1.546206 |
| 260 | 2.379310 | 0.000000 | 1.789045 |
| 261 | 1.981481 | 0.000000 | 1.595301 |
| 262 | 1.606299 | 0.000000 | 1.597917 |
| 263 | 1.598802 | 0.020619 | 1.170607 |
| 264 | 0.818182 | 0.000000 | 0.711364 |
| 265 | 2.060000 | 0.000000 | 1.410317 |
| 266 | 0.785075 | 0.000000 | 0.670611 |
| 267 | 1.540698 | 0.000000 | 1.367665 |
| 268 | 1.809160 | 0.000000 | 1.097070 |
| 269 | 0.883562 | 0.000000 | 0.838358 |
| 270 | 1.783582 | 0.000000 | 0.943535 |
| 271 | 1.132000 | 0.000000 | 0.882122 |
| 272 | 1.208696 | 0.000000 | 0.746063 |
| 273 | 1.154412 | 0.001515 | 0.833830 |
| 274 | 1.205021 | 0.000000 | 0.978513 |
| 275 | 1.850000 | 0.000000 | 1.530842 |
| 276 | 1.296970 | 0.000000 | 1.111978 |
| 277 | 1.553191 | 0.019231 | 1.017735 |
| 278 | 1.510345 | 0.000000 | 1.332932 |
| 279 | 0.847134 | 0.000000 | 1.002928 |
| 280 | 1.337423 | 0.000000 | 1.531221 |
| 281 | 1.318841 | 0.000000 | 0.934397 |
| 282 | 1.765766 | 0.000000 | 0.941694 |
| 283 | 1.700935 | 0.000000 | 1.024385 |
| 284 | 2.632353 | 0.000000 | 1.207436 |
| 285 | 1.468354 | 0.000000 | 0.773128 |
| 286 | 1.122172 | 0.000000 | 0.753581 |
| 287 | 1.545455 | 0.000000 | 1.602863 |
| 288 | 1.710280 | 0.000000 | 1.429821 |
| 289 | 0.914729 | 0.000000 | 0.722475 |
| 290 | 0.939716 | 0.001988 | 0.818081 |
| 291 | 1.151220 | 0.000000 | 1.130814 |
| 292 | 1.550898 | 0.000000 | 1.285063 |
| 293 | 1.011070 | 0.000000 | 0.747454 |
| 294 | 0.715385 | 0.001572 | 0.719670 |
| 295 | 0.610169 | 0.000000 | 0.578262 |
| 296 | 0.807339 | 0.000000 | 0.909463 |
| 297 | 0.908108 | 0.003333 | 0.705687 |
| 298 | 0.971963 | 0.000000 | 0.815777 |
| 299 | 1.000000 | 0.000000 | 0.679341 |
| 300 | 1.950311 | 0.000000 | 1.468694 |
| 301 | 0.800000 | 0.003063 | 0.557631 |
| 302 | 1.522843 | 0.000000 | 0.992741 |
| 303 | 1.461988 | 0.000000 | 0.999932 |
| 304 | 2.039370 | 0.000000 | 1.382224 |

**Figure 4.6: Confidence of each sample for different classifiers.**

## 4.2.1 Summary of Hypothesis #2

| SAMPLE NUMBER | CLASSIFIER-1 | CLASSIFIER-2 | CLASSIFIER-3 | ACTUAL-CLASS | PREDICTED CLASS |
|---|---|---|---|---|---|
| 0 | 0 | 0 | 0 | 0 | 0 |
| 1 | 0 | 0 | 0 | 0 | 0 |
| 2 | 0 | 0 | 0 | 0 | 0 |
| 3 | 0 | 0 | 0 | 0 | 0 |
| 4 | 0 | 0 | 0 | 0 | 0 |
| 5 | 1 | 1 | 1 | 1 | 1 |
| 6 | 1 | 1 | 1 | 1 | 1 |
| 7 | 1 | 1 | 1 | 1 | 1 |
| 8 | 1 | 1 | 1 | 1 | 1 |
| 9 | 1 | 1 | 1 | 1 | 1 |
| 10 | 2 | 2 | 2 | 2 | 2 |
| 11 | 2 | 2 | 2 | 2 | 2 |
| 12 | 2 | 2 | 2 | 2 | 2 |
| 13 | 2 | 2 | 2 | 2 | 2 |
| 14 | 2 | 2 | 2 | 2 | 2 |
| 15 | 3 | 3 | 3 | 3 | 3 |
| 16 | 3 | 3 | 3 | 3 | 3 |
| 17 | 3 | 3 | 3 | 3 | 3 |
| 18 | 3 | 27 | 3 | 3 | 3 |
| 19 | 3 | 3 | 3 | 3 | 3 |
| 20 | 25 | 4 | 25 | 4 | 25 |
| 21 | 4 | 4 | 4 | 4 | 4 |
| 22 | 4 | 4 | 4 | 4 | 4 |
| 23 | 4 | 4 | 4 | 4 | 4 |
| 24 | 4 | 4 | 4 | 4 | 4 |
| 25 | 5 | 5 | 5 | 5 | 5 |
| 26 | 5 | 5 | 5 | 5 | 5 |
| 27 | 5 | 5 | 5 | 5 | 5 |
| 28 | 5 | 5 | 5 | 5 | 5 |
| 29 | 5 | 5 | 5 | 5 | 5 |
| 30 | 6 | 9 | 6 | 6 | 6 |
| 31 | 6 | 9 | 63 | 6 | 6 |
| 32 | 6 | 9 | 6 | 6 | 6 |
| 33 | 6 | 6 | 6 | 6 | 6 |
| 34 | 6 | 6 | 6 | 6 | 6 |
| 35 | 7 | 7 | 7 | 7 | 7 |
| 36 | 7 | 7 | 7 | 7 | 7 |
| 37 | 7 | 7 | 7 | 7 | 7 |
| 38 | 7 | 7 | 7 | 7 | 7 |
| 39 | 7 | 7 | 7 | 7 | 7 |
| 40 | 8 | 8 | 8 | 8 | 8 |
| 41 | 8 | 8 | 8 | 8 | 8 |
| 42 | 9 | 8 | 8 | 8 | 9 |
| 43 | 8 | 9 | 25 | 8 | 8 |
| 44 | 63 | 34 | 25 | 8 | 63 |
| 45 | 9 | 9 | 9 | 9 | 9 |
| 46 | 8 | 9 | 9 | 9 | 8 |
| 47 | 9 | 9 | 4 | 9 | 9 |
| 48 | 9 | 9 | 9 | 9 | 9 |
| 49 | 2 | 9 | 63 | 9 | 2 |
| 50 | 10 | 13 | 10 | 10 | 10 |
| 51 | 10 | 56 | 12 | 10 | 10 |
| 52 | 10 | 63 | 63 | 10 | 63 |
| 53 | 10 | 10 | 63 | 10 | 63 |
| 54 | 47 | 56 | 12 | 10 | 47 |
| 55 | 11 | 58 | 11 | 11 | 11 |
| 56 | 11 | 10 | 11 | 11 | 11 |
| 57 | 11 | 52 | 10 | 11 | 11 |
| 58 | 11 | 63 | 63 | 11 | 63 |
| 59 | 56 | 56 | 37 | 11 | 37 |
| 60 | 12 | 12 | 12 | 12 | 12 |
| 61 | 12 | 12 | 12 | 12 | 12 |
| 62 | 12 | 10 | 12 | 12 | 12 |
| 63 | 12 | 10 | 12 | 12 | 12 |
| 64 | 12 | 10 | 12 | 12 | 12 |
| 65 | 14 | 14 | 14 | 13 | 14 |
| 66 | 14 | 14 | 63 | 13 | 14 |
| 67 | 11 | 10 | 13 | 13 | 11 |
| 68 | 14 | 10 | 10 | 13 | 14 |
| 69 | 14 | 10 | 10 | 13 | 10 |
| 70 | 14 | 14 | 14 | 14 | 14 |
| 71 | 14 | 14 | 14 | 14 | 14 |
| 72 | 14 | 14 | 14 | 14 | 14 |
| 73 | 14 | 14 | 14 | 14 | 14 |
| 74 | 35 | 30 | 63 | 14 | 35 |
| 75 | 15 | 15 | 63 | 15 | 15 |
| 76 | 34 | 34 | 34 | 15 | 34 |
| 77 | 15 | 15 | 15 | 15 | 15 |
| 78 | 15 | 21 | 15 | 15 | 15 |
| 79 | 15 | 15 | 15 | 15 | 15 |
| 80 | 16 | 16 | 16 | 16 | 16 |
| 81 | 16 | 16 | 16 | 16 | 16 |
| 82 | 16 | 16 | 16 | 16 | 16 |
| 83 | 16 | 16 | 16 | 16 | 16 |
| 84 | 16 | 16 | 16 | 16 | 16 |
| 85 | 17 | 15 | 19 | 17 | 17 |
| 86 | 17 | 9 | 17 | 17 | 17 |
| 87 | 15 | 20 | 35 | 17 | 15 |
| 88 | 17 | 17 | 63 | 17 | 17 |
| 89 | 17 | 19 | 63 | 17 | 17 |
| 90 | 25 | 8 | 25 | 18 | 25 |
| 91 | 4 | 8 | 63 | 18 | 63 |
| 92 | 63 | 8 | 63 | 18 | 63 |
| 93 | 23 | 30 | 63 | 18 | 23 |
| 94 | 17 | 8 | 63 | 18 | 17 |
| 95 | 19 | 19 | 63 | 19 | 19 |
| 96 | 19 | 19 | 19 | 19 | 19 |
| 97 | 19 | 19 | 19 | 19 | 19 |
| 98 | 19 | 19 | 19 | 19 | 19 |
| 99 | 19 | 19 | 19 | 19 | 19 |
| 100 | 20 | 19 | 19 | 20 | 20 |
| 101 | 20 | 20 | 20 | 20 | 20 |
| 102 | 20 | 20 | 20 | 20 | 20 |
| 103 | 20 | 20 | 20 | 20 | 20 |
| 104 | 20 | 20 | 20 | 20 | 20 |
| 105 | 52 | 21 | 21 | 21 | 52 |
| 106 | 21 | 21 | 21 | 21 | 21 |
| 107 | 21 | 21 | 21 | 21 | 21 |

| | | | | |
|---|---|---|---|---|
| 108 | 21 | 21 | 21 | 21 | 21 |
| 109 | 21 | 21 | 21 | 21 | 21 |
| 110 | 22 | 22 | 22 | 22 | 22 |
| 111 | 22 | 22 | 22 | 22 | 22 |
| 112 | 22 | 22 | 22 | 22 | 22 |
| 113 | 22 | 22 | 22 | 22 | 22 |
| 114 | 22 | 22 | 22 | 22 | 22 |
| 115 | 23 | 23 | 23 | 23 | 23 |
| 116 | 23 | 23 | 23 | 23 | 23 |
| 117 | 23 | 23 | 63 | 23 | 23 |
| 118 | 23 | 23 | 63 | 23 | 23 |
| 119 | 23 | 23 | 63 | 23 | 23 |
| 120 | 24 | 24 | 24 | 24 | 24 |
| 121 | 24 | 24 | 24 | 24 | 24 |
| 122 | 24 | 24 | 24 | 24 | 24 |
| 123 | 24 | 24 | 24 | 24 | 24 |
| 124 | 25 | 24 | 63 | 25 | 25 |
| 125 | 25 | 25 | 25 | 25 | 25 |
| 126 | 25 | 25 | 25 | 25 | 25 |
| 127 | 25 | 25 | 25 | 25 | 25 |
| 128 | 25 | 25 | 25 | 25 | 25 |
| 129 | 24 | 25 | 63 | 25 | 24 |
| 130 | 26 | 26 | 10 | 26 | 26 |
| 131 | 26 | 26 | 26 | 26 | 26 |
| 132 | 26 | 26 | 26 | 26 | 26 |
| 133 | 26 | 26 | 26 | 26 | 26 |
| 134 | 26 | 10 | 25 | 26 | 26 |
| 135 | 27 | 27 | 27 | 27 | 27 |
| 136 | 3 | 27 | 25 | 27 | 3 |
| 137 | 3 | 28 | 44 | 27 | 27 |
| 138 | 27 | 27 | 37 | 27 | 27 |
| 139 | 27 | 27 | 25 | 27 | 27 |
| 140 | 28 | 17 | 5 | 28 | 28 |
| 141 | 28 | 36 | 51 | 28 | 51 |
| 142 | 28 | 28 | 63 | 28 | 28 |
| 143 | 28 | 28 | 51 | 28 | 51 |
| 144 | 28 | 28 | 51 | 28 | 28 |
| 145 | 29 | 29 | 29 | 29 | 29 |
| 146 | 29 | 32 | 29 | 29 | 29 |
| 147 | 29 | 32 | 29 | 29 | 29 |
| 148 | 29 | 31 | 63 | 29 | 29 |
| 149 | 29 | 29 | 29 | 29 | 29 |
| 150 | 30 | 30 | 30 | 30 | 30 |
| 151 | 30 | 30 | 30 | 30 | 30 |
| 152 | 30 | 30 | 30 | 30 | 30 |
| 153 | 30 | 30 | 30 | 30 | 30 |
| 154 | 30 | 30 | 30 | 30 | 30 |
| 155 | 31 | 31 | 31 | 31 | 31 |
| 156 | 31 | 31 | 31 | 31 | 31 |
| 157 | 31 | 31 | 31 | 31 | 31 |
| 158 | 29 | 32 | 29 | 31 | 29 |
| 159 | 29 | 29 | 63 | 31 | 63 |
| 160 | 32 | 32 | 32 | 32 | 32 |
| 161 | 32 | 32 | 32 | 32 | 32 |
| 162 | 32 | 32 | 32 | 32 | 32 |
| 163 | 32 | 32 | 32 | 32 | 32 |
| 164 | 32 | 32 | 32 | 32 | 32 |
| 165 | 33 | 33 | 33 | 33 | 33 |
| 166 | 33 | 33 | 33 | 33 | 33 |
| 167 | 33 | 33 | 33 | 33 | 33 |
| 168 | 33 | 33 | 33 | 33 | 33 |
| 169 | 33 | 33 | 33 | 33 | 33 |
| 170 | 33 | 32 | 25 | 34 | 33 |
| 171 | 29 | 32 | 29 | 34 | 29 |
| 172 | 29 | 9 | 63 | 34 | 29 |
| 173 | 32 | 34 | 25 | 34 | 32 |
| 174 | 34 | 34 | 30 | 34 | 34 |
| 175 | 35 | 35 | 35 | 35 | 35 |
| 176 | 35 | 35 | 35 | 35 | 35 |
| 177 | 35 | 35 | 35 | 35 | 35 |
| 178 | 35 | 35 | 35 | 35 | 35 |
| 179 | 35 | 35 | 35 | 35 | 35 |
| 180 | 36 | 36 | 36 | 36 | 36 |
| 181 | 36 | 36 | 10 | 36 | 36 |
| 182 | 36 | 42 | 25 | 36 | 36 |
| 183 | 36 | 36 | 36 | 36 | 36 |
| 184 | 36 | 36 | 36 | 36 | 36 |
| 185 | 56 | 39 | 42 | 37 | 56 |
| 186 | 5 | 36 | 42 | 37 | 42 |
| 187 | 36 | 55 | 25 | 37 | 36 |
| 188 | 37 | 37 | 37 | 37 | 37 |
| 189 | 37 | 36 | 25 | 37 | 37 |
| 190 | 38 | 38 | 38 | 38 | 38 |
| 191 | 38 | 38 | 38 | 38 | 38 |
| 192 | 38 | 37 | 25 | 38 | 38 |
| 193 | 42 | 36 | 37 | 38 | 42 |
| 194 | 38 | 38 | 38 | 38 | 38 |
| 195 | 57 | 36 | 10 | 39 | 57 |
| 196 | 39 | 39 | 39 | 39 | 39 |
| 197 | 39 | 39 | 39 | 39 | 39 |
| 198 | 39 | 39 | 39 | 39 | 39 |
| 199 | 39 | 39 | 39 | 39 | 39 |
| 200 | 40 | 43 | 40 | 40 | 40 |
| 201 | 40 | 62 | 40 | 40 | 40 |
| 202 | 40 | 43 | 40 | 40 | 40 |
| 203 | 43 | 40 | 40 | 40 | 43 |
| 204 | 43 | 62 | 43 | 40 | 43 |
| 205 | 41 | 41 | 41 | 41 | 41 |
| 206 | 41 | 41 | 41 | 41 | 41 |
| 207 | 41 | 41 | 41 | 41 | 41 |
| 208 | 41 | 41 | 41 | 41 | 41 |
| 209 | 41 | 41 | 41 | 41 | 41 |
| 210 | 42 | 42 | 42 | 42 | 42 |
| 211 | 42 | 42 | 42 | 42 | 42 |
| 212 | 42 | 42 | 42 | 42 | 42 |
| 213 | 42 | 42 | 42 | 42 | 42 |
| 214 | 42 | 42 | 42 | 42 | 42 |
| 215 | 60 | 60 | 60 | 43 | 60 |
| 216 | 60 | 60 | 10 | 43 | 60 |
| 217 | 60 | 60 | 60 | 43 | 60 |
| 218 | 60 | 60 | 60 | 43 | 60 |
| 219 | 60 | 52 | 45 | 43 | 60 |
| 220 | 44 | 44 | 53 | 44 | 44 |
| 221 | 44 | 44 | 44 | 44 | 44 |
| 222 | 44 | 47 | 44 | 44 | 44 |
| 223 | 27 | 44 | 57 | 44 | 27 |
| 224 | 55 | 62 | 55 | 44 | 55 |
| 225 | 45 | 45 | 45 | 45 | 45 |
| 226 | 45 | 45 | 63 | 45 | 45 |
| 227 | 45 | 45 | 45 | 45 | 45 |
| 228 | 45 | 45 | 45 | 45 | 45 |
| 229 | 45 | 45 | 45 | 45 | 45 |
| 230 | 46 | 46 | 46 | 46 | 46 |
| 231 | 46 | 46 | 46 | 46 | 46 |
| 232 | 46 | 46 | 37 | 46 | 37 |
| 233 | 46 | 46 | 46 | 46 | 46 |
| 234 | 46 | 46 | 46 | 46 | 46 |
| 235 | 19 | 49 | 19 | 47 | 19 |
| 236 | 47 | 48 | 63 | 47 | 47 |
| 237 | 49 | 47 | 51 | 47 | 49 |
| 238 | 47 | 56 | 37 | 47 | 47 |
| 239 | 48 | 49 | 63 | 47 | 48 |
| 240 | 48 | 48 | 48 | 48 | 48 |
| 241 | 48 | 48 | 48 | 48 | 48 |
| 242 | 47 | 47 | 48 | 48 | 48 |
| 243 | 43 | 60 | 19 | 48 | 49 |
| 244 | 48 | 49 | 19 | 48 | 19 |
| 245 | 49 | 48 | 49 | 49 | 49 |
| 246 | 49 | 49 | 36 | 49 | 36 |
| 247 | 12 | 14 | 63 | 49 | 12 |
| 248 | 48 | 48 | 50 | 49 | 48 |
| 249 | 57 | 53 | 57 | 49 | 57 |
| 250 | 50 | 50 | 50 | 50 | 50 |
| 251 | 50 | 11 | 55 | 50 | 50 |
| 252 | 50 | 58 | 25 | 50 | 50 |
| 253 | 50 | 58 | 10 | 50 | 10 |
| 254 | 50 | 11 | 37 | 50 | 50 |
| 255 | 51 | 51 | 51 | 51 | 51 |
| 256 | 51 | 51 | 51 | 51 | 51 |
| 257 | 57 | 48 | 57 | 51 | 57 |
| 258 | 51 | 51 | 51 | 51 | 51 |
| 259 | 51 | 51 | 51 | 51 | 51 |
| 260 | 52 | 62 | 52 | 52 | 52 |
| 261 | 52 | 62 | 42 | 52 | 52 |
| 262 | 52 | 62 | 42 | 52 | 52 |
| 263 | 60 | 60 | 25 | 52 | 60 |
| 264 | 60 | 60 | 63 | 52 | 60 |
| 265 | 53 | 53 | 53 | 53 | 53 |
| 266 | 53 | 53 | 63 | 53 | 53 |
| 267 | 54 | 58 | 63 | 53 | 54 |
| 268 | 53 | 53 | 63 | 53 | 53 |
| 269 | 53 | 53 | 63 | 53 | 53 |
| 270 | 33 | 54 | 23 | 54 | 33 |
| 271 | 60 | 53 | 63 | 54 | 60 |
| 272 | 29 | 54 | 29 | 54 | 29 |
| 273 | 29 | 53 | 63 | 54 | 29 |
| 274 | 60 | 53 | 63 | 54 | 60 |
| 275 | 55 | 56 | 55 | 55 | 55 |
| 276 | 55 | 55 | 13 | 55 | 55 |
| 277 | 55 | 35 | 10 | 55 | 55 |
| 278 | 55 | 10 | 55 | 55 | 55 |
| 279 | 55 | 43 | 51 | 55 | 51 |
| 280 | 56 | 55 | 44 | 56 | 44 |
| 281 | 56 | 10 | 63 | 56 | 56 |
| 282 | 56 | 56 | 56 | 56 | 56 |
| 283 | 56 | 56 | 56 | 56 | 56 |
| 284 | 56 | 56 | 56 | 56 | 56 |
| 285 | 25 | 30 | 63 | 57 | 25 |
| 286 | 57 | 57 | 25 | 57 | 57 |
| 287 | 55 | 62 | 55 | 57 | 55 |
| 288 | 55 | 30 | 63 | 57 | 55 |
| 289 | 55 | 30 | 63 | 57 | 55 |
| 290 | 35 | 60 | 63 | 58 | 35 |
| 291 | 35 | 58 | 10 | 58 | 35 |
| 292 | 58 | 43 | 10 | 58 | 58 |
| 293 | 35 | 36 | 35 | 58 | 35 |
| 294 | 60 | 60 | 35 | 58 | 35 |
| 295 | 35 | 35 | 35 | 59 | 35 |
| 296 | 35 | 35 | 35 | 59 | 35 |
| 297 | 35 | 35 | 35 | 59 | 35 |
| 298 | 35 | 35 | 35 | 59 | 35 |
| 299 | 35 | 35 | 35 | 59 | 35 |
| 300 | 60 | 60 | 60 | 60 | 60 |
| 301 | 60 | 60 | 60 | 60 | 60 |
| 302 | 60 | 60 | 60 | 60 | 60 |
| 303 | 60 | 60 | 60 | 60 | 60 |
| 304 | 60 | 60 | 60 | 60 | 60 |
| 305 | 61 | 61 | 61 | 61 | 61 |
| 306 | 61 | 61 | 61 | 61 | 61 |
| 307 | 61 | 61 | 61 | 61 | 61 |
| 308 | 59 | 62 | 59 | 61 | 59 |
| 309 | 35 | 62 | 63 | 61 | 35 |
| 310 | 48 | 49 | 55 | 62 | 48 |
| 311 | 52 | 52 | 36 | 62 | 52 |
| 312 | 52 | 62 | 57 | 62 | 52 |
| 313 | 52 | 52 | 55 | 62 | 52 |
| 314 | 61 | 58 | 10 | 62 | 61 |
| 315 | 63 | 63 | 63 | 63 | 63 |
| 316 | 63 | 63 | 63 | 63 | 63 |
| 317 | 63 | 63 | 63 | 63 | 63 |
| 318 | 63 | 63 | 63 | 63 | 63 |
| 319 | 63 | 63 | 63 | 63 | 63 |

**Figure 4.7: Summary of Hypothesis #2**

### 4.2.2 CONCLUSION FROM HYPOTHESIS #2

The first conclusion that can be drawn from Hypothesis #2 is that the confidence of the classifier-2 is zero in most of the cases, which suggests that this classifier will have lesser impact on the voting of the confidences during the formation of the hypothesis.

A very interesting observation is that from sample number 45 to 49 which belongs to class index #9 out of all the three classifiers, only classifier #2 is predicting this class correctly, but due to zero confidence of classifier #2 on those samples, the correct classification is bypassed and wrong prediction is obtained by the hypothesis. This shows that the technique used for the formation of the confidence matrix may not be reliable at all time.

In this method of hypothesis formation some observations are as follows:

Class index #13 and #43 is always getting miss-classified as class index #14 and #60 respectively.

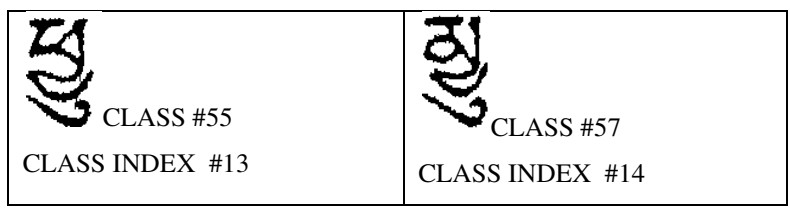

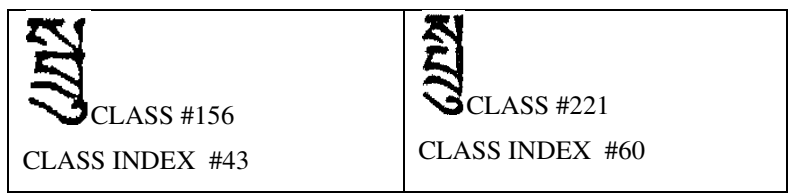

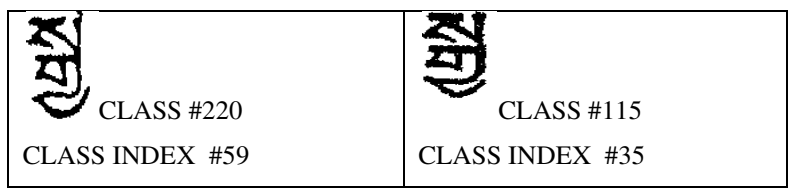

Apart from these classes which exhibit unusual similarity, there are some other classes for which the hypothesis is not able to correctly predict the output and making random errors.

Those class indices are #18(5/5 misclassification), #54(5/5 misclassification), #57(4/5 misclassification), #58(4/5 misclassification), #62 (5/5 misclassification).

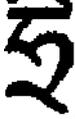
CLASS #67
CLASS INDEX #18

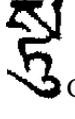
CLASS #181
CLASS INDEX #54

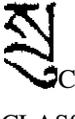
CLASS #217
CLASS INDEX #57

CLASS #219
CLASS INDEX #58

CLASS #224
CLASS INDEX #62

Also the best class in which every single classifier is giving the correct prediction which leads to correct overall prediction is class index #42.

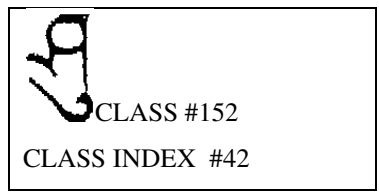
CLASS #152
CLASS INDEX #42

Again, for this hypothesis to work decently, only those samples are considered which is correctly being classified by at least one of the classifiers. The number of such samples are 255. And the number of correctly predicted samples by the hypothesis is 226. So the accuracy comes to 88.62 %.

**ACCURACY = 88.62 %**

The samples discarded are: 44, 54, 59, 65, 66, 67, 68, 69, 74, 76, 87, 90, 91, 92, 93, 94, 136, 158, 159, 170, 171, 172, 185, 186, 187, 193, 195, 204, 215, 216, 217, 218, 219, 224, 235, 239, 243, 247, 248, 249, 257, 263, 264, 267, 271, 273, 274, 285, 287, 288, 289, 290, 293, 294, 295, 296, 297, 298, 299, 308, 309, 310, 311, 313, 314.

## 4.3   HYPOTHESIS #3

### 4.3.1   Analysis of Classifier #1

**Figure 4.8: Confusion Matrix of Classifier #1**

**Conclusion:** Accuracy of Classifier #1 is 77.81 %.

## 4.3.2   Analysis of Classifier #2

**Figure 4.9: Confusion matrix of classifier #2**

**Conclusion:** Accuracy of Classifier #2 = 77.18 %.

### 4.3.3 Analysis of Classifier #3

**Fig 4.10: Confusion matrix of classifier #3**

**Conclusion:** Accuracy of Classifier #3 = 60.62 %.

### 4.3.4 Confidence Matrix

```
THE CONFIDENCE OF EACH CLASSIFIER ON INDIVIDUAL CLASSES

CLASS   CLASSIFIER-1   CLASSIFIER-2   CLASSIFIER-3

0       1.000000       1.000000       1.000000
1       0.800000       1.000000       0.800000
2       1.000000       1.000000       1.000000
3       1.000000       1.000000       0.800000
4       1.000000       1.000000       1.000000
5       0.800000       1.000000       0.800000
6       1.000000       0.600000       0.000000
7       1.000000       1.000000       0.800000
8       1.000000       1.000000       0.600000
9       0.800000       0.800000       0.200000
10      0.200000       0.600000       0.000000
11      1.000000       0.400000       0.200000
12      0.600000       0.600000       0.600000
13      0.800000       0.600000       0.800000
14      1.000000       0.800000       0.600000
15      1.000000       1.000000       1.000000
16      1.000000       1.000000       0.800000
17      0.600000       0.400000       0.400000
18      0.200000       0.200000       0.200000
19      1.000000       1.000000       0.800000
20      0.800000       1.000000       0.400000
21      1.000000       1.000000       1.000000
22      1.000000       1.000000       0.800000
23      1.000000       0.200000       1.000000
24      1.000000       0.800000       1.000000
25      1.000000       0.800000       1.000000
26      0.800000       0.800000       0.600000
27      0.600000       0.600000       0.400000
28      0.800000       1.000000       0.600000
29      0.800000       1.000000       0.600000
30      0.800000       1.000000       0.800000
31      1.000000       1.000000       1.000000
32      0.800000       1.000000       0.600000
33      1.000000       1.000000       1.000000
34      0.600000       0.600000       0.400000
35      1.000000       1.000000       0.800000
36      1.000000       0.800000       1.000000
37      0.200000       0.400000       0.200000
38      0.200000       0.200000       0.200000
39      0.800000       0.800000       0.800000
40      0.400000       0.200000       0.000000
41      1.000000       1.000000       1.000000
42      1.000000       1.000000       0.600000
43      0.000000       0.000000       0.000000
44      0.800000       1.000000       0.200000
45      1.000000       1.000000       1.000000
46      0.800000       0.800000       0.600000
47      0.400000       0.200000       0.000000
48      0.400000       0.600000       0.200000
49      1.000000       1.000000       0.800000
50      1.000000       1.000000       0.800000
51      1.000000       1.000000       1.000000
52      0.800000       0.800000       0.600000
53      0.600000       0.600000       0.500000
54      0.800000       1.000000       0.200000
55      0.800000       0.800000       0.600000
56      0.600000       0.800000       0.600000
57      0.800000       0.600000       0.400000
58      0.400000       0.600000       0.200000
59      0.000000       0.000000       0.000000
60      0.800000       0.800000       0.400000
61      1.000000       1.000000       1.000000
62      0.400000       0.600000       0.400000
63      1.000000       1.000000       1.000000
```

**Fig 4.11: Confidence matrix of hypothesis #3**.

## 4.3.5  Summary of Hypothesis #3

| SAMPLE NUMBER | CLASSIFIER-1 | CLASSIFIER-2 | CLASSIFIER-3 | ACTUAL-CLASS | PREDICTED-CLASS |
|---|---|---|---|---|---|
| 0 | 0 | 0 | 0 | 0 | 0 |
| 1 | 0 | 0 | 0 | 0 | 0 |
| 2 | 0 | 0 | 0 | 0 | 0 |
| 3 | 0 | 0 | 0 | 0 | 0 |
| 4 | 8 | 0 | 63 | 0 | 63 |
| 5 | 1 | 1 | 1 | 1 | 1 |
| 6 | 1 | 1 | 1 | 1 | 1 |
| 7 | 1 | 1 | 1 | 1 | 1 |
| 8 | 1 | 1 | 1 | 1 | 1 |
| 9 | 1 | 1 | 1 | 1 | 1 |
| 10 | 2 | 2 | 2 | 2 | 2 |
| 11 | 2 | 2 | 2 | 2 | 2 |
| 12 | 2 | 2 | 2 | 2 | 2 |
| 13 | 2 | 2 | 2 | 2 | 2 |
| 14 | 2 | 2 | 2 | 2 | 2 |
| 15 | 3 | 3 | 3 | 3 | 3 |
| 16 | 3 | 3 | 3 | 3 | 3 |
| 17 | 3 | 3 | 3 | 3 | 3 |
| 18 | 27 | 28 | 63 | 3 | 63 |
| 19 | 3 | 3 | 3 | 3 | 3 |
| 20 | 23 | 4 | 63 | 3 | 63 |
| 21 | 4 | 4 | 4 | 4 | 4 |
| 22 | 4 | 4 | 4 | 4 | 4 |
| 23 | 4 | 4 | 4 | 4 | 4 |
| 24 | 4 | 4 | 4 | 4 | 4 |
| 25 | 5 | 5 | 5 | 5 | 5 |
| 26 | 5 | 5 | 5 | 5 | 5 |
| 27 | 5 | 5 | 5 | 5 | 5 |
| 28 | 5 | 5 | 5 | 5 | 5 |
| 29 | 5 | 5 | 5 | 5 | 5 |
| 30 | 6 | 9 | 6 | 6 | 6 |
| 31 | 6 | 9 | 63 | 6 | 63 |
| 32 | 6 | 9 | 6 | 6 | 6 |
| 33 | 6 | 9 | 6 | 6 | 6 |
| 34 | 6 | 6 | 6 | 6 | 6 |
| 35 | 7 | 7 | 7 | 7 | 7 |
| 36 | 7 | 7 | 7 | 7 | 7 |
| 37 | 7 | 7 | 7 | 7 | 7 |
| 38 | 7 | 7 | 7 | 7 | 7 |
| 39 | 7 | 7 | 7 | 7 | 7 |
| 40 | 8 | 8 | 8 | 8 | 8 |
| 41 | 29 | 8 | 63 | 8 | 63 |
| 42 | 8 | 8 | 8 | 8 | 8 |
| 43 | 9 | 8 | 8 | 8 | 8 |
| 44 | 23 | 23 | 63 | 8 | 23 |
| 45 | 9 | 9 | 9 | 9 | 9 |
| 46 | 9 | 9 | 63 | 9 | 9 |
| 47 | 9 | 9 | 8 | 9 | 9 |
| 48 | 9 | 9 | 8 | 9 | 9 |
| 49 | 17 | 9 | 63 | 9 | 63 |
| 50 | 12 | 14 | 63 | 10 | 63 |
| 51 | 13 | 14 | 63 | 10 | 63 |
| 52 | 10 | 56 | 63 | 10 | 63 |
| 53 | 55 | 35 | 63 | 10 | 63 |
| 54 | 12 | 56 | 63 | 10 | 63 |
| 55 | 11 | 11 | 11 | 11 | 11 |
| 56 | 11 | 11 | 11 | 11 | 11 |
| 57 | 11 | 11 | 11 | 11 | 11 |
| 58 | 45 | 36 | 63 | 11 | 63 |
| 59 | 45 | 56 | 63 | 11 | 63 |
| 60 | 12 | 12 | 12 | 12 | 12 |
| 61 | 12 | 12 | 12 | 12 | 12 |
| 62 | 12 | 14 | 12 | 12 | 12 |
| 63 | 12 | 12 | 12 | 12 | 12 |
| 64 | 12 | 11 | 12 | 12 | 12 |
| 65 | 55 | 14 | 63 | 13 | 63 |
| 66 | 14 | 14 | 11 | 13 | 14 |
| 67 | 14 | 10 | 63 | 13 | 63 |
| 68 | 14 | 10 | 11 | 13 | 14 |
| 69 | 14 | 10 | 11 | 13 | 14 |
| 70 | 14 | 14 | 14 | 14 | 14 |
| 71 | 14 | 14 | 14 | 14 | 14 |
| 72 | 14 | 14 | 14 | 14 | 14 |
| 73 | 14 | 14 | 14 | 14 | 14 |
| 74 | 29 | 30 | 63 | 14 | 63 |
| 75 | 15 | 19 | 63 | 15 | 63 |
| 76 | 26 | 19 | 21 | 15 | 21 |
| 77 | 15 | 15 | 15 | 15 | 15 |
| 78 | 15 | 21 | 15 | 15 | 15 |
| 79 | 15 | 15 | 15 | 15 | 15 |
| 80 | 16 | 16 | 16 | 16 | 16 |
| 81 | 16 | 16 | 16 | 16 | 16 |
| 82 | 16 | 16 | 16 | 16 | 16 |
| 83 | 16 | 16 | 16 | 16 | 16 |
| 84 | 16 | 16 | 16 | 16 | 16 |
| 85 | 17 | 15 | 63 | 17 | 63 |
| 86 | 17 | 15 | 15 | 17 | 15 |
| 87 | 29 | 21 | 35 | 17 | 21 |
| 88 | 17 | 17 | 63 | 17 | 17 |
| 89 | 17 | 19 | 63 | 17 | 63 |
| 90 | 23 | 8 | 63 | 18 | 63 |
| 91 | 23 | 8 | 63 | 18 | 63 |
| 92 | 63 | 8 | 63 | 18 | 63 |
| 93 | 4 | 8 | 63 | 18 | 63 |
| 94 | 23 | 29 | 63 | 18 | 63 |
| 95 | 19 | 19 | 63 | 19 | 63 |
| 96 | 19 | 19 | 19 | 19 | 19 |
| 97 | 19 | 19 | 19 | 19 | 19 |
| 98 | 19 | 19 | 19 | 19 | 19 |
| 107 | 21 | 21 | 21 | 21 | 21 |
| 108 | 21 | 21 | 21 | 21 | 21 |
| 109 | 21 | 21 | 21 | 22 | 22 |
| 110 | 22 | 22 | 21 | 22 | 22 |
| 111 | 22 | 22 | 63 | 22 | 22 |
| 112 | 22 | 22 | 22 | 22 | 63 |
| 113 | 25 | 22 | 63 | 22 | 63 |
| 114 | 22 | 22 | 23 | 23 | 23 |
| 115 | 23 | 23 | 23 | 23 | 23 |
| 116 | 23 | 23 | 23 | 23 | 23 |
| 117 | 23 | 23 | 63 | 23 | 23 |
| 118 | 23 | 35 | 23 | 23 | 23 |
| 119 | 29 | 29 | 63 | 23 | 29 |
| 120 | 24 | 24 | 24 | 24 | 24 |
| 121 | 24 | 24 | 24 | 24 | 24 |
| 122 | 24 | 24 | 24 | 24 | 24 |
| 123 | 24 | 24 | 24 | 24 | 24 |
| 124 | 25 | 25 | 25 | 24 | 25 |
| 125 | 25 | 25 | 25 | 25 | 25 |
| 126 | 25 | 25 | 25 | 25 | 25 |
| 127 | 25 | 25 | 63 | 25 | 25 |
| 128 | 25 | 25 | 63 | 25 | 25 |
| 129 | 25 | 25 | 63 | 25 | 25 |
| 130 | 26 | 26 | 26 | 26 | 26 |
| 131 | 26 | 26 | 26 | 26 | 26 |
| 132 | 26 | 26 | 26 | 26 | 26 |
| 133 | 26 | 26 | 26 | 26 | 26 |
| 134 | 26 | 26 | 26 | 26 | 26 |
| 135 | 27 | 27 | 27 | 27 | 27 |
| 136 | 26 | 26 | 63 | 27 | 26 |
| 137 | 27 | 27 | 25 | 27 | 27 |
| 138 | 27 | 27 | 27 | 27 | 27 |
| 139 | 27 | 27 | 27 | 27 | 27 |
| 140 | 28 | 17 | 63 | 28 | 63 |
| 141 | 29 | 17 | 50 | 28 | 50 |
| 142 | 29 | 17 | 63 | 28 | 63 |
| 143 | 29 | 17 | 63 | 28 | 63 |
| 144 | 29 | 17 | 63 | 28 | 63 |
| 145 | 29 | 29 | 29 | 29 | 29 |
| 146 | 29 | 29 | 29 | 29 | 29 |
| 147 | 29 | 31 | 29 | 29 | 29 |
| 148 | 29 | 29 | 29 | 29 | 29 |
| 149 | 29 | 29 | 29 | 29 | 29 |
| 150 | 30 | 30 | 30 | 30 | 30 |
| 151 | 30 | 30 | 30 | 30 | 30 |
| 152 | 30 | 30 | 30 | 30 | 30 |
| 153 | 30 | 30 | 30 | 30 | 30 |
| 154 | 30 | 30 | 30 | 30 | 31 |
| 155 | 31 | 31 | 31 | 31 | 31 |
| 156 | 31 | 31 | 31 | 31 | 31 |
| 157 | 31 | 31 | 31 | 31 | 31 |
| 158 | 29 | 32 | 29 | 31 | 29 |
| 159 | 29 | 29 | 29 | 31 | 29 |
| 160 | 32 | 32 | 32 | 32 | 32 |
| 161 | 32 | 32 | 32 | 32 | 32 |
| 162 | 32 | 32 | 32 | 32 | 32 |
| 163 | 32 | 32 | 63 | 32 | 32 |
| 164 | 32 | 32 | 32 | 32 | 32 |
| 165 | 33 | 33 | 63 | 33 | 33 |
| 166 | 33 | 33 | 33 | 33 | 33 |
| 167 | 33 | 33 | 33 | 33 | 33 |
| 168 | 33 | 33 | 33 | 33 | 33 |
| 169 | 33 | 33 | 33 | 33 | 33 |
| 170 | 33 | 17 | 63 | 34 | 63 |
| 171 | 9 | 29 | 63 | 34 | 17 |
| 172 | 17 | 17 | 63 | 34 | 63 |
| 173 | 33 | 17 | 63 | 34 | 63 |
| 174 | 34 | 34 | 63 | 34 | 34 |
| 175 | 59 | 35 | 35 | 35 | 35 |
| 176 | 35 | 35 | 35 | 35 | 35 |
| 177 | 35 | 35 | 35 | 35 | 35 |
| 178 | 35 | 35 | 35 | 35 | 35 |
| 179 | 35 | 35 | 35 | 35 | 35 |
| 180 | 36 | 36 | 36 | 36 | 36 |
| 181 | 36 | 36 | 63 | 36 | 36 |
| 182 | 55 | 22 | 50 | 36 | 22 |
| 183 | 36 | 36 | 36 | 36 | 36 |
| 184 | 36 | 36 | 36 | 36 | 36 |
| 185 | 39 | 39 | 39 | 37 | 25 |
| 186 | 25 | 17 | 25 | 37 | 37 |
| 187 | 39 | 39 | 39 | 37 | 39 |
| 188 | 37 | 38 | 63 | 37 | 37 |
| 189 | 37 | 38 | 38 | 37 | 63 |
| 190 | 38 | 39 | 38 | 38 | 39 |
| 191 | 38 | 39 | 8 | 38 | 39 |
| 192 | 38 | 39 | 39 | 38 | 39 |
| 193 | 37 | 39 | 63 | 38 | 63 |
| 194 | 38 | 38 | 38 | 38 | 38 |
| 195 | 14 | 36 | 50 | 39 | 14 |
| 196 | 39 | 39 | 39 | 39 | 39 |
| 197 | 8 | 9 | 25 | 39 | 25 |
| 198 | 8 | 38 | 25 | 39 | 25 |
| 199 | 8 | 8 | 25 | 39 | 8 |
| 200 | 40 | 52 | 40 | 40 | 52 |
| 201 | 40 | 52 | 40 | 40 | 52 |
| 202 | 40 | 40 | 40 | 40 | 40 |
| 203 | 43 | 43 | 40 | 40 | 63 |
| 204 | 43 | 43 | 43 | 40 | 63 |
| 205 | 41 | 41 | 41 | 41 | 41 |
| 206 | 41 | 41 | 63 | 41 | 41 |
| 207 | 41 | 41 | 41 | 41 | 41 |
| 208 | 41 | 41 | 41 | 41 | 41 |
| 209 | 41 | 41 | 41 | 41 | 41 |
| 210 | 42 | 42 | 26 | 42 | 42 |
| 211 | 42 | 42 | 42 | 42 | 42 |
| 212 | 37 | 38 | 42 | 42 | 26 |
| 213 | 42 | 42 | 42 | 42 | 42 |
| 214 | 42 | 42 | 63 | 42 | 42 |
| 215 | 60 | 60 | 63 | 43 | 60 |
| 216 | 60 | 52 | 63 | 43 | 63 |

| 217 | 60 | 58 | 60 | 43 | 60 |
|-----|----|----|----|----|----|
| 218 | 60 | 60 | 63 | 43 | 60 |
| 219 | 60 | 52 | 63 | 43 | 63 |
| 220 | 44 | 44 | 63 | 44 | 44 |
| 221 | 44 | 44 | 44 | 44 | 44 |
| 222 | 44 | 44 | 44 | 44 | 44 |
| 223 | 56 | 56 | 56 | 44 | 56 |
| 224 | 11 | 12 | 12 | 44 | 12 |
| 225 | 45 | 45 | 45 | 45 | 45 |
| 226 | 45 | 45 | 63 | 45 | 45 |
| 227 | 45 | 45 | 45 | 45 | 45 |
| 228 | 45 | 45 | 45 | 45 | 45 |
| 229 | 45 | 45 | 45 | 45 | 45 |
| 230 | 46 | 46 | 46 | 46 | 46 |
| 231 | 46 | 46 | 46 | 46 | 46 |
| 232 | 37 | 46 | 63 | 46 | 63 |
| 233 | 46 | 46 | 46 | 46 | 46 |
| 234 | 46 | 46 | 63 | 46 | 46 |
| 235 | 48 | 49 | 63 | 47 | 63 |
| 236 | 47 | 49 | 63 | 47 | 63 |
| 237 | 51 | 49 | 63 | 47 | 63 |
| 238 | 27 | 56 | 63 | 47 | 63 |
| 239 | 48 | 49 | 63 | 47 | 63 |
| 240 | 48 | 48 | 48 | 48 | 48 |
| 241 | 48 | 48 | 48 | 48 | 48 |
| 242 | 48 | 60 | 48 | 48 | 60 |
| 243 | 48 | 60 | 63 | 48 | 63 |
| 244 | 49 | 49 | 63 | 48 | 49 |
| 245 | 49 | 49 | 49 | 49 | 49 |
| 246 | 55 | 21 | 50 | 49 | 63 |
| 247 | 43 | 50 | 63 | 49 | 63 |
| 248 | 11 | 47 | 63 | 49 | 63 |
| 249 | 14 | 21 | 40 | 49 | 63 |
| 250 | 50 | 50 | 50 | 50 | 50 |
| 251 | 48 | 49 | 63 | 50 | 63 |
| 252 | 50 | 58 | 50 | 50 | 50 |
| 253 | 50 | 58 | 50 | 50 | 50 |
| 254 | 21 | 49 | 63 | 50 | 63 |
| 255 | 51 | 51 | 51 | 51 | 51 |
| 256 | 51 | 51 | 25 | 51 | 25 |
| 257 | 55 | 21 | 25 | 51 | 25 |
| 258 | 51 | 51 | 51 | 51 | 51 |
| 259 | 51 | 51 | 51 | 51 | 51 |
| 260 | 52 | 62 | 63 | 52 | 62 |
| 261 | 62 | 62 | 51 | 52 | 62 |
| 262 | 62 | 62 | 51 | 52 | 62 |
| 263 | 50 | 10 | 50 | 52 | 50 |
| 264 | 35 | 60 | 63 | 52 | 63 |
| 265 | 53 | 53 | 53 | 53 | 53 |
| 266 | 53 | 53 | 63 | 53 | 63 |
| 267 | 54 | 58 | 63 | 53 | 63 |
| 268 | 53 | 40 | 63 | 53 | 63 |
| 269 | 53 | 53 | 63 | 53 | 63 |
| 270 | 60 | 54 | 63 | 54 | 63 |
| 271 | 60 | 53 | 63 | 54 | 63 |
| 272 | 60 | 54 | 63 | 54 | 63 |
| 273 | 58 | 54 | 63 | 54 | 63 |
| 274 | 60 | 46 | 63 | 54 | 63 |
| 275 | 55 | 55 | 55 | 55 | 55 |
| 276 | 55 | 55 | 63 | 55 | 55 |
| 277 | 55 | 55 | 55 | 55 | 55 |
| 278 | 55 | 13 | 63 | 55 | 63 |
| 279 | 58 | 14 | 63 | 55 | 63 |
| 280 | 56 | 36 | 36 | 56 | 36 |
| 281 | 56 | 56 | 63 | 56 | 56 |
| 282 | 56 | 56 | 56 | 56 | 56 |
| 283 | 56 | 56 | 63 | 56 | 56 |
| 284 | 56 | 56 | 56 | 56 | 56 |
| 285 | 8 | 29 | 63 | 57 | 29 |
| 286 | 55 | 29 | 50 | 57 | 63 |
| 287 | 14 | 12 | 63 | 57 | 63 |
| 288 | 27 | 38 | 25 | 57 | 25 |
| 289 | 59 | 57 | 25 | 57 | 25 |
| 290 | 35 | 40 | 63 | 58 | 63 |
| 291 | 59 | 56 | 63 | 58 | 63 |
| 292 | 58 | 58 | 58 | 58 | 58 |
| 293 | 35 | 56 | 63 | 58 | 63 |
| 294 | 35 | 60 | 63 | 58 | 63 |
| 295 | 51 | 59 | 63 | 59 | 63 |
| 296 | 35 | 35 | 35 | 59 | 35 |
| 297 | 35 | 35 | 35 | 59 | 35 |
| 298 | 35 | 35 | 35 | 59 | 35 |
| 299 | 35 | 35 | 63 | 59 | 35 |
| 300 | 60 | 60 | 63 | 60 | 60 |
| 301 | 60 | 60 | 63 | 60 | 60 |
| 302 | 60 | 60 | 60 | 60 | 60 |
| 303 | 60 | 58 | 60 | 60 | 60 |
| 304 | 60 | 60 | 63 | 60 | 60 |
| 305 | 61 | 61 | 63 | 61 | 61 |
| 306 | 61 | 61 | 61 | 61 | 61 |
| 307 | 61 | 61 | 61 | 61 | 61 |
| 308 | 59 | 59 | 59 | 61 | 63 |
| 309 | 59 | 59 | 63 | 61 | 63 |
| 310 | 48 | 49 | 63 | 62 | 63 |
| 311 | 27 | 52 | 63 | 62 | 63 |
| 312 | 52 | 52 | 63 | 62 | 52 |
| 313 | 52 | 52 | 63 | 62 | 52 |
| 314 | 61 | 58 | 60 | 62 | 61 |
| 315 | 63 | 63 | 63 | 63 | 63 |
| 316 | 63 | 63 | 63 | 63 | 63 |
| 317 | 63 | 63 | 63 | 63 | 63 |

**Fig 4.12: Summary of Hypothesis #3**

### 4.3.6 Conclusion from Hypothesis #3

From the above results we observe that samples from some class labels are misclassified in to some other particular class labels. Class indices #10, #18, #47, #54 are always getting misclassified as class index #63. Class #59 is getting classified in to Class #35. They are shown below:

Some class labels which are performing poorly in this Hypothesis are Class #13, #40, #52, #57, #62.

| | |
|---|---|
| 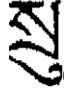CLASS #217<br>CLASS INDEX #57 | 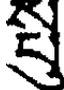CLASS #224<br>CLASS INDEX #62 |

For the hypothesis to work correctly, those samples are considered on which at least one classifier has given a correct prediction. Using the above table, we find that there are 233 such samples, out of which the number of correctly classified samples by the hypothesis are 194. Thus the final accuracy is 83.26%.

**ACCURACY = 83.26 %**

The samples discarded are: 18, 44, 50, 51, 52, 53, 54, 58, 59, 65, 66, 67, 68, 69, 74, 76, 87, 119, 124, 136, 141, 142, 143, 144, 158, 159, 170, 171, 172, 173, 182, 185, 186, 187, 192, 193, 195, 197, 199, 204, 215, 216, 217, 218, 219, 223, 224, 235, 236, 237, 238, 239, 244, 246, 247, 248, 249, 251, 254, 257, 261, 262, 263, 264, 267, 271, 274, 279, 285, 286, 287, 288, 289, 290, 293, 294, 296, 297, 298, 299, 308, 309, 310, 311, 312, 313, 314.

## 4.4 Hypothesis #4

### 4.4.1 Analysis of Classifier #1

**Figure 4.13: Confusion matrix of classifier #1.**

**Conclusion:** Accuracy of classifier #1 = 77.81 %.

### 4.4.2   Analysis of classifier #2

**Figure 4.14: Confusion matrix of classifier #2.**

**Conclusion:** Accuracy of classifier #2 = 63.12 %.

### 4.4.3 Analysis of classifier #3

**Fig 4.15: Confusion matrix of classifier #3**

**Conclusion:** Accuracy of classifier #3 = 4.37 %. Most of the samples are getting incorrectly classified.

### 4.4.4 Confidence matrix

```
THE CONFIDENCE OF EACH CLASSIFIER ON INDIVIDUAL CLASSES

CLASS   CLASSIFIER-1    CLASSIFIER-2    CLASSIFIER-3

0       1.000000        0.800000        0.000000
1       0.800000        1.000000        0.000000
2       1.000000        1.000000        0.000000
3       1.000000        0.600000        0.000000
4       1.000000        1.000000        0.000000
5       0.800000        1.000000        0.000000
6       1.000000        0.400000        0.200000
7       1.000000        0.400000        0.000000
8       1.000000        0.800000        0.400000
9       0.800000        0.800000        0.000000
10      0.200000        0.000000        0.000000
11      1.000000        0.600000        0.800000
12      0.600000        0.600000        0.000000
13      0.800000        0.600000        0.000000
14      1.000000        0.400000        0.000000
15      1.000000        1.000000        0.000000
16      1.000000        0.800000        0.000000
17      0.600000        0.400000        0.000000
18      0.200000        0.200000        0.000000
19      1.000000        1.000000        0.000000
20      0.800000        0.600000        0.000000
21      1.000000        1.000000        0.000000
22      1.000000        0.600000        0.000000
23      1.000000        0.200000        0.000000
24      1.000000        0.800000        0.000000
25      1.000000        0.200000        0.200000
26      0.800000        0.800000        0.000000
27      0.600000        0.800000        0.000000
28      0.800000        0.600000        0.000000
29      0.800000        1.000000        0.400000
30      0.800000        1.000000        0.000000
31      1.000000        0.600000        0.000000
32      0.800000        0.800000        0.000000
33      1.000000        1.000000        0.000000
34      0.600000        0.400000        0.000000
35      1.000000        0.800000        0.000000
36      1.000000        1.000000        0.000000
37      0.200000        0.200000        0.000000
38      0.200000        0.000000        0.000000
39      0.800000        0.800000        0.000000
40      0.400000        0.200000        0.000000
41      1.000000        1.000000        0.000000
42      1.000000        1.000000        0.000000
43      0.000000        0.000000        0.000000
44      0.800000        1.000000        0.000000
45      1.000000        0.800000        0.000000
46      0.800000        0.800000        0.000000
47      0.400000        0.200000        0.000000
48      0.400000        0.200000        0.000000
49      1.000000        1.000000        0.000000
50      1.000000        0.800000        0.000000
51      1.000000        0.600000        0.000000
52      0.800000        0.600000        0.000000
53      0.600000        0.200000        0.000000
54      0.800000        0.400000        0.000000
55      0.800000        0.800000        0.400000
56      0.600000        0.600000        0.000000
57      0.800000        0.600000        0.000000
58      0.400000        0.600000        0.000000
59      0.000000        0.000000        0.400000
60      0.800000        0.400000        0.000000
61      1.000000        1.000000        0.000000
62      0.400000        0.400000        0.000000
63      1.000000        0.600000        0.000000
```

**Figure 4.16: Confidence matrix of Hypothesis #4**

## 4.4.5 Summary of Hypothesis #4

| SAMPLE NUMBER | CLASSIFIER-1 | CLASSIFIER-2 | CLASSIFIER-3 | ACTUAL-CLASS | PREDICTED CLASS |
|---|---|---|---|---|---|
| 0 | 0 | 0 | 47 | 0 | 0 |
| 1 | 0 | 0 | 47 | 0 | 0 |
| 2 | 0 | 0 | 47 | 0 | 0 |
| 3 | 0 | 0 | 47 | 0 | 0 |
| 4 | 0 | 63 | 39 | 0 | 0 |
| 5 | 1 | 1 | 11 | 1 | 1 |
| 6 | 1 | 1 | 25 | 1 | 1 |
| 7 | 1 | 1 | 39 | 1 | 1 |
| 8 | 1 | 1 | 47 | 1 | 1 |
| 9 | 1 | 1 | 11 | 1 | 1 |
| 10 | 2 | 2 | 47 | 2 | 2 |
| 11 | 2 | 2 | 47 | 2 | 2 |
| 12 | 2 | 2 | 47 | 2 | 2 |
| 13 | 2 | 2 | 47 | 2 | 2 |
| 14 | 2 | 2 | 47 | 2 | 2 |
| 15 | 3 | 3 | 47 | 3 | 3 |
| 16 | 3 | 16 | 47 | 3 | 3 |
| 17 | 3 | 3 | 47 | 3 | 3 |
| 18 | 27 | 27 | 50 | 3 | 27 |
| 19 | 3 | 24 | 25 | 3 | 3 |
| 20 | 23 | 4 | 23 | 4 | 23 |
| 21 | 4 | 4 | 47 | 4 | 4 |
| 22 | 4 | 4 | 47 | 4 | 4 |
| 23 | 4 | 4 | 47 | 4 | 4 |
| 24 | 4 | 4 | 11 | 4 | 4 |
| 25 | 5 | 5 | 47 | 5 | 5 |
| 26 | 5 | 5 | 59 | 5 | 5 |
| 27 | 5 | 5 | 47 | 5 | 5 |
| 28 | 5 | 5 | 47 | 5 | 5 |
| 29 | 5 | 5 | 47 | 5 | 5 |
| 30 | 6 | 8 | 55 | 6 | 6 |
| 31 | 6 | 8 | 11 | 6 | 6 |
| 32 | 6 | 23 | 6 | 6 | 6 |
| 33 | 6 | 6 | 55 | 6 | 6 |
| 34 | 6 | 6 | 11 | 6 | 6 |
| 35 | 7 | 7 | 47 | 7 | 7 |
| 36 | 7 | 7 | 47 | 7 | 7 |
| 37 | 7 | 7 | 47 | 7 | 7 |
| 38 | 7 | 0 | 47 | 7 | 7 |
| 39 | 7 | 7 | 47 | 7 | 7 |
| 40 | 8 | 8 | 11 | 8 | 8 |
| 41 | 8 | 8 | 8 | 8 | 8 |
| 42 | 8 | 8 | 11 | 8 | 8 |
| 43 | 9 | 8 | 8 | 8 | 8 |
| 44 | 23 | 23 | 11 | 8 | 23 |
| 45 | 9 | 9 | 39 | 9 | 9 |
| 46 | 9 | 8 | 59 | 9 | 9 |
| 47 | 9 | 8 | 11 | 9 | 11 |
| 48 | 9 | 8 | 11 | 9 | 11 |
| 49 | 8 | 8 | 11 | 9 | 8 |
| 50 | 12 | 11 | 39 | 10 | 12 |
| 51 | 13 | 11 | 62 | 10 | 13 |
| 52 | 10 | 27 | 55 | 10 | 27 |
| 53 | 55 | 14 | 11 | 10 | 55 |
| 54 | 12 | 3 | 55 | 10 | 12 |
| 55 | 11 | 11 | 11 | 11 | 11 |
| 56 | 11 | 11 | 11 | 11 | 11 |
| 57 | 11 | 11 | 11 | 11 | 11 |
| 58 | 45 | 42 | 11 | 11 | 45 |
| 59 | 45 | 59 | 11 | 11 | 45 |
| 60 | 12 | 12 | 47 | 12 | 12 |
| 61 | 12 | 12 | 44 | 12 | 12 |
| 62 | 12 | 39 | 50 | 12 | 12 |
| 63 | 12 | 12 | 47 | 12 | 12 |
| 64 | 12 | 11 | 47 | 12 | 12 |
| 65 | 55 | 14 | 11 | 13 | 55 |
| 66 | 14 | 14 | 11 | 13 | 14 |
| 67 | 14 | 55 | 55 | 13 | 55 |
| 68 | 14 | 8 | 55 | 13 | 14 |
| 69 | 14 | 14 | 11 | 13 | 14 |
| 70 | 14 | 14 | 6 | 14 | 14 |
| 71 | 14 | 61 | 11 | 14 | 61 |
| 72 | 14 | 36 | 11 | 14 | 36 |
| 73 | 14 | 36 | 11 | 14 | 36 |
| 74 | 29 | 11 | 55 | 14 | 29 |
| 75 | 15 | 19 | 48 | 15 | 19 |
| 76 | 26 | 27 | 11 | 15 | 27 |
| 77 | 15 | 15 | 29 | 15 | 15 |
| 78 | 15 | 21 | 44 | 15 | 21 |
| 79 | 15 | 15 | 47 | 15 | 15 |
| 80 | 16 | 16 | 47 | 16 | 16 |
| 81 | 16 | 16 | 47 | 16 | 16 |
| 82 | 16 | 16 | 27 | 16 | 16 |
| 83 | 16 | 16 | 58 | 16 | 16 |
| 84 | 16 | 58 | 59 | 16 | 16 |
| 85 | 17 | 15 | 59 | 17 | 15 |
| 86 | 17 | 15 | 39 | 17 | 15 |
| 87 | 29 | 15 | 11 | 17 | 15 |
| 88 | 17 | 17 | 11 | 17 | 17 |
| 89 | 17 | 15 | 55 | 17 | 15 |
| 90 | 23 | 44 | 23 | 18 | 44 |
| 91 | 23 | 12 | 23 | 18 | 23 |
| 92 | 6 | 57 | 29 | 18 | 6 |
| 93 | 4 | 57 | 11 | 18 | 4 |
| 94 | 23 | 63 | 23 | 18 | 23 |
| 95 | 19 | 17 | 22 | 19 | 19 |
| 96 | 19 | 19 | 11 | 19 | 19 |
| 97 | 19 | 19 | 25 | 19 | 19 |
| 98 | 19 | 19 | 55 | 19 | 19 |
| 99 | 19 | 19 | 27 | 19 | 19 |
| 100 | 19 | 48 | 59 | 20 | 19 |
| 101 | 20 | 20 | 47 | 20 | 20 |
| 102 | 20 | 20 | 47 | 20 | 20 |
| 103 | 19 | 48 | 29 | 20 | 19 |
| 104 | 20 | 20 | 47 | 20 | 20 |
| 105 | 52 | 7 | 11 | 21 | 52 |
| 107 | 21 | 21 | 27 | 21 | 21 |
| 108 | 21 | 21 | 50 | 21 | 21 |
| 109 | 21 | 22 | 27 | 21 | 21 |
| 110 | 22 | 21 | 11 | 22 | 22 |
| 111 | 22 | 57 | 11 | 22 | 22 |
| 112 | 22 | 34 | 11 | 22 | 22 |
| 113 | 25 | 59 | 11 | 22 | 25 |
| 114 | 22 | 22 | 47 | 22 | 22 |
| 115 | 23 | 23 | 47 | 23 | 23 |
| 116 | 23 | 23 | 11 | 23 | 23 |
| 117 | 23 | 23 | 23 | 23 | 23 |
| 118 | 23 | 12 | 59 | 23 | 23 |
| 119 | 29 | 29 | 23 | 23 | 29 |
| 120 | 24 | 24 | 59 | 24 | 24 |
| 121 | 24 | 27 | 50 | 24 | 24 |
| 122 | 24 | 24 | 47 | 24 | 24 |
| 123 | 24 | 24 | 47 | 24 | 24 |
| 124 | 25 | 25 | 25 | 24 | 25 |
| 125 | 25 | 25 | 44 | 24 | 25 |
| 126 | 25 | 25 | 55 | 25 | 25 |
| 127 | 25 | 25 | 11 | 25 | 25 |
| 128 | 25 | 21 | 11 | 25 | 25 |
| 129 | 25 | 26 | 11 | 25 | 25 |
| 130 | 26 | 26 | 47 | 26 | 26 |
| 131 | 26 | 26 | 27 | 26 | 26 |
| 132 | 26 | 26 | 47 | 26 | 26 |
| 133 | 26 | 26 | 47 | 26 | 26 |
| 134 | 26 | 26 | 55 | 26 | 26 |
| 135 | 27 | 27 | 47 | 27 | 27 |
| 136 | 26 | 57 | 11 | 27 | 26 |
| 137 | 27 | 27 | 25 | 27 | 27 |
| 138 | 27 | 27 | 59 | 27 | 27 |
| 139 | 27 | 27 | 47 | 27 | 27 |
| 140 | 28 | 33 | 59 | 28 | 33 |
| 141 | 28 | 28 | 11 | 28 | 29 |
| 142 | 29 | 28 | 11 | 28 | 29 |
| 143 | 29 | 28 | 59 | 28 | 29 |
| 144 | 29 | 28 | 59 | 28 | 29 |
| 145 | 29 | 29 | 29 | 29 | 29 |
| 146 | 29 | 24 | 50 | 29 | 29 |
| 147 | 29 | 29 | 50 | 29 | 29 |
| 148 | 29 | 14 | 47 | 29 | 29 |
| 149 | 29 | 29 | 29 | 29 | 29 |
| 150 | 29 | 30 | 11 | 30 | 30 |
| 151 | 30 | 30 | 47 | 30 | 30 |
| 152 | 30 | 30 | 47 | 30 | 30 |
| 153 | 30 | 34 | 16 | 30 | 30 |
| 154 | 30 | 30 | 47 | 30 | 30 |
| 155 | 31 | 31 | 55 | 31 | 31 |
| 156 | 31 | 31 | 27 | 31 | 31 |
| 157 | 31 | 31 | 47 | 31 | 31 |
| 158 | 29 | 25 | 29 | 31 | 29 |
| 159 | 32 | 29 | 50 | 31 | 29 |
| 160 | 32 | 32 | 29 | 32 | 32 |
| 161 | 32 | 32 | 29 | 32 | 32 |
| 162 | 32 | 32 | 29 | 32 | 32 |
| 163 | 32 | 32 | 27 | 32 | 32 |
| 164 | 32 | 32 | 55 | 32 | 32 |
| 165 | 33 | 33 | 59 | 33 | 33 |
| 166 | 33 | 33 | 11 | 33 | 33 |
| 167 | 33 | 33 | 55 | 33 | 33 |
| 168 | 33 | 53 | 47 | 33 | 33 |
| 169 | 33 | 33 | 47 | 33 | 33 |
| 170 | 33 | 27 | 8 | 34 | 33 |
| 171 | 9 | 34 | 54 | 34 | 9 |
| 172 | 17 | 59 | 8 | 34 | 17 |
| 173 | 33 | 57 | 14 | 34 | 33 |
| 174 | 34 | 34 | 29 | 34 | 11 |
| 175 | 59 | 48 | 11 | 35 | 35 |
| 176 | 35 | 35 | 47 | 35 | 35 |
| 177 | 35 | 35 | 11 | 35 | 35 |
| 178 | 35 | 35 | 47 | 35 | 35 |
| 179 | 35 | 35 | 47 | 35 | 35 |
| 180 | 36 | 36 | 59 | 36 | 36 |
| 181 | 36 | 36 | 59 | 36 | 36 |
| 182 | 55 | 55 | 11 | 36 | 55 |
| 183 | 36 | 36 | 47 | 36 | 36 |
| 184 | 36 | 36 | 47 | 36 | 36 |
| 185 | 39 | 40 | 11 | 37 | 39 |
| 186 | 25 | 17 | 44 | 37 | 25 |
| 187 | 29 | 62 | 11 | 37 | 29 |
| 188 | 37 | 30 | 27 | 37 | 30 |
| 189 | 37 | 52 | 44 | 37 | 52 |
| 190 | 38 | 22 | 11 | 38 | 11 |
| 191 | 38 | 16 | 11 | 38 | 16 |
| 192 | 38 | 59 | 11 | 38 | 11 |
| 193 | 8 | 55 | 50 | 38 | 8 |
| 194 | 38 | 12 | 25 | 38 | 12 |
| 195 | 14 | 36 | 9 | 39 | 36 |
| 196 | 39 | 39 | 27 | 39 | 39 |
| 197 | 8 | 9 | 39 | 39 | 8 |
| 198 | 8 | 14 | 11 | 39 | 8 |
| 199 | 8 | 8 | 27 | 39 | 8 |
| 200 | 40 | 52 | 36 | 40 | 52 |
| 201 | 40 | 53 | 36 | 40 | 40 |
| 202 | 40 | 40 | 36 | 40 | 40 |
| 203 | 43 | 43 | 47 | 40 | 63 |
| 204 | 43 | 43 | 47 | 40 | 63 |
| 205 | 41 | 41 | 11 | 41 | 41 |
| 206 | 41 | 52 | 11 | 41 | 41 |
| 207 | 41 | 41 | 27 | 41 | 41 |
| 208 | 41 | 41 | 47 | 41 | 41 |
| 209 | 41 | 41 | 11 | 41 | 41 |
| 210 | 42 | 20 | 11 | 42 | 42 |
| 211 | 42 | 42 | 55 | 42 | 42 |
| 212 | 37 | 38 | 8 | 42 | 8 |
| 213 | 42 | 42 | 47 | 42 | 42 |
| 214 | 42 | 56 | 11 | 42 | 42 |
| 215 | 60 | 60 | 59 | 43 | 60 |
| 216 | 60 | 21 | 55 | 43 | 21 |
| 217 | 60 | 47 | 47 | 43 | 60 |
| 218 | 60 | 60 | 48 | 43 | 60 |
| 219 | 60 | 52 | 48 | 43 | 60 |
| 220 | 44 | 44 | 27 | 44 | 44 |

| | | | | |
|---|---|---|---|---|
| 217 | 60 | 47 | 47 | 43 | 60 |
| 218 | 60 | 60 | 48 | 43 | 60 |
| 219 | 60 | 52 | 48 | 43 | 60 |
| 220 | 44 | 44 | 27 | 44 | 44 |
| 221 | 44 | 44 | 47 | 44 | 44 |
| 222 | 44 | 44 | 25 | 44 | 44 |
| 223 | 56 | 56 | 59 | 44 | 56 |
| 224 | 11 | 15 | 55 | 44 | 15 |
| 225 | 45 | 16 | 47 | 45 | 45 |
| 226 | 45 | 48 | 27 | 45 | 45 |
| 227 | 45 | 45 | 47 | 45 | 45 |
| 228 | 45 | 45 | 47 | 45 | 45 |
| 229 | 45 | 45 | 11 | 45 | 45 |
| 230 | 46 | 46 | 39 | 46 | 46 |
| 231 | 46 | 46 | 59 | 46 | 46 |
| 232 | 37 | 46 | 59 | 46 | 46 |
| 233 | 46 | 46 | 47 | 46 | 46 |
| 234 | 46 | 46 | 11 | 46 | 46 |
| 235 | 48 | 8 | 55 | 47 | 8 |
| 236 | 47 | 49 | 8 | 47 | 49 |
| 237 | 51 | 49 | 59 | 47 | 51 |
| 238 | 27 | 48 | 59 | 47 | 27 |
| 239 | 48 | 59 | 55 | 47 | 55 |
| 240 | 48 | 16 | 55 | 48 | 16 |
| 241 | 48 | 48 | 59 | 48 | 48 |
| 242 | 48 | 18 | 25 | 48 | 48 |
| 243 | 48 | 60 | 29 | 48 | 60 |
| 244 | 52 | 11 | 59 | 48 | 52 |
| 245 | 49 | 49 | 47 | 49 | 49 |
| 246 | 55 | 21 | 11 | 49 | 21 |
| 247 | 11 | 11 | 55 | 49 | 11 |
| 248 | 11 | 14 | 59 | 49 | 11 |
| 249 | 11 | 15 | 11 | 49 | 11 |
| 250 | 50 | 50 | 47 | 50 | 50 |
| 251 | 48 | 9 | 11 | 50 | 11 |
| 252 | 50 | 59 | 29 | 50 | 50 |
| 253 | 50 | 14 | 55 | 50 | 50 |
| 254 | 14 | 14 | 11 | 50 | 14 |
| 255 | 51 | 51 | 47 | 51 | 51 |
| 256 | 51 | 51 | 8 | 51 | 51 |
| 257 | 55 | 19 | 11 | 51 | 19 |
| 258 | 51 | 20 | 11 | 51 | 51 |
| 259 | 51 | 51 | 50 | 51 | 51 |
| 260 | 52 | 62 | 11 | 52 | 52 |
| 261 | 62 | 62 | 11 | 52 | 62 |
| 262 | 62 | 14 | 11 | 52 | 11 |
| 263 | 50 | 14 | 11 | 52 | 50 |
| 264 | 60 | 17 | 59 | 52 | 60 |
| 265 | 53 | 53 | 47 | 53 | 53 |
| 266 | 53 | 52 | 14 | 53 | 53 |
| 267 | 54 | 8 | 39 | 53 | 54 |
| 268 | 53 | 8 | 9 | 53 | 8 |
| 269 | 59 | 59 | 9 | 53 | 63 |
| 270 | 60 | 8 | 23 | 54 | 60 |
| 271 | 60 | 46 | 39 | 54 | 60 |
| 272 | 60 | 8 | 8 | 54 | 8 |
| 273 | 59 | 8 | 33 | 54 | 8 |
| 274 | 60 | 59 | 23 | 54 | 60 |
| 275 | 55 | 56 | 8 | 55 | 55 |
| 276 | 55 | 10 | 39 | 55 | 55 |
| 277 | 55 | 29 | 47 | 55 | 29 |
| 278 | 55 | 10 | 55 | 55 | 55 |
| 279 | 58 | 14 | 55 | 55 | 58 |
| 280 | 56 | 42 | 11 | 56 | 42 |
| 281 | 56 | 36 | 27 | 56 | 36 |
| 282 | 56 | 56 | 50 | 56 | 56 |
| 283 | 56 | 56 | 55 | 56 | 56 |
| 284 | 56 | 56 | 44 | 56 | 56 |
| 285 | 8 | 8 | 11 | 57 | 8 |
| 286 | 55 | 55 | 11 | 57 | 55 |
| 287 | 14 | 15 | 55 | 57 | 15 |
| 288 | 27 | 17 | 25 | 57 | 27 |
| 289 | 59 | 17 | 50 | 57 | 17 |
| 290 | 35 | 8 | 59 | 58 | 35 |
| 291 | 11 | 27 | 59 | 58 | 11 |
| 292 | 58 | 43 | 47 | 58 | 58 |
| 293 | 35 | 36 | 22 | 58 | 36 |
| 294 | 14 | 36 | 59 | 58 | 36 |
| 295 | 51 | 59 | 59 | 59 | 51 |
| 296 | 35 | 35 | 47 | 59 | 35 |
| 297 | 35 | 35 | 59 | 59 | 35 |
| 298 | 35 | 35 | 59 | 59 | 35 |
| 299 | 35 | 35 | 59 | 59 | 35 |
| 300 | 60 | 60 | 48 | 60 | 60 |
| 301 | 60 | 60 | 48 | 60 | 60 |
| 302 | 60 | 60 | 47 | 60 | 60 |
| 303 | 60 | 58 | 47 | 60 | 60 |
| 304 | 60 | 60 | 48 | 60 | 60 |
| 305 | 61 | 61 | 59 | 61 | 61 |
| 306 | 61 | 61 | 27 | 61 | 61 |
| 307 | 61 | 61 | 47 | 61 | 61 |
| 308 | 59 | 61 | 59 | 61 | 61 |
| 309 | 59 | 56 | 59 | 61 | 56 |
| 310 | 48 | 3 | 22 | 62 | 3 |
| 311 | 27 | 42 | 55 | 62 | 42 |
| 312 | 52 | 59 | 11 | 62 | 52 |
| 313 | 52 | 49 | 55 | 62 | 49 |
| 314 | 61 | 35 | 29 | 62 | 61 |
| 315 | 63 | 63 | 47 | 63 | 63 |
| 316 | 63 | 48 | 47 | 63 | 63 |
| 317 | 63 | 63 | 11 | 63 | 63 |
| 318 | 63 | 48 | 47 | 63 | 63 |
| 319 | 63 | 63 | 47 | 63 | 63 |

**Figure 4.17: Summary of Hypothesis #4.**

### 4.4.6   Conclusion from Hypothesis #4

Samples from class #28 are getting misclassified as class #29 (4/5 times). Similarly, class #43, #54, #59 is getting misclassified in to class #60, #60, #35 (4/5 times). The visual comparison is shown below:

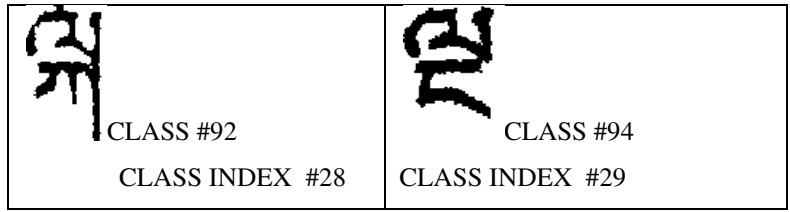

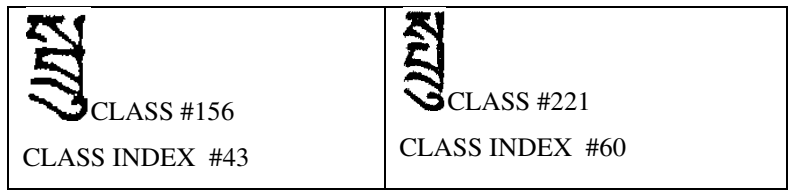

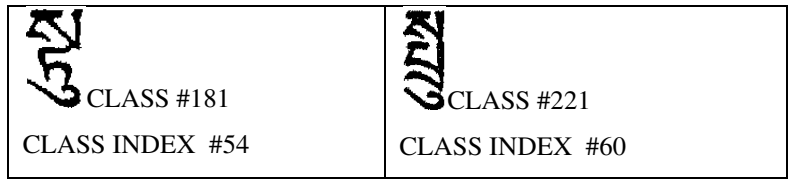

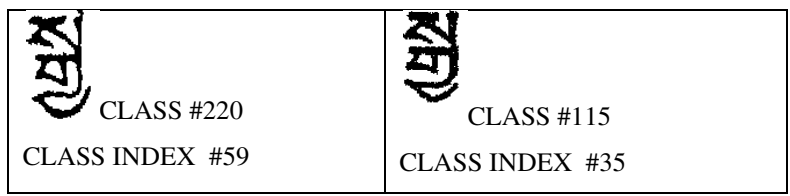

The class indices #10, #13, #14, #18, #28, #54, #62 cannot be correctly classified by the hypothesis as most of the samples are being misclassified by the individual classifiers as well.

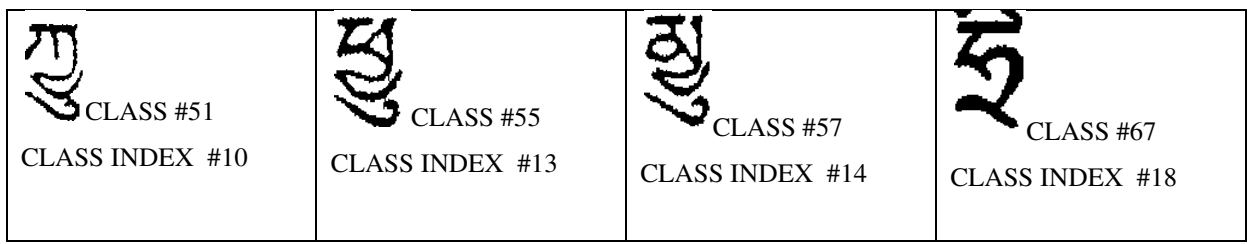

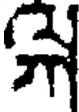CLASS #92

CLASS INDEX #28

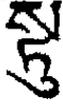CLASS #181

CLASS INDEX #54

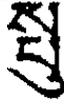CLASS #224

CLASS INDEX #62

Again, for this hypothesis to work, those samples will be considered which has been correctly classified by at least one of the three classifiers. Number of such samples are 227. Also the number of correctly classified samples by the hypothesis is 190. So the accuracy amounts to 83.7 %.

**ACCURACY = 83.7 %**

The samples discarded are: 4, 18, 44, 49, 50, 51, 53, 54, 65, 66, 67, 68, 69, 74, 76, 87, 90, 91, 92, 93, 94, 100, 103, 105, 113, 124, 136, 158, 159, 170, 172, 173, 175, 182, 185, 186, 187, 192, 193, 195, 203, 204, 212, 215, 216, 217, 218, 219, 223, 224, 235, 237, 238, 239, 244, 246, 247, 248, 249, 251, 254, 257, 261, 262, 263, 264, 267, 269, 270, 271, 272, 273, 274, 285, 286, 287, 288, 289, 290, 291, 293, 294, 295, 296, 297, 298, 299, 309, 310, 311, 312, 313, 314.

## 4.5 Hypothesis #5

| SAMPLE NUMBER | CLASSIFIER-1 | CLASSIFIER-2 | CLASSIFIER-3 | ACTUAL-CLASS | PREDICTED CLASS | CONF #1 | CONF #2 | CONF #3 | RESULT |
|---|---|---|---|---|---|---|---|---|---|
| 0 | 0 | 0 | 47 | 0 | 0 | 0.888863 | 0.637107 | 0.171695 | 1.525977 |
| 1 | 0 | 0 | 47 | 0 | 0 | 0.732068 | 0.433350 | 0.124467 | 1.185408 |
| 2 | 0 | 0 | 47 | 0 | 0 | 0.392043 | 0.531262 | 0.121453 | 0.923395 |
| 3 | 0 | 0 | 47 | 0 | 0 | 0.544335 | 0.441504 | 0.142546 | 0.985835 |
| 4 | 8 | 63 | 39 | 0 | 63 | 0.773432 | 1.322408 | 0.262587 | 1.322408 |
| 5 | 1 | 1 | 11 | 1 | 1 | 0.957078 | 0.645170 | 0.243587 | 1.602248 |
| 6 | 1 | 1 | 25 | 1 | 1 | 0.405631 | 0.381659 | 0.144678 | 0.787350 |
| 7 | 1 | 1 | 39 | 1 | 1 | 1.022778 | 0.396176 | 0.230033 | 1.418954 |
| 8 | 1 | 1 | 47 | 1 | 1 | 0.685798 | 0.636064 | 0.179499 | 1.321862 |
| 9 | 1 | 1 | 11 | 1 | 1 | 0.808550 | 0.896328 | 0.239506 | 1.704878 |
| 10 | 2 | 2 | 47 | 2 | 2 | 0.739186 | 0.570180 | 0.181784 | 1.300366 |
| 11 | 2 | 2 | 47 | 2 | 2 | 0.801762 | 0.738201 | 0.150050 | 1.539963 |
| 12 | 2 | 2 | 47 | 2 | 2 | 0.840166 | 0.610413 | 0.132387 | 1.450578 |
| 13 | 2 | 2 | 47 | 2 | 2 | 0.707421 | 0.676642 | 0.131910 | 1.384063 |
| 14 | 2 | 2 | 47 | 2 | 2 | 0.411753 | 0.270825 | 0.131698 | 0.682578 |
| 15 | 3 | 3 | 47 | 3 | 3 | 0.732296 | 0.057030 | 0.208450 | 1.789326 |
| 16 | 3 | 3 | 16 | 3 | 3 | 0.707101 | 0.568526 | 0.307486 | 1.275627 |
| 17 | 3 | 3 | 47 | 3 | 3 | 0.622939 | 0.325186 | 0.331635 | 0.948125 |
| 18 | 3 | 3 | 16 | 3 | 3 | 0.818351 | 0.706397 | 0.384155 | 1.524748 |
| 19 | 27 | 27 | 50 | 3 | 27 | 0.609864 | 0.769293 | 0.161650 | 0.769293 |
| 20 | 23 | 4 | 23 | 3 | 23 | 0.995230 | 0.723881 | 0.249595 | 1.244825 |
| 21 | 4 | 4 | 47 | 4 | 4 | 0.974585 | 1.032225 | 0.261381 | 2.006810 |
| 22 | 4 | 4 | 47 | 4 | 4 | 1.443812 | 1.364342 | 0.311438 | 2.808154 |
| 23 | 4 | 4 | 47 | 4 | 4 | 0.679324 | 0.807143 | 0.232368 | 1.486467 |
| 24 | 4 | 4 | 11 | 4 | 4 | 0.878283 | 1.029396 | 0.274152 | 1.907680 |
| 25 | 5 | 5 | 47 | 5 | 5 | 0.609112 | 0.713090 | 0.203244 | 1.322202 |
| 26 | 5 | 5 | 59 | 5 | 5 | 1.180949 | 1.307183 | 0.393516 | 2.488131 |
| 27 | 5 | 5 | 47 | 5 | 5 | 0.807275 | 1.185180 | 0.278104 | 1.992463 |
| 28 | 5 | 5 | 47 | 5 | 5 | 0.495483 | 0.569720 | 0.175756 | 1.065203 |
| 29 | 5 | 5 | 47 | 5 | 5 | 0.531964 | 0.339493 | 0.149362 | 0.671458 |
| 30 | 5 | 8 | 55 | 5 | 5 | 0.400585 | 0.377558 | 0.208875 | 0.400585 |
| 31 | 6 | 6 | 11 | 6 | 6 | 0.418615 | 0.297931 | 0.230072 | 0.418615 |
| 32 | 6 | 23 | 6 | 6 | 6 | 0.484617 | 0.298678 | 0.220696 | 0.705313 |
| 33 | 6 | 6 | 55 | 6 | 6 | 0.728366 | 0.343690 | 0.231605 | 1.072045 |
| 34 | 6 | 6 | 11 | 6 | 6 | 0.516336 | 0.776687 | 0.233410 | 1.293063 |
| 35 | 7 | 7 | 47 | 7 | 7 | 0.583816 | 0.567355 | 0.165172 | 1.151171 |
| 36 | 7 | 7 | 47 | 7 | 7 | 0.644844 | 0.583824 | 0.183578 | 1.228668 |
| 37 | 7 | 7 | 47 | 7 | 7 | 0.849235 | 0.927452 | 0.209593 | 1.776677 |
| 38 | 7 | 0 | 47 | 7 | 7 | 0.406411 | 0.522099 | 0.136917 | 0.522099 |
| 39 | 7 | 7 | 47 | 7 | 7 | 0.653851 | 1.928136 | 0.163658 | 2.581988 |
| 40 | 7 | 8 | 11 | 7 | 8 | 0.662252 | 1.027991 | 0.322920 | 1.690243 |
| 41 | 8 | 8 | 8 | 8 | 8 | 0.587417 | 0.476633 | 0.360658 | 1.064051 |
| 42 | 8 | 8 | 11 | 8 | 8 | 0.671938 | 0.326958 | 0.296299 | 0.997996 |
| 43 | 8 | 8 | 11 | 8 | 8 | 0.811858 | 0.835778 | 0.368405 | 1.205182 |
| 44 | 23 | 8 | 11 | 8 | 23 | 0.810629 | 1.376313 | 0.523577 | 1.386341 |
| 45 | 9 | 9 | 39 | 9 | 9 | 0.562565 | 0.897002 | 0.320669 | 1.459567 |
| 46 | 9 | 9 | 59 | 9 | 9 | 0.686528 | 0.857179 | 0.401759 | 0.857179 |
| 47 | 9 | 8 | 11 | 9 | 9 | 0.665379 | 0.611283 | 0.388043 | 0.665379 |
| 48 | 9 | 9 | 11 | 9 | 9 | 0.501963 | 0.384243 | 0.275407 | 0.501963 |
| 49 | 8 | 8 | 11 | 9 | 9 | 0.557126 | 0.243405 | 0.288332 | 0.800531 |
| 50 | 12 | 11 | 39 | 10 | 12 | 0.395974 | 0.248445 | 0.148724 | 0.395974 |
| 51 | 11 | 11 | 62 | 10 | 10 | 0.544674 | 0.281648 | 0.189016 | 0.544674 |
| 52 | 10 | 27 | 55 | 10 | 10 | 0.387350 | 0.238678 | 0.157942 | 0.387350 |
| 53 | 55 | 14 | 11 | 10 | 10 | 0.461775 | 0.474786 | 0.199733 | 0.474786 |
| 54 | 12 | 3 | 55 | 10 | 12 | 0.488611 | 0.179383 | 0.198381 | 0.488611 |
| 55 | 11 | 11 | 11 | 11 | 11 | 0.771839 | 1.069617 | 0.372694 | 2.214150 |
| 56 | 11 | 11 | 11 | 11 | 11 | 0.715475 | 0.560025 | 0.442762 | 1.718263 |
| 57 | 11 | 11 | 11 | 11 | 11 | 0.833368 | 0.341839 | 0.398051 | 2.174059 |
| 58 | 45 | 42 | 11 | 11 | 45 | 0.772931 | 0.700544 | 0.462354 | 0.772931 |
| 59 | 42 | 59 | 11 | 11 | 45 | 0.718770 | 0.475419 | 0.364407 | 0.718770 |
| 60 | 12 | 12 | 47 | 12 | 12 | 0.508189 | 0.433174 | 0.216165 | 0.941363 |
| 61 | 12 | 12 | 44 | 12 | 12 | 1.097313 | 1.113849 | 0.395977 | 2.211161 |
| 62 | 12 | 33 | 50 | 12 | 12 | 0.790581 | 0.612741 | 0.238313 | 0.790581 |
| 63 | 12 | 12 | 47 | 12 | 12 | 0.451545 | 0.224336 | 0.184357 | 0.675881 |
| 64 | 12 | 12 | 47 | 12 | 12 | 0.921817 | 0.976065 | 0.389033 | 0.976065 |
| 65 | 55 | 14 | 14 | 13 | 14 | 0.781565 | 0.785952 | 0.261321 | 1.047273 |
| 66 | 14 | 14 | 11 | 13 | 13 | 0.687200 | 0.743952 | 0.209324 | 1.431052 |
| 67 | 56 | 55 | 47 | 13 | 55 | 0.889681 | 0.681023 | 0.262584 | 0.943607 |
| 68 | 14 | 8 | 55 | 13 | 13 | 0.979814 | 0.738203 | 0.248841 | 0.979814 |
| 69 | 14 | 14 | 11 | 13 | 14 | 0.349728 | 0.177368 | 0.121921 | 0.349728 |
| 70 | 8 | 14 | 6 | 14 | 14 | 0.767546 | 0.753867 | 0.303342 | 1.521412 |
| 71 | 14 | 61 | 11 | 14 | 61 | 0.611929 | 0.770300 | 0.331732 | 0.770300 |
| 72 | 14 | 36 | 11 | 14 | 14 | 0.876333 | 0.668509 | 0.329079 | 0.876333 |
| 73 | 14 | 36 | 11 | 14 | 14 | 0.782317 | 0.749982 | 0.312377 | 0.782317 |
| 74 | 28 | 11 | 55 | 14 | 14 | 1.018330 | 1.568976 | 0.406266 | 1.568976 |
| 75 | 15 | 26 | 48 | 15 | 15 | 1.053562 | 0.883649 | 0.308986 | 1.053562 |
| 76 | 15 | 27 | 11 | 15 | 27 | 1.059836 | 1.410218 | 0.331562 | 1.410218 |
| 77 | 15 | 15 | 29 | 15 | 15 | 0.903777 | 0.851842 | 0.283836 | 1.755619 |
| 78 | 15 | 21 | 44 | 15 | 15 | 1.095045 | 0.655388 | 0.458845 | 1.095045 |
| 79 | 15 | 15 | 47 | 15 | 15 | 0.390250 | 1.585348 | 0.449589 | 2.575608 |
| 80 | 16 | 16 | 47 | 16 | 16 | 0.599544 | 0.893761 | 0.375264 | 1.493395 |
| 81 | 16 | 16 | 18 | 16 | 16 | 0.998946 | 0.725521 | 0.434645 | 2.217111 |
| 82 | 16 | 16 | 27 | 16 | 16 | 0.458834 | 0.314792 | 0.218419 | 1.773626 |
| 83 | 16 | 16 | 58 | 16 | 16 | 0.984439 | 0.847638 | 0.398339 | 2.240416 |
| 84 | 16 | 16 | 58 | 16 | 16 | 1.073611 | 0.611905 | 0.617684 | 1.605586 |
| 85 | 17 | 17 | 11 | 17 | 17 | 0.729480 | 0.546507 | 0.334917 | 0.729480 |
| 86 | 17 | 15 | 39 | 17 | 17 | 0.596662 | 0.758639 | 0.284171 | 0.758639 |
| 87 | 29 | 15 | 11 | 17 | 17 | 0.568780 | 0.837130 | 0.256414 | 0.837130 |
| 88 | 17 | 17 | 47 | 17 | 17 | 0.876333 | 0.321414 | 0.274611 | 0.728993 |
| 89 | 17 | 17 | 55 | 17 | 17 | 0.452579 | 0.582469 | 0.257097 | 0.582469 |
| 90 | 23 | 44 | 23 | 18 | 18 | 1.149720 | 1.241766 | 0.237241 | 1.386362 |
| 91 | 23 | 23 | 11 | 18 | 18 | 1.435355 | 0.250586 | 0.156284 | 0.591639 |
| 92 | 6 | 57 | 29 | 18 | 57 | 0.322889 | 0.343870 | 0.159199 | 0.343870 |
| 93 | 4 | 57 | 11 | 18 | 4 | 0.366793 | 0.165441 | 0.144549 | 0.366793 |
| 94 | 63 | 23 | 11 | 18 | 63 | 0.395136 | 0.170896 | 0.161536 | 0.516672 |
| 95 | 15 | 17 | 22 | 19 | 19 | 0.753887 | 0.595681 | 0.243643 | 0.753887 |
| 96 | 19 | 19 | 11 | 19 | 19 | 0.871575 | 0.899306 | 0.260222 | 1.770082 |
| 97 | 19 | 19 | 25 | 19 | 19 | 0.600024 | 0.747229 | 0.281902 | 1.347253 |
| 98 | 19 | 19 | 55 | 19 | 19 | 0.873542 | 1.568465 | 0.514875 | 2.442008 |
| 99 | 19 | 19 | 47 | 19 | 19 | 0.339249 | 0.907869 | 0.377939 | 1.847118 |
| 100 | 19 | 48 | 59 | 20 | 19 | 0.449187 | 0.823994 | 0.184955 | 0.823994 |
| 101 | 20 | 20 | 47 | 20 | 20 | 1.075226 | 0.729183 | 0.229855 | 1.804409 |
| 102 | 20 | 20 | 47 | 20 | 20 | 0.364011 | 0.311068 | 0.145879 | 0.675079 |
| 103 | 20 | 48 | 29 | 20 | 20 | 0.527493 | 0.426480 | 0.175306 | 0.527493 |
| 104 | 20 | 20 | 47 | 20 | 20 | 0.729740 | 1.300147 | 0.209639 | 2.037886 |
| 105 | 52 | 7 | 11 | 21 | 52 | 0.341128 | 0.486231 | 0.174591 | 0.341128 |
| 106 | 21 | 21 | 59 | 21 | 21 | 0.963865 | 0.632236 | 0.220072 | 1.596101 |
| 107 | 21 | 21 | 21 | 21 | 21 | 0.724855 | 0.998748 | 0.232091 | 1.723043 |
| 108 | 21 | 21 | 50 | 21 | 21 | 0.387727 | 1.306366 | 0.376204 | 2.294693 |
| 109 | 21 | 21 | 27 | 21 | 21 | 1.006649 | 1.250812 | 0.368923 | 1.250812 |
| 110 | 22 | 21 | 11 | 22 | 21 | 0.923924 | 1.223311 | 0.262639 | 1.223311 |
| 111 | 22 | 57 | 11 | 22 | 22 | 0.714843 | 0.535463 | 0.251242 | 0.714843 |
| 112 | 22 | 34 | 11 | 22 | 22 | 0.751003 | 0.689488 | 0.327526 | 0.751003 |
| 113 | 22 | 59 | 11 | 22 | 25 | 0.896720 | 0.827102 | 0.249674 | 0.896720 |
| 114 | 23 | 22 | 47 | 22 | 22 | 0.769491 | 1.596452 | 0.242628 | 2.355944 |
| 115 | 23 | 47 | 23 | 23 | 23 | 1.058850 | 1.894705 | 0.441190 | 2.453395 |
| 116 | 23 | 23 | 11 | 23 | 23 | 1.010653 | 1.558219 | 0.584496 | 2.568872 |
| 117 | 23 | 23 | 23 | 23 | 23 | 0.579518 | 1.771941 | 0.256329 | 2.607788 |
| 118 | 23 | 12 | 59 | 23 | 12 | 1.051087 | 1.066647 | 0.360121 | 1.066647 |
| 119 | 23 | 29 | 29 | 23 | 23 | 0.918531 | 1.851813 | 0.575564 | 2.770344 |
| 120 | 24 | 24 | 59 | 24 | 24 | 0.624289 | 1.112579 | 0.159664 | 1.736868 |
| 121 | 24 | 50 | 50 | 24 | 24 | 1.295086 | 0.567199 | 0.244355 | 1.295086 |
| 122 | 24 | 24 | 47 | 24 | 24 | 0.969695 | 0.865237 | 0.525253 | 1.834921 |
| 123 | 24 | 24 | 47 | 24 | 24 | 1.008576 | 0.598846 | 0.333944 | 1.607222 |
| 124 | 24 | 25 | 25 | 24 | 24 | 0.582080 | 0.767248 | 0.151067 | 1.500395 |
| 125 | 25 | 25 | 44 | 25 | 25 | 0.788100 | 0.506200 | 0.339583 | 1.091300 |
| 126 | 25 | 25 | 55 | 25 | 25 | 0.772743 | 0.336342 | 0.408757 | 1.109085 |
| 127 | 25 | 25 | 11 | 25 | 25 | 0.794630 | 0.445793 | 0.361079 | 1.240422 |
| 128 | 25 | 21 | 11 | 25 | 25 | 0.678032 | 0.342438 | 0.338142 | 0.678032 |
| 129 | 25 | 25 | 11 | 25 | 25 | 0.811597 | 0.476830 | 0.385773 | 1.288427 |
| 130 | 26 | 26 | 47 | 26 | 26 | 0.878070 | 1.164791 | 0.334026 | 2.042862 |
| 131 | 26 | 26 | 27 | 26 | 26 | 0.581115 | 0.712109 | 0.303492 | 1.293223 |
| 132 | 26 | 26 | 47 | 26 | 26 | 0.564640 | 0.580188 | 0.279876 | 1.124808 |
| 133 | 26 | 26 | 47 | 26 | 26 | 0.491062 | 0.554272 | 0.267934 | 1.045334 |
| 134 | 26 | 26 | 55 | 26 | 26 | 1.109141 | 1.454799 | 0.510254 | 2.563940 |
| 135 | 27 | 27 | 47 | 27 | 27 | 0.635340 | 0.535171 | 0.363869 | 1.174511 |
| 136 | 26 | 57 | 11 | 27 | 27 | 0.480109 | 0.392612 | 0.219566 | 0.480109 |
| 137 | 27 | 27 | 25 | 27 | 27 | 0.511270 | 0.727604 | 0.242788 | 1.238874 |
| 138 | 27 | 27 | 59 | 27 | 27 | 0.423937 | 0.415955 | 0.216082 | 0.833652 |
| 139 | 27 | 27 | 47 | 27 | 27 | 0.584295 | 0.368521 | 0.302400 | 0.952816 |
| 140 | 28 | 33 | 59 | 28 | 28 | 0.477765 | 0.535805 | 0.249788 | 0.535805 |
| 141 | 29 | 29 | 11 | 28 | 28 | 0.589992 | 0.942420 | 0.315450 | 0.942420 |
| 142 | 29 | 29 | 11 | 28 | 28 | 0.289364 | 0.598627 | 0.168899 | 0.598627 |
| 143 | 29 | 29 | 59 | 28 | 28 | 0.550003 | 0.754051 | 0.320198 | 0.754051 |
| 144 | 29 | 29 | 11 | 28 | 28 | 0.694473 | 1.354999 | 0.345018 | 1.354999 |
| 145 | 29 | 29 | 29 | 28 | 29 | 1.058193 | 0.499064 | 0.361700 | 1.918957 |
| 146 | 29 | 24 | 50 | 29 | 24 | 0.835341 | 0.936053 | 0.318002 | 0.936053 |
| 147 | 29 | 50 | 29 | 29 | 29 | 1.089883 | 1.152738 | 0.375110 | 2.242621 |
| 148 | 29 | 29 | 14 | 29 | 14 | 0.509583 | 0.523034 | 0.255111 | 0.523034 |
| 149 | 29 | 29 | 47 | 29 | 29 | 0.627556 | 0.498676 | 0.298751 | 1.424983 |
| 150 | 30 | 30 | 11 | 30 | 30 | 0.549357 | 0.878454 | 0.189281 | 1.427811 |
| 151 | 30 | 30 | 47 | 30 | 30 | 0.993237 | 0.736562 | 0.272762 | 1.729799 |
| 152 | 30 | 30 | 47 | 30 | 30 | 0.440602 | 0.684472 | 0.187432 | 1.125074 |
| 153 | 30 | 30 | 47 | 30 | 30 | 0.493889 | 0.497358 | 0.212616 | 0.497358 |
| 154 | 30 | 30 | 47 | 30 | 30 | 0.864505 | 0.590355 | 0.265275 | 1.854860 |
| 155 | 31 | 31 | 55 | 31 | 31 | 0.719533 | 0.528208 | 0.211647 | 1.247741 |
| 156 | 31 | 31 | 27 | 31 | 31 | 0.496153 | 0.523289 | 0.166233 | 1.019442 |
| 157 | 31 | 31 | 47 | 31 | 31 | 0.606708 | 0.506821 | 0.169214 | 1.113529 |
| 158 | 29 | 31 | 29 | 31 | 29 | 0.713547 | 0.490901 | 0.177584 | 0.891131 |
| 159 | 29 | 29 | 50 | 31 | 31 | 0.599639 | 0.982862 | 0.177018 | 1.582500 |
| 160 | 32 | 32 | 29 | 32 | 32 | 0.428955 | 0.276244 | 0.156214 | 0.704299 |
| 161 | 32 | 32 | 29 | 32 | 32 | 0.637326 | 0.559264 | 0.172831 | 1.196590 |
| 162 | 32 | 32 | 29 | 32 | 32 | 0.327877 | 0.260778 | 0.167976 | 0.588656 |
| 163 | 32 | 32 | 27 | 32 | 32 | 0.438128 | 0.304566 | 0.153272 | 0.742634 |
| 164 | 32 | 32 | 55 | 32 | 32 | 0.843104 | 1.462538 | 0.209672 | 2.305642 |
| 165 | 33 | 33 | 59 | 33 | 33 | 0.593956 | 0.532287 | 0.197301 | 1.126243 |
| 166 | 33 | 33 | 11 | 33 | 33 | 0.549761 | 0.396033 | 0.212462 | 0.345734 |
| 167 | 33 | 33 | 55 | 33 | 33 | 0.594302 | 0.647315 | 0.234979 | 1.241617 |
| 168 | 33 | 33 | 53 | 33 | 33 | 0.774398 | 0.652037 | 0.275881 | 0.774398 |
| 169 | 33 | 33 | 47 | 33 | 33 | 0.758837 | 0.499525 | 0.250278 | 1.258362 |
| 170 | 33 | 27 | 8 | 33 | 34 | 0.791364 | 0.816106 | 0.221390 | 0.816106 |
| 171 | 9 | 27 | 34 | 54 | 27 | 0.289854 | 0.182649 | 0.063895 | 0.289854 |
| 172 | 17 | 17 | 59 | 8 | 17 | 0.503609 | 0.331856 | 0.163250 | 0.503609 |
| 173 | 33 | 57 | 14 | 14 | 57 | 0.585462 | 0.684229 | 0.225239 | 0.684229 |
| 174 | 34 | 57 | 54 | 29 | 34 | 0.372474 | 0.345307 | 0.122210 | 0.717780 |
| 175 | 35 | 35 | 48 | 11 | 35 | 0.721179 | 1.083170 | 0.259604 | 1.083170 |
| 176 | 35 | 35 | 11 | 47 | 35 | 0.808426 | 0.297622 | 0.254171 | 1.106048 |
| 177 | 35 | 35 | 11 | 35 | 35 | 0.955104 | 1.114844 | 0.444183 | 2.069948 |
| 178 | 35 | 35 | 47 | 35 | 35 | 0.861913 | 1.118060 | 0.289344 | 1.979973 |
| 179 | 35 | 35 | 47 | 35 | 35 | 0.926293 | 0.906858 | 0.273982 | 1.832952 |
| 180 | 36 | 36 | 59 | 35 | 36 | 0.945548 | 0.804400 | 0.333681 | 1.749348 |
| 181 | 36 | 36 | 59 | 36 | 36 | 0.883278 | 0.745941 | 0.362095 | 1.629219 |
| 182 | 36 | 55 | 11 | 36 | 36 | 0.745443 | 1.169599 | 0.317909 | 1.915042 |
| 183 | 36 | 36 | 47 | 36 | 36 | 0.808960 | 1.229152 | 0.418982 | 2.057212 |
| 184 | 36 | 36 | 47 | 36 | 36 | 0.899222 | 1.069343 | 0.405202 | 1.958565 |
| 185 | 39 | 40 | 11 | 37 | 37 | 0.736572 | 0.953723 | 0.405922 | 0.953723 |
| 186 | 25 | 17 | 44 | 37 | 37 | 0.461842 | 0.456881 | 0.192447 | 0.461842 |
| 187 | 39 | 12 | 62 | 11 | 39 | 0.492293 | 0.170134 | 0.174802 | 0.492203 |
| 188 | 37 | 30 | 27 | 37 | 37 | 0.510380 | 0.321643 | 0.226234 | 0.510380 |
| 189 | 37 | 37 | 44 | 37 | 37 | 0.620934 | 0.559506 | 0.226931 | 0.620934 |
| 190 | 38 | 22 | 11 | 38 | 38 | 0.418951 | 0.255600 | 0.186326 | 0.418951 |
| 191 | 38 | 16 | 11 | 38 | 38 | 0.349523 | 0.133088 | 0.166170 | 0.349523 |
| 192 | 38 | 38 | 11 | 38 | 38 | 0.405865 | 0.200430 | 0.164830 | 0.405865 |
| 193 | 9 | 19 | 50 | 38 | 8 | 0.360906 | 0.147204 | 0.172337 | 0.360906 |
| 194 | 38 | 12 | 25 | 38 | 12 | 0.426546 | 0.433564 | 0.202327 | 0.433564 |
| 195 | 14 | 38 | 9 | 39 | 39 | 0.567585 | 0.439249 | 0.278418 | 0.567585 |
| 196 | 39 | 39 | 27 | 39 | 39 | 0.754583 | 0.755017 | 0.379146 | 1.509600 |
| 197 | 8 | 19 | 9 | 39 | 8 | 0.899600 | 1.287096 | 0.425110 | 1.287096 |
| 198 | 8 | 8 | 14 | 39 | 39 | 0.682050 | 0.544741 | 0.291384 | 0.682050 |
| 199 | 8 | 8 | 27 | 33 | 39 | 0.799601 | 0.534686 | 0.341684 | 1.334267 |
| 200 | 40 | 8 | 82 | 40 | 40 | 1.025432 | 0.903501 | 0.261744 | 1.025432 |
| 201 | 40 | 53 | 53 | 40 | 53 | 0.517389 | 0.594626 | 0.314160 | 0.594626 |
| 202 | 40 | 40 | 36 | 40 | 40 | 0.858959 | 0.965762 | 0.262300 | 1.822721 |
| 203 | 40 | 43 | 43 | 40 | 43 | 0.872232 | 0.945779 | 0.275985 | 1.818011 |
| 204 | 43 | 43 | 47 | 40 | 43 | 0.826754 | 0.881380 | 0.299562 | 1.708134 |
| 205 | 41 | 41 | 11 | 44 | 41 | 0.773592 | 0.496132 | 0.148328 | 1.269724 |
| 206 | 41 | 41 | 52 | 41 | 41 | 0.575631 | 0.767252 | 0.185364 | 0.767252 |
| 207 | 41 | 41 | 27 | 41 | 41 | 0.776762 | 0.423839 | 0.135671 | 1.200601 |
| 208 | 41 | 41 | 47 | 41 | 41 | 0.675078 | 0.591790 | 0.152832 | 1.266883 |
| 209 | 41 | 41 | 11 | 41 | 41 | 0.482559 | 0.670683 | 0.159546 | 1.153142 |
| 210 | 42 | 42 | 11 | 42 | 42 | 0.964635 | 1.006400 | 0.444163 | 1.006400 |
| 211 | 42 | 42 | 55 | 42 | 42 | 0.635341 | 0.582349 | 0.243237 | 1.217630 |
| 212 | 37 | 42 | 38 | 8 | 42 | 0.605926 | 0.220757 | 0.184469 | 0.720057 |
| 213 | 42 | 42 | 8 | 47 | 42 | 0.936725 | 1.070240 | 0.329869 | 2.006365 |
| 214 | 42 | 42 | 56 | 11 | 42 | 0.592217 | 0.281885 | 0.213567 | 0.592217 |
| 215 | 60 | 60 | 11 | 42 | 42 | 0.511110 | 0.346307 | 0.153456 | 0.857417 |
| 216 | 60 | 60 | 21 | 43 | 60 | 0.842172 | 0.467813 | 0.230733 | 0.842172 |
| 217 | 60 | 47 | 47 | 43 | 47 | 0.454182 | 0.315711 | 0.216642 | 0.532353 |
| 218 | 60 | 60 | 48 | 43 | 43 | 0.640034 | 0.426965 | 0.228591 | 1.066998 |
| 219 | 60 | 60 | 52 | 48 | 43 | 0.519840 | 0.416511 | 0.185319 | 0.519840 |
| 220 | 44 | 44 | 27 | 44 | 44 | 0.585050 | 0.353174 | 0.301429 | 0.938224 |
| 221 | 44 | 44 | 47 | 44 | 44 | 0.805593 | 0.883701 | 0.368220 | 1.689205 |
| 222 | 44 | 44 | 25 | 44 | 44 | 0.635250 | 0.724029 | 0.328938 | 1.359279 |
| 223 | 44 | 56 | 59 | 44 | 56 | 0.734319 | 0.850441 | 0.384154 | 1.594760 |
| 224 | 11 | 15 | 55 | 44 | 11 | 0.811519 | 0.470609 | 0.540030 | 0.811519 |
| 225 | 45 | 16 | 47 | 45 | 45 | 1.037029 | 0.728701 | 0.451579 | 1.037029 |
| 226 | 45 | 48 | 27 | 45 | 45 | 0.601327 | 0.485211 | 0.238532 | 0.601327 |
| 227 | 45 | 47 | 45 | 45 | 45 | 0.715069 | 0.678169 | 0.364474 | 1.393227 |
| 228 | 45 | 47 | 47 | 45 | 45 | 0.347269 | 0.910842 | 0.452719 | 1.858111 |
| 229 | 45 | 45 | 11 | 45 | 45 | 0.984427 | 1.146123 | 0.332775 | 2.130550 |
| 230 | 46 | 39 | 46 | 46 | 46 | 0.786100 | 1.037099 | 0.296636 | 1.823199 |
| 231 | 46 | 38 | 59 | 46 | 46 | 0.452851 | 0.528257 | 0.219153 | 0.381108 |
| 232 | 37 | 46 | 59 | 46 | 46 | 0.805908 | 0.857272 | 0.267643 | 0.857272 |
| 233 | 46 | 46 | 47 | 46 | 46 | 0.948893 | 0.840165 | 0.259321 | 1.789068 |
| 234 | 46 | 46 | 11 | 46 | 46 | 0.553603 | 0.463540 | 0.202110 | 1.022942 |
| 235 | 48 | 8 | 55 | 47 | 8 | 0.663229 | 1.003147 | 0.280032 | 1.003147 |
| 236 | 47 | 49 | 8 | 47 | 49 | 0.546107 | 0.786794 | 0.298880 | 0.786794 |
| 237 | 51 | 49 | 59 | 47 | 47 | 0.586701 | 0.729504 | 0.305286 | 0.729504 |
| 238 | 27 | 48 | 59 | 47 | 48 | 0.462666 | 0.580429 | 0.243067 | 0.580429 |
| 239 | 48 | 59 | 55 | 47 | 48 | 0.576875 | 0.545322 | 0.235870 | 0.576875 |
| 240 | 48 | 16 | 59 | 47 | 48 | 0.831811 | 1.626834 | 0.321317 | 1.626834 |
| 241 | 16 | 48 | 59 | 48 | 16 | 1.069588 | 0.867332 | 0.375435 | 1.947920 |
| 242 | 48 | 18 | 25 | 48 | 48 | 0.734281 | 0.803905 | 0.290975 | 0.803905 |
| 243 | 18 | 60 | 29 | 48 | 48 | 0.791838 | 1.270221 | 0.389989 | 1.270221 |
| 244 | 52 | 11 | 59 | 48 | 48 | 0.879694 | 0.880245 | 0.350480 | 0.880245 |
| 245 | 49 | 49 | 47 | 49 | 49 | 0.571589 | 0.576817 | 0.246013 | 1.148406 |
| 246 | 55 | 21 | 11 | 49 | 21 | 0.851377 | 0.960891 | 0.258012 | 0.960891 |
| 247 | 11 | 11 | 55 | 49 | 11 | 0.851377 | 1.463474 | 0.323243 | 2.336950 |
| 248 | 11 | 14 | 59 | 49 | 14 | 0.640237 | 0.934815 | 0.246274 | 0.934815 |
| 249 | 11 | 15 | 11 | 49 | 15 | 0.829023 | 1.268116 | 0.408585 | 1.268116 |
| 250 | 50 | 11 | 47 | 50 | 50 | 0.347245 | 0.671307 | 0.293590 | 1.618552 |
| 251 | 48 | 9 | 11 | 50 | 50 | 0.871870 | 0.335133 | 0.293076 | 0.871870 |
| 252 | 50 | 16 | 29 | 50 | 50 | 0.974087 | 0.775903 | 0.340711 | 0.974087 |
| 253 | 50 | 14 | 55 | 50 | 50 | 0.816255 | 0.815518 | 0.301116 | 0.816255 |
| 254 | 14 | 14 | 11 | 50 | 50 | 0.760321 | 0.861127 | 0.306382 | 1.621448 |
| 255 | 51 | 14 | 47 | 51 | 51 | 0.965293 | 1.473124 | 0.385696 | 2.438417 |
| 256 | 51 | 51 | 8 | 51 | 51 | 1.047160 | 1.178636 | 0.415376 | 2.225795 |
| 257 | 55 | 19 | 11 | 51 | 51 | 0.997682 | 1.950022 | 0.307563 | 1.950022 |
| 258 | 51 | 20 | 11 | 51 | 20 | 0.945252 | 1.475967 | 0.263157 | 1.475967 |
| 259 | 51 | 50 | 51 | 51 | 51 | 0.904068 | 0.821425 | 0.254727 | 1.806293 |
| 260 | 52 | 52 | 11 | 51 | 52 | 1.114462 | 1.170144 | 0.337890 | 1.170144 |
| 261 | 62 | 62 | 11 | 11 | 62 | 0.899949 | 0.895830 | 0.347093 | 1.795779 |
| 262 | 62 | 62 | 11 | 62 | 62 | 0.999364 | 0.853567 | 0.521862 | 0.999364 |
| 263 | 50 | 14 | 11 | 52 | 52 | 0.407713 | 0.351810 | 0.197821 | 0.407719 |
| 264 | 60 | 17 | 59 | 52 | 17 | 0.467094 | 0.949336 | 0.274667 | 0.949336 |
| 265 | 53 | 47 | 53 | 52 | 47 | 0.725703 | 0.802878 | 0.290992 | 1.528582 |
| 266 | 53 | 52 | 14 | 53 | 53 | 0.451040 | 0.761376 | 0.207538 | 0.761376 |
| 267 | 54 | 8 | 39 | 53 | 8 | 0.469424 | 0.648763 | 0.219880 | 0.648763 |
| 268 | 53 | 9 | 53 | 53 | 53 | 0.817494 | 0.442173 | 0.244495 | 0.817494 |
| 269 | 59 | 59 | 9 | 53 | 59 | 0.438685 | 0.343088 | 0.137334 | 1.381773 |
| 270 | 60 | 8 | 23 | 54 | 8 | 0.711573 | 0.755879 | 0.237945 | 0.755879 |
| 271 | 60 | 46 | 39 | 54 | 60 | 0.462012 | 0.240845 | 0.202442 | 0.462012 |
| 272 | 60 | 8 | 8 | 54 | 60 | 0.478205 | 0.259113 | 0.215541 | 0.478205 |
| 273 | 59 | 8 | 39 | 54 | 59 | 0.389823 | 0.268330 | 0.207040 | 0.389823 |
| 274 | 60 | 59 | 23 | 54 | 60 | 0.495024 | 0.351387 | 0.210285 | 0.495024 |
| 275 | 60 | 56 | 8 | 55 | 56 | 0.877035 | 1.107597 | 0.372917 | 1.107597 |
| 276 | 55 | 10 | 39 | 55 | 55 | 0.697706 | 0.665032 | 0.309101 | 0.697706 |
| 277 | 55 | 29 | 47 | 55 | 55 | 0.972993 | 0.575819 | 0.420040 | 0.972993 |
| 278 | 55 | 10 | 55 | 55 | 55 | 0.994170 | 0.762327 | 0.461178 | 1.490348 |
| 279 | 58 | 14 | 55 | 55 | 58 | 0.603209 | 0.327126 | 0.354333 | 0.603209 |
| 280 | 56 | 42 | 11 | 56 | 56 | 0.890921 | 0.600428 | 0.308596 | 0.890921 |
| 281 | 56 | 27 | 50 | 56 | 56 | 0.602777 | 0.832587 | 0.313911 | 0.832587 |
| 282 | 56 | 56 | 50 | 56 | 56 | 0.949795 | 0.844934 | 0.388320 | 1.794729 |
| 283 | 56 | 56 | 55 | 56 | 56 | 0.750595 | 0.668299 | 0.343099 | 1.418894 |
| 284 | 56 | 56 | 44 | 56 | 56 | 0.867866 | 0.834162 | 0.384817 | 1.702028 |
| 285 | 8 | 8 | 11 | 57 | 8 | 0.527675 | 0.706077 | 0.277818 | 1.313762 |
| 286 | 55 | 55 | 11 | 57 | 57 | 0.485205 | 0.594839 | 0.394915 | 1.080044 |
| 287 | 14 | 15 | 55 | 57 | 15 | 1.034410 | 1.302910 | 0.420959 | 1.302910 |
| 288 | 27 | 17 | 25 | 57 | 17 | 1.002554 | 1.358199 | 0.303986 | 1.358199 |
| 289 | 59 | 17 | 50 | 57 | 57 | 0.529213 | 0.390703 | 0.243632 | 0.529213 |
| 290 | 35 | 8 | 59 | 58 | 35 | 0.514966 | 0.348888 | 0.273455 | 0.514966 |
| 291 | 11 | 59 | 59 | 58 | 11 | 0.586121 | 0.503782 | 0.255572 | 0.586121 |
| 292 | 58 | 43 | 47 | 58 | 58 | 0.666353 | 0.568037 | 0.254407 | 0.666353 |
| 293 | 35 | 36 | 22 | 58 | 35 | 0.508213 | 0.352421 | 0.235847 | 0.508213 |
| 294 | 14 | 59 | 59 | 58 | 14 | 0.513688 | 0.330428 | 0.258363 | 0.513688 |
| 295 | 51 | 59 | 59 | 58 | 59 | 0.643446 | 0.452768 | 0.365010 | 0.817718 |
| 296 | 35 | 35 | 47 | 59 | 35 | 1.331703 | 1.876630 | 0.487048 | 3.208332 |
| 297 | 35 | 35 | 59 | 59 | 35 | 1.140376 | 1.018409 | 0.389885 | 2.158785 |
| 298 | 35 | 35 | 59 | 59 | 35 | 1.758301 | 2.138520 | 0.562658 | 3.896821 |
| 299 | 35 | 35 | 59 | 59 | 35 | 1.231825 | 1.444060 | 0.401431 | 2.675885 |
| 300 | 60 | 60 | 48 | 60 | 60 | 0.667955 | 0.536301 | 0.240845 | 1.204255 |
| 301 | 60 | 60 | 60 | 60 | 60 | 0.544296 | 0.382930 | 0.306954 | 1.527166 |
| 302 | 60 | 48 | 47 | 60 | 60 | 0.588527 | 0.460612 | 0.299016 | 1.049139 |
| 303 | 60 | 58 | 47 | 60 | 60 | 0.865668 | 0.671793 | 0.380642 | 0.865668 |
| 304 | 60 | 60 | 48 | 60 | 60 | 0.975483 | 0.497615 | 0.263917 | 1.373098 |
| 305 | 61 | 48 | 59 | 60 | 61 | 0.910114 | 0.864000 | 0.295290 | 1.774115 |
| 306 | 61 | 61 | 47 | 61 | 61 | 0.795085 | 0.686811 | 0.249776 | 1.481896 |
| 307 | 61 | 61 | 47 | 61 | 61 | 1.158559 | 1.113342 | 0.356597 | 2.272501 |
| 308 | 59 | 61 | 47 | 59 | 61 | 0.836969 | 0.785480 | 0.314395 | 1.151364 |
| 309 | 59 | 56 | 59 | 61 | 61 | 0.963165 | 0.968998 | 0.324360 | 1.287525 |
| 310 | 48 | 3 | 42 | 62 | 62 | 0.645217 | 0.639942 | 0.330686 | 0.645217 |
| 311 | 27 | 42 | 55 | 62 | 62 | 0.389434 | 0.278537 | 0.251381 | 0.389434 |
| 312 | 52 | 59 | 11 | 62 | 59 | 0.447439 | 0.540349 | 0.284342 | 0.540349 |
| 313 | 59 | 49 | 55 | 62 | 62 | 0.684904 | 0.682475 | 0.352181 | 0.684904 |
| 314 | 61 | 35 | 29 | 62 | 61 | 0.449315 | 0.360302 | 0.263239 | 0.449315 |
| 315 | 63 | 63 | 47 | 63 | 63 | 0.738502 | 1.796971 | 0.174767 | 2.533474 |
| 316 | 63 | 48 | 47 | 63 | 63 | 0.543913 | 0.694755 | 0.142410 | 0.694755 |
| 317 | 63 | 47 | 11 | 63 | 63 | 1.083947 | 0.921003 | 0.188097 | 2.004950 |
| 318 | 63 | 48 | 47 | 63 | 63 | 0.548442 | 0.245267 | 0.181528 | 0.548442 |
| 319 | 63 | 63 | 47 | 63 | 63 | 0.946377 | 0.752785 | 0.154451 | 1.699162 |

**Figure 4.18: Summary of Hypothesis #5**

### 4.5.1 Conclusions from Hypothesis #5

The above table shows the predicted class labels from each classifiers along with the confidence values of each classifier on that sample which leads to the final decision making.

From the above table we can observe that class index #54 and #57 are difficult to classify correctly as each one of the single classifier fails to classify them. Also class index #59 and #43 are getting misclassified to class index #35 and #60 every single time, just like in the previous cases. A new observation reveals that class index #13 and #14 are indistinguishable, i.e. every time class #13 predicts #14 as its output.

| | |
|---|---|
| 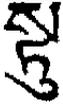 CLASS #181<br>CLASS INDEX #54 | 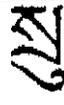 CLASS #217<br>CLASS INDEX #57 |
| 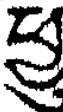 CLASS #55<br>CLASS INDEX #13 | 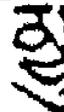 CLASS #57<br>CLASS INDEX #14 |
| 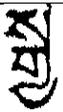 CLASS #220<br>CLASS INDEX #59 | 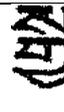 CLASS #115<br>CLASS INDEX #35 |

For this hypothesis to work decently, those samples are considered on which at least one classifier has predicted the correct class index. The number of such samples is 228. Also the number of correctly classified samples is 191. Thus the accuracy stands at 83.77 %.

**ACCURACY = 83.77 %**

The discarded samples are: 4, 18, 44, 49, 50, 51, 53, 54, 65, 66, 67, 68, 69, 74, 76, 87, 90, 91, 92, 93 ,94, 100, 103, 105, 113, 124, 136, 158, 159, 170, 172, 173, 175, 182, 185, 186, 187, 192, 193, 195, 203, 204, 212, 215, 216, 217, 218, 219, 223, 224, 235, 237, 238, 239, 244, 246, 247, 248, 249, 251, 254, 257, 261, 262, 263, 264, 267, 269, 270, 271, 272, 273, 274, 285, 286, 287, 288, 289, 290, 291, 293, 294, 296, 297, 298, 299, 309, 310, 311, 312, 313, 314.

# *Chapter 5*

## *Conclusion*

After analysing the obtained results, we can conclude that the theory of multi-hypothesis classifier is a success. In all the five cases above, the overall accuracy increases. Here is a brief summary of the results

| HYPOTHESIS | CLASSIFIER #1 | CLASSIFIER #2 | CLASSIFIER #3 | FINAL ACCURACY | % age CHANGE |
|---|---|---|---|---|---|
| Hypothesis #1 | 84.06 % | 79.06 % | 74.37 % | 87.45 % | +3.39 % |
| Hypothesis #2 | 84.06 % | 79.06 % | 74.37 % | 88.63 % | +4.57 % |
| Hypothesis #3 | 77.81 % | 77.18 % | 60.63 % | 83.26 % | +5.45 % |
| Hypothesis #4 | 77.81 % | 63.13 % | 4.38 % | 83.70 % | +5.89 % |
| Hypothesis #5 | 77.81 % | 63.13 % | 4.38 % | 83.77 % | +5.96 % |

**Table 5.1: Comparison of results of different hypotheses.**

- The highest accuracy obtained among the above hypotheses is by Hypothesis #2 at 88.63%

- The highest percentage increase among the above is by Hypothesis #5 at +5.96%.

- Classifier #1 has been consistently good performing best among the three classifiers.

- Some class labels are quite visually similar in appearance and could not be distinguished by the classifiers, e.g. class index #35 and #59, class index #43 and #60 among other few.

# *Chapter 6*

## *Future Scope*

Active research has been going on and there is still lot to be done in this field. Some future enhancements are suggested below:

1) **Increase number of Classifiers:** Some more number of classifiers can be used such as SVM, ANN, Fringe maps and the overall effect of using theses can be analysed.

2) **Sampling techniques:** One particular observation drawn from this research is the conflicts of results given on the test set and the validation set by a particular classifier for their respective sample numbers, i.e. for samples belonging to a particular class label, the prediction by a particular classifier on the data of test and validation set conflicts, which resulted in some samples being discarded and also affected in the overall accuracy. We have used a random sample selector to select our data randomly. A better way could be to use sampling techniques. Sampling techniques ensure that the selected samples are well represented by that class. As data plays a huge role in deciding the accuracy, intelligent way of sampling the data could be a major boost to improve the accuracy.

3) **Forming the perfect ensemble:** The basic assumption made on this research is that the classifiers are mutually exclusive i.e. statistically independent. Although we haven't exactly explored how to achieve this or how to prove if they are mutually complementary. Work can be pursued in this direction to form the perfect ensemble, which will have a direct impact on the accuracy.